\documentclass[letterpaper, 10 pt, journal, twoside]{IEEEtran}
\usepackage{amsmath}
\DeclareMathOperator{\st}{s.t.}
\usepackage{amssymb}    
\usepackage{mathtools} 
\usepackage{siunitx}
\usepackage{physics}
\usepackage{graphicx}
\usepackage{subfigure}
\usepackage[ruled,vlined]{algorithm2e}
\usepackage{tabularx}
\usepackage{enumitem}
\usepackage{balance}
\usepackage{fancyhdr}
\usepackage[hidelinks]{hyperref}
\newtheorem{definition}{Definition}
\newcommand{\T}{^\mathsf{T}}
\IEEEoverridecommandlockouts

\fancypagestyle{firstpage}
{
    \setlength{\headheight}{35pt}
    
    \fancyhead[L]{}    
    \fancyhead[R]{}
    \fancyhead[C]{This paper has been accepted for publication in the IEEE Transactions on Robotics, 2023. \\ Please cite the paper as: J. Huang, X. Chu, X. Ma, and K. W. S. Au, ``Deformable object manipulation with constraints using path set planning and tracking," IEEE Trans. Robot., vol. 39. no. 6, pp. 4671-4690, 2023.}
    \fancyfoot[C]{}
}

\begin{document}
\title{Deformable Object Manipulation With Constraints Using Path Set Planning and Tracking}
\author{Jing Huang, Xiangyu Chu, Xin Ma,  and Kwok Wai Samuel Au
\thanks{
This work was supported in part by the Chow Yuk Ho Technology Centre for Innovative Medicine of The Chinese University of Hong Kong, in part by the Multi-Scale Medical Robotics Centre, AIR@InnoHK, and in part by the Research Grants Council (RGC) of Hong Kong under Grants 14209118, 14209719, and 14211320. 
\textit{(Corresponding author: Kwok Wai Samuel Au.)}
}
\thanks{The authors are with the Department of Mechanical and Automation Engineering, The Chinese University of Hong Kong, Hong Kong, China. Jing Huang, Xiangyu Chu, and Kwok Wai Samuel Au are also with the Multi-Scale Medical Robotics Center, Hong Kong (e-mail: \href{mailto:huangjing@mae.cuhk.edu.hk}{huang\hfill jing@mae.cuhk.edu.hk}; \href{mailto:xychu@mae.cuhk.edu.hk}{xychu@mae.cuhk.edu.hk};  \href{mailto:maxin1988maxin@gmail.com}{maxin1988maxin@gmail.com};\hfill \href{mailto:samuelau@cuhk.edu.hk}{samuelau@cuhk.e-\hfill du.hk}).}
}

\pagenumbering{arabic}
\maketitle
\thispagestyle{firstpage}

\begin{abstract}
In robotic deformable object manipulation (DOM) applications, constraints arise commonly from environments and task-specific requirements. Enabling DOM with constraints is therefore crucial for its deployment in practice. However, dealing with constraints turns out to be challenging due to many inherent factors such as inaccessible deformation models of deformable objects (DOs) and varying environmental setups. 
This article presents a systematic manipulation framework for DOM subject to constraints by proposing a novel path set planning and tracking scheme. First, constrained DOM tasks are formulated into a versatile optimization formalism which enables dynamic constraint imposition. Because of the lack of the local optimization objective and high state dimensionality, the formulated problem is not analytically solvable.
To address this, planning of the path set, which collects paths of DO feedback points, is proposed subsequently to offer feasible path and motion references for DO in constrained setups. Both theoretical analyses and computationally efficient algorithmic implementation of path set planning are discussed. Lastly, a control architecture combining path set tracking and constraint handling is designed for task execution. The effectiveness of our methods is validated in a variety of DOM tasks with constrained experimental settings.
\end{abstract}
\IEEEpeerreviewmaketitle

\section{Introduction}
\label{Introduction Section}
\begin{figure}[t]
    \minipage{1 \columnwidth}
    \centering
    \includegraphics[width= 0.92 \columnwidth]{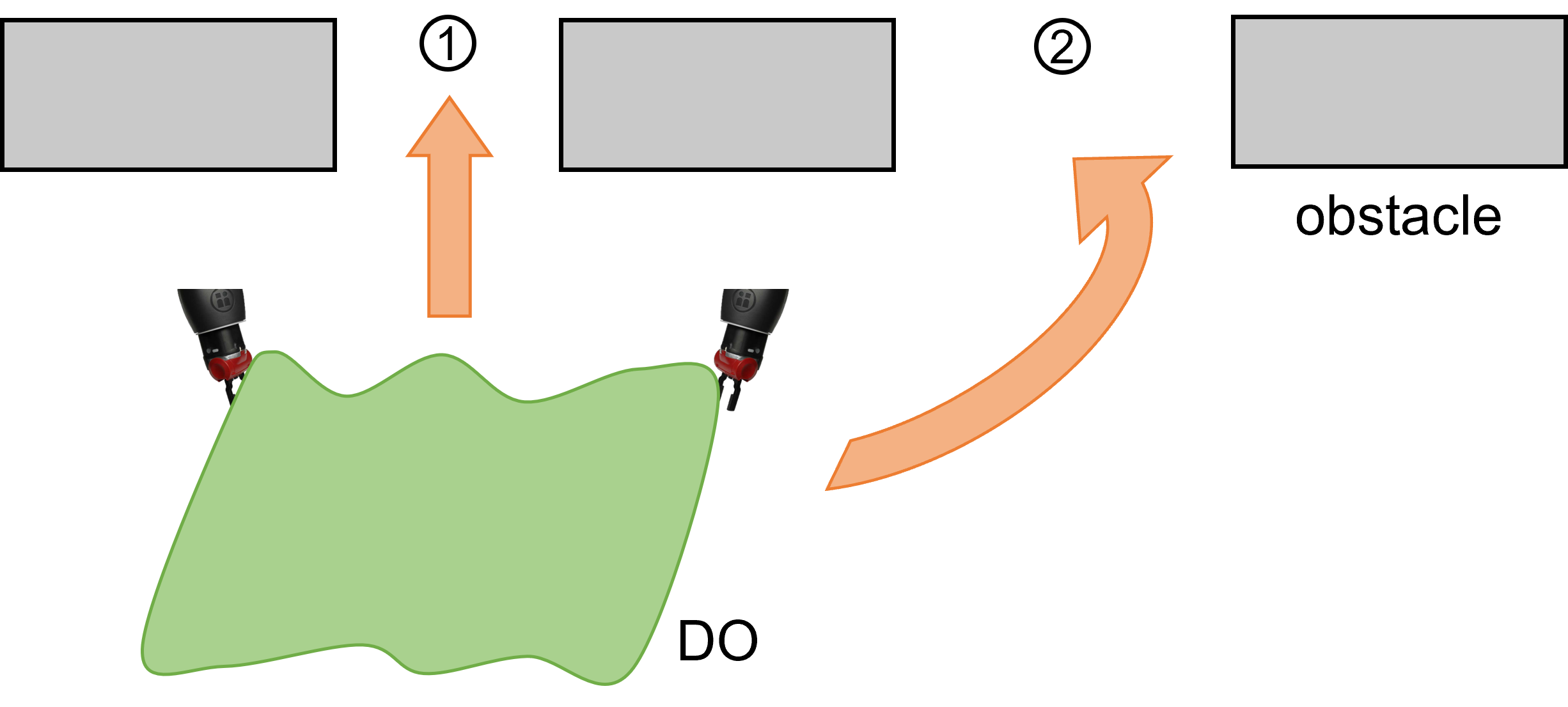}  
    \endminipage \hfill
    \caption{Piece of cloth is intended to be led by the robot to pass the narrow passage $1$ or $2$ to reach the desired position. Which passage (1 is closer but narrower) to pass and how to pass it are unknown. Path set planning aims to find an appropriate spatial path for a DO in constrained environments.}
    \label{Introduction Example}
    \end{figure}

\IEEEPARstart{D}{eformable} object manipulation (DOM) yields a fundamental branch of robotic manipulation with broad domestic and industrial applications. Though significant progress has been achieved \cite{D. Navarro-Alarcon 2014 ijrr}-\cite{D. Mcconachie 2018 TASE}, autonomous DOM is far from ready for deployment in common real-life tasks. Aside from longstanding challenges such as unattainable models of deformable objects (DOs) and unstructured task procedures \cite{J. Sanchez 2018, H. Yin 2021}, one core limitation hindering DOM from practical usage is that existing methodologies are unable to conduct tasks subject to constraints well (see Fig. \ref{Introduction Example}). Constraints are quite ubiquitous in DOM for the common presence of constrained environments and task-specific setups. For instance, avoidance of collision with environment obstacles \cite{F. Lamiraux 2001, D. Mcconachie 2020} and regulation of DO over-deformation \cite{D. Berenson 2013, D. Mcconachie 2018 TASE, J. Huang 2021} impose various forms of constraints. Given the inherent complexity of deformation modeling, control, and interaction in most DOM tasks, performing them with constraints is nontrivial.

Compared to rigid object manipulation, constraints in DOM can usually be harder to be managed. Firstly, the environments and task procedures in DOM are usually unstructured, which makes interactions hardly predictable and therefore results in a high collision risk. Secondly, DOs' time-varying modalities of shape and size dramatically increase the task indeterminacy and often require extra regulation. In consequence, the capability to cope with constraints, particularly the constraints related to various interactions and DOs' deformation, is crucial for the utility and applicability of newly developed DOM methods. Nevertheless, conducting DOM in the presence of constraints is difficult and remains insufficiently studied to date. Previous works focused more on the deformation control problem and usually assumed unconstrained setups \cite{D. Navarro-Alarcon 2014 ijrr}-\cite{D. Berenson 2013}, \cite{F. Ficuciello 2018}-\cite{Z. Hu 2018}. Even if some constraints were involved, they would rely on explicit deformation modeling or simulation to determine feasible DO states and robot motions prior to manipulation, leading to DO-specific and task-specific methods with quite high complexity \cite{F. Lamiraux 2001, M. Saha 2007, O. Roussel 2015}. Currently, a general approach able to effectively handle constraints in DOM is still absent.

\begin{figure}[t]
    \minipage{1 \columnwidth}
    \centering
    \includegraphics[width=  1 \columnwidth]{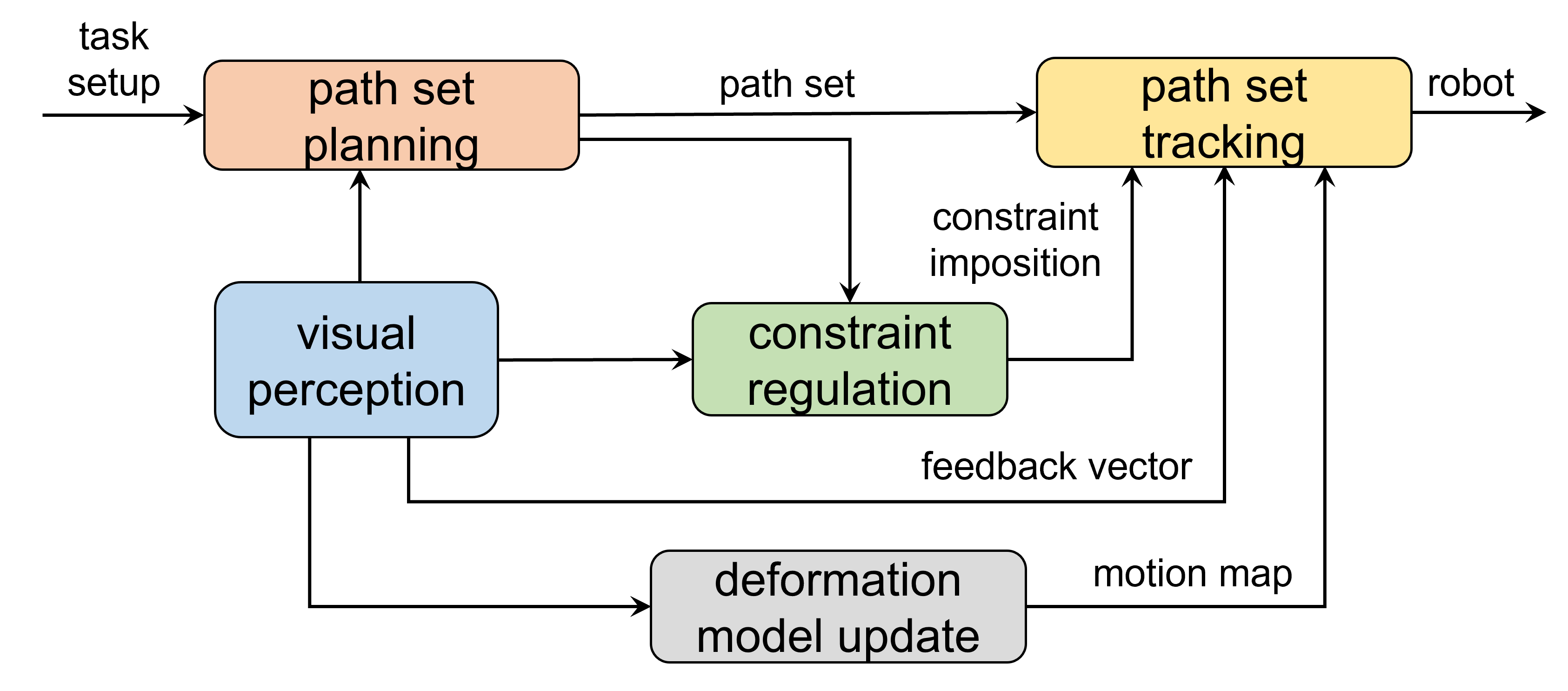}    
    \endminipage \hfill
    \caption{Block diagram illustrating the pipeline of the proposed manipulation framework for DOM with constraints.}
    \label{Task Block Diagram}
    \end{figure}

This article investigates DOM with constraints and proposes a novel path set planning and tracking pipeline. As illustrated in Fig. \ref{Task Block Diagram}, constraint regulation, path set planning and tracking modules in the proposed manipulation framework constitute new additions to conventional pure control approaches \cite{D. Navarro-Alarcon 2014 ijrr}-\cite{D. Mcconachie 2018 TASE}. Taking into account representative constraints, DOM tasks with constraints are first formulated into a versatile optimization formalism which enables dynamic constraint imposition. This imposition mechanism is implemented through constraint regulation in order to reinforce the task feasibility under constraints. With constraints, the formulated problem is not directly solvable by pure control approaches, necessitating the introduction of planning methodologies. There are no unified and easily tractable motion/path planning methods in DOM. To address this, path set planning for the visual feedback vector is presented in this article. The path set collects feedback points' paths and encodes essential path and motion references for the DO. Equally importantly, it is efficiently obtainable based on the modified optimal planners that involve passage encoding and selection without relying on explicit deformation modeling or simulation. Both theoretical analyses and algorithmic implementation for path set planning are detailed. Finally, the control architecture subsumes constraint regulation, path set tracking, and local minimum handling to execute the task in a path set tracking manner. To sum up, this scheme integrates path planning into constrained DOM efficiently via a generic task-level planning and local-level control paradigm.

\subsection{Related Work}
The related work of this article mainly includes manipulation with constraints, deformation control, and motion/path planning for DOM. Manipulation with constraints is a classical topic in robotics. Early works dealt with constraints by exploiting the redundancy in robot degree of freedom (DOF) with the task-priority framework \cite{A. Liegeois 1977}, \cite{B. Siciliano 1991}, where constraints were imposed as secondary tasks, e.g., configuration singularities \cite{S. Chiaverini 1997}, joint limits \cite{T. F. Chan 1995, B. Siciliano book}, and workspace obstacles \cite{A. A. Maciejewski 1985 ijrr}. 
Based on a hierarchy of quadratic programs, the framework incorporating inequalities was proposed and elaborated in \cite{O. Kanoun TRO}, \cite{A. Escande 2014}, but secondary tasks were still satisfied in the least-square sense. Task sequencing was introduced in \cite{N. Mansard TRO}, \cite{N. Mansard TRO 2009} which separated the global task into several subtasks and dynamically activated subtasks for constraint imposition. 

Deformation control has been extensively studied mostly in unconstrained settings \cite{J. Sanchez 2018, H. Yin 2021}. Model-based methods were broadly used in early studies, where the DO was endowed with an explicit physical model \cite{F. Lamiraux 2001}, \cite{S. Hirai 2000}-\cite{R. Jansen 2009}. While precise deformation simulation is achievable using finite element methods (FEMs) \cite{F. Ficuciello 2018, S. Patil 2010}, the cumbrous tuning and high computational cost inhibit its real-time application. Model-free approaches have thus become much more prevalent recently. In general, the model-free property is achieved through numerical methods \cite{D. Navarro-Alarcon 2013 TRO}, \cite{F. Alambeigi 2018}, or the adoption of generic model assumptions applicable to most DOs \cite{D. Navarro-Alarcon 2014 ijrr}-\cite{D. Mcconachie 2018 TASE}, \cite{Z. Hu 2018}. Numerical methods approximate the unknown map from robot motions to deformation measurements using techniques like fast Gaussian process regression \cite{Z. Hu 2018} and Broyden's method \cite{C. G. Broyden 1965}. Once an approximate map is available, the deformation control will take feedback control such as passivity-based control \cite{D. Navarro-Alarcon 2013 TRO}, and direct error-driven control \cite{F. Alambeigi 2018}. Commonly-used generic model assumptions include the affine deformation model \cite{D. Navarro-Alarcon 2014 ijrr}, energy-based formulation of DOs \cite{D. Navarro-Alarcon 2016 TRO}, and diminishing rigidity Jacobian \cite{D. Berenson 2013, D. Mcconachie 2018 TASE}, many of which are embedded into an adaptive control framework. Recent years have seen an increasing number of machine learning techniques applied to DOM, which can also be classified as model-free approaches. Some works use deep neural networks to learn the deformation models \cite{Z. Hu 2019}-\cite{M. Yu 2023}. Rather than separately addressing the unknown models, the action policy of robots is more and more often directly learned by reinforcement learning \cite{B. Thananjeyan 2017}-\cite{R. Jangir 2020}. Although learning-based approaches now are restricted to the specific trained tasks, they are promising due to their potential to combine both model prior and data \cite{H. Yin 2021}.

The presence of constraints makes DOM far more difficult and DOM with constraints has not been sufficiently studied to date. Environmental collision and the elastic limit were taken into account when planning paths for elastic objects \cite{F. Lamiraux 2001} and elastic rods \cite{O. Roussel 2015}, where an explicit geometrical representation of DOs and the elasticity model were utilized. 
When manipulating deformable linear objects (DLOs), the stretching range was constrained in \cite{M. Saha 2007}. In \cite{A. Wang 2019}, the manipulation plan of the DO (e.g., a rope) was learned from raw images by a generative adversarial network (GAN) wherein environment obstacles were processed by learning feasible DO paths. Avoidance of excessive stretching and collision was implemented by null-space projection based on the assessed deformation model in \cite{D. Berenson 2013, D. Mcconachie 2018 TASE}.
In existing work, constraint imposition highly relies on the adopted manipulation framework. Specifically, in DOM planning phase, constraints are validity check conditions. In DOM control, constraints are resolved with the classical task-priority strategy in manipulator control without a higher level of integration into the holistic task.

Manipulation planning has been investigated in many DOM tasks. A survey on model-based DOM planning can be found in \cite{P. Jimenez 2012}. Most existing works are application-specific, e.g., manipulation of DLOs, planar objects, and clothes  \cite{M. Saha 2007, M. Moll 2006}-\cite{A. Doumanoglou 2016}. Some studies focus on elementary motion paradigms such as folding and bending \cite{Liang Lu 2000}. In general, only standard planning algorithms are utilized in these works to find a path to the target state characterized by extracted DO geometrical or topological properties. Then the robot follows the planned path by using the prespecified actuating relation between the robot and DO. More relevant to our proposed work is DO path planning where a feasible path is searched for to connect DO's initial and target configurations in complex environments. In such problems, sampling-based approaches are employed \cite{P. Jimenez 2012}. First, random DO configurations are sampled based on the preset deformation model assumption, whose feasibility is later examined by criteria such as collision, internal energy, and geometric properties. A nominal feasible path is then found for robot execution \cite{O. Burchan Bayzait 2002, R. Gayle 2005}. The main drawback of this avenue is that sample generation and examination are complicated and rely on computationally costly simulation. Moreover, it is not resilient to model inaccuracy and actuation error since no feedback is exploited. To alleviate these, DO state prediction, motion planning, and control were combined in an interleaved way in \cite{D. Mcconachie 2020} with many task specifications. Recently, planning in the latent space of the learned DO dynamics model has been more and more explored \cite{W. Yan 2021, M. Lippi 2023}.

 \subsection{Contributions}
To endow robots with the crucial capability to conduct DOM under constraints, this article provides an efficient and general approach to combining path planning and control. Specifically, the following key contributions are made:
\begin{enumerate}
    \item A novel manipulation framework for DOM with constraints based on path set planning and tracking. In this framework, dynamic constraint imposition is employed in the task formulation. The strategy to determine constraint imposition states is also developed.
    \item Comprehensive analyses of the formulated path set planning problem for the visual feedback vector in the feature deformation description method. Novel, general, and efficient algorithms for path set generation compatible with existing optimal planners are proposed. 
    \item A holistic control architecture integrating constraint regulation, path set tracking, and local minimum resolution to accomplish the constrained manipulation task in a path set tracking manner.
\end{enumerate}

The remainder of this article is structured as follows. Section \ref{Formulation Section} formulates DOM tasks with constraints after specifying typical constraints and task setups. Section \ref{Planning Section} details the concept and key properties of path sets for feedback vectors as well as path planning preparations. Section \ref{Path Set Transfer Section} proposes algorithms for path set generation. Section \ref{Control Section} introduces path set tracking control in DOM execution. Section \ref{Experiment Section} presents experimental results. Finally, Section \ref{Conclusion Section} concludes this article.

\section{Formulation of DOM Tasks With Constraints}
\label{Formulation Section}
In this section, we first specify representative constraints in DOM considered in this work and introduce the task setups. Then the task formulation with dynamic constraint imposition is given in an optimization formalism.

\subsection{Constraint Specification}
\subsubsection{DO-Obstacle Collision Constraint}
 Restricted and confined workspace is ubiquitous in manipulative tasks. Unless stated otherwise, the constraint of being collision-free between the DO and environment is imposed by default. Suppose $\mathcal{O}$ is the set of DO points $\mathbf{p}_O \in \mathbb{R}^3$. $\mathcal{E}$ is an obstacle composed of points $\mathbf{p}_E \in \mathbb{R}^3$. The DO distance to $\mathcal{E}$ takes the minimum among all the point distances, i.e.,
\begin{equation}
    d(\mathcal{O, E}) = \min_{\mathbf{p}_O \in \mathcal{O}, \mathbf{p}_E \in \mathcal{E}} \| \mathbf{p}_O - \mathbf{p}_E \|_2.
\end{equation}
The repulsive potential is usually calculated as $ P_{\mathcal{E}}(\mathcal{O}) = \frac{k_{\mathcal{E}}}{2 d^2(\mathcal{O, E})}, k_{\mathcal{E}} \in \mathbb{R}_{+}$. The collision constraint requires 
\begin{equation}
    c_1: \; d(\mathcal{O, E}) > 0.
\end{equation}
$P_{\mathcal{E}}(\mathcal{O})$ needs to be smaller than a threshold.

\subsubsection{Robot-Obstacle Collision Constraint}
For the interaction between the robot and obstacles, only collision of the end-effector $\mathbf{r} \in \mathbb{R}^3$ is considered. Robot's distance to the obstacle $d(\mathbf{r}, \mathcal{E})$ and the repulsive potential $P_{\mathcal{E}}(\mathbf{r})$ are defined analogously as above. Thus the obstacle avoidance constraint for the robot is
\begin{equation}
    c_2: \; d(\mathbf{r}, \mathcal{E}) > 0.
\end{equation}
There usually exists rigid fixture between DO and end-effector, and $d(\mathcal{O}, \mathcal{E}) > 0$ naturally guarantees $d(\mathbf{r}, \mathcal{E}) > 0$. However, to accurately and safely control the robot and for analysis clarity, the distinction between them is preferable.

\subsubsection{DO Shape Constraint}
A distinctive class of constraints in DOM originate from the DO. To depict deformation, we consider the classical discrete description method. A small number of key feedback points are exploited, based on which deformation features are constructed to extract deformation properties of interest.
Specifically, $K$ feedback points $\mathbf{p}_{s_i} \in \mathbb{R}^3$ are picked on the DO with associated image projections $\mathbf{s}_i \in \mathbb{R}^2$. The deformation feature $\mathbf{y} \in \mathbb{R}^m$ is structured by
\begin{equation}
\label{visual feature eq}
    \mathbf{y} = \mathcal{F}(S)
\end{equation}
where $S = [\mathbf{s}_1\T \; \mathbf{s}_2\T \; ... \; \mathbf{s}_K\T]\T \in \mathbb{R}^{2K}$ is the visual feedback vector. $\mathcal{F}(\cdot): \mathbb{R}^{2K} \mapsto \mathbb{R}^m$ for $m \leq 2K$ is the feature extraction function. While features are versatile in deformation description, an underlying risk is undesired over-deformation, e.g., over-compression or over-stretch \cite{D. Mcconachie 2020}, \cite{J. Huang 2021}, which may cause severe DO damages such as plastic deformation and pathological damages of living tissues \cite{X. Chu RA-L 2018}. To avoid this, the DO shape constraint is introduced to manage vulnerable shape characteristics $\mathbf{h} \in \mathbb{R}^h$ not directly controlled in $\mathbf{y}$. $\mathbf{h}$ is constrained in a range 
\begin{equation}
    c_3: \; \mathbf{\underline{h}} \leq \mathbf{h} \leq \overline{\mathbf{h}}.
\end{equation}
$\mathbf{h}$ is assumed to be given by $\mathbf{h} = \mathcal{H}(S)$ also from $S$.

\subsection{Task Formulation With Dynamic Constraint Imposition}
The goal of a DOM task is typically framed as manipulating the DO to achieve the desired feature $\mathbf{y}_d$ specifying the deformation target. Let $S_d$ be the corresponding desired feedback vector for $\mathbf{y}_d$. Note $S_d$ need not be uniquely determined by $\mathbf{y}_d = \mathcal{F}(S_d)$ since $\mathcal{F}(\cdot)$ can be non-injective. For instance, $\mathbf{y}$ may partially rely on the relative distribution of $S$, e.g., the centroid of $S$. We assume $\mathbf{y}$ is at least \textit{complete}, which implies that there exists some feedback point in $S$ with a determinate target position. The completeness of $\mathbf{y}$ is required in the planning phase to eliminate the indeterminacy of $S_d$. In manipulation, the end-effector and the DO are connected rigidly at a fixed grasping position so that robot motions $\Delta \mathbf{r}$ are seamlessly transferred to the DO \cite{J. Huang 2021}. With the visual feedback, the manipulator is controlled kinematically by specifying the end-effector velocity $\mathbf{\dot{r}}$.

\begin{figure}[t]
    \minipage{1 \columnwidth}
    \centering
    \includegraphics[width= 1 \columnwidth]{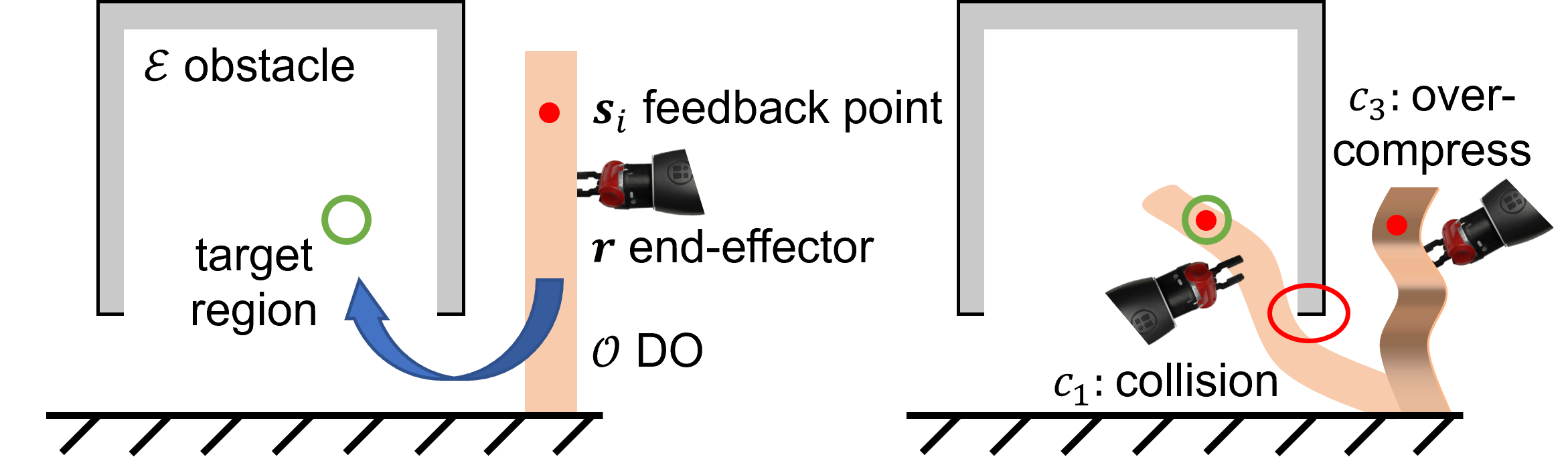}    
    \endminipage \hfill
    \caption{Robot needs to manipulate the DO to pass through a narrow passage. In this task, over-compression and collision with the environment have a high probability to happen.}
    \label{Narrow Passage Task}
    \end{figure}

Consider a DOM task subject to the constraints specified above. In practice, to efficiently accomplish a task, some constraints need not be strictly enforced throughout the entire task. One motivating example is shown in Fig. \ref{Narrow Passage Task}. To move the DO (e.g. a sponge strip) into the constrained target region, over-compression of the DO is performed first. When passing the narrow passage, the DO can contact the wall easily. However, such temporary constraint violations should be allowed for better task feasibility and efficiency. Therefore, appropriate constraint regulation is necessary to achieve dynamic constraint activation. To this end, constraints involved in the task are indexed from $1$ to $l$ as $c_i, \, i = 1, 2, ..., l$. A binary activation vector $\mathbf{a} = [a_1 \, a_2 \, ... \, a_l]\T \in \{0, 1\}^l$ with $a_i = 1$ if and only if $c_i$ is active encodes the constraint activation state.

For releasable constraints in $c_i$, the binary classification of $a_i$ is conducted according to task conditions. Suppose $T \in \mathbb{R}_{+}$ is the total task time and all the respected constraints are formulated as an equality $c_i = 0$ \textit{in form}. Then a general formalism of DOM tasks with constraints is
\begin{equation}
\label{Concept Optimizaion}
    \begin{split}
    &\operatorname*{min} \; \| \mathbf{y}(T) - \mathbf{y}_d \|^2 \\
    &\st  \; \mathbf{c}(t)\T \mathbf{a}(t) = 0, \;\; 0 \leq t \leq T\\
    \end{split}
\end{equation}
where the constraint vector $\mathbf{c} \in \mathbb{R}^l$ aggregates all the indexed constraints. For a feasible task, the final norm of the error term $\mathbf{e}_y(t) = \mathbf{y}(t) - \mathbf{y}_d$ can fall to zero, i.e., $\text{min} \, \|\mathbf{y}(T) - \mathbf{y}_d\|^2 = 0$. So the problem to be solved is finding a robot motion policy of $\dot{\mathbf{r}}(t)$ that achieves the minimum error norm while fulfills the active constraints in $\mathbf{c}(t)$ at each time step $t$.

The constraints introduced in the last subsection are considered in task formulation. To get a concise form consistent with (\ref{Concept Optimizaion}) and facilitate dynamic constraint imposition, the following modifications are applied. First, constraints are appointed as $c_1$ to $c_3$ in order. $c_i = 0$ formally implies no violation of $c_i$. Otherwise, $c_i = 1$. Second, the indicators $a_i$ are specified. For safety reasons, the end-effector collision constraint ($c_2$) always remains active, so $a_2 \equiv 1$, while the DO-obstacle collision constraint and DO shape constraint ($c_1$ and $c_3$) are allowed to be relaxed on some conditions to ensure the feasibility and efficiency of task execution. Therefore,
\begin{equation}
    \mathbf{a} = 
    [a_1 \;\; 1 \;\; a_3]\T.
\end{equation}

Incorporating specifications of $\mathbf{c}$ and $\mathbf{a}$ above, the task in the form of (\ref{Concept Optimizaion}) is
\begin{equation}
\label{Compact Optimizaion}
    \begin{split}
    &\operatorname*{min} \; \| \mathbf{y}(T) - \mathbf{y}_d \|^2 \\
    &\st  \; c_1 a_1 + c_2 + c_3 a_3 = 0.
    \end{split}
\end{equation}
This formulation enables dynamic constraint imposition, making it more versatile and practical, but it is hardly solvable for 
\begin{enumerate}
\item There is no local optimization objective available relating the current feature $\mathbf{y}(t)$ and the desired $\mathbf{y}_d$ which meanwhile meets the constraints in $\mathbf{c}(t)$; 
\item The objective function and constraints are formulated by different state variables.
\end{enumerate}
As such, pure control approaches become inapplicable. Relying on explicit deformation modeling or simulation will be unscalable, sophisticated, and computationally costly. In this article, path planning is resorted to for providing both local motion reference and global path reference for the DO.

\section{Path Set of Visual Feedback Vector in DOM}
\label{Planning Section}
In order to establish the relationship between intermediate $S(t), \mathbf{y}(t)$ and their final states, path planning is imperative to find reasonable reference paths for $\mathbf{s}_i$. This helps to alleviate the task indeterminacy and to figure out feasible DO motions in complex environments. This section elaborates on the concept of path set for the feedback vector $S$, related properties, and preparations for path set planning.

\subsection{Optimal Path Planning for An Individual Feedback Point}
We first briefly discuss optimal path planning for an individual feedback point in the DOM context. Suppose point $\mathbf{s}_{i}$ in a complete deformation feature vector $\mathbf{y}$ has the definite target region $\mathcal{S}_{goal}$, a feasible path from its initial position $\mathbf{s}_{i,init}$ to $\mathcal{S}_{goal}$ optimizing a user-defined cost function is obtainable using optimal planers such as sampling-based algorithms \cite{Karaman S. 2011 ICRA}-\cite{J.D. Gammell 2014}. Any optimal planner can be utilized for single point path planning and RRT$^*$ (optimal rapidly-exploring random tree) is taken as the subroutine in our implementation.
\begin{algorithm}[t]
\nl $V \leftarrow \{ \mathbf{s}_{i,init} \}$; $E \leftarrow \emptyset$;
$\delta \leftarrow k_\delta d({\mathbf{s}_{i},  \mathcal{C}})$\;
\nl \For {$i = 1, 2, ..., N$} {
    \nl $\mathbf{s}_{rand} \leftarrow \texttt{SampleFree}(\delta)$\;
    \nl $\mathbf{s}_{nearest} \leftarrow \texttt{Nearest}(G = (V, E), \mathbf{s}_{rand})$\;
    \nl $\mathbf{s}_{new} \leftarrow \texttt{Steer}(\mathbf{s}_{nearest}, \mathbf{s}_{rand})$\;
    \nl \If{\textnormal{\texttt{ObstacleFree}$(\mathbf{s}_{nearest}, \mathbf{s}_{new})$}} {
        \nl $S_{near} \leftarrow \texttt{Near}(G = (V, E), \mathbf{s}_{new}, r_{near})$\;
        \nl $V \leftarrow V \cup \{ \mathbf{s}_{new} \}$\;
        \nl $\mathbf{s}_{min} \leftarrow \texttt{GetParent}(S_{near}, \mathbf{s}_{new})$\;
        \nl $E \leftarrow E \cup \{(\mathbf{s}_{min}, \mathbf{s}_{new})\}$\;
        \nl $\texttt{Rewire}(G = (V, E), S_{near}, \mathbf{s}_{min}, \mathbf{s}_{new})$\;
    }
    }
    \nl \Return $G = (V, E)$\;
\caption{RRT$^*$ for An Individual Feedback Point.}
\label{RRTStar}
\end{algorithm}

Constraints in (\ref{Compact Optimizaion}) cannot be enforced at the path planning stage of a single point. One crucial quantity in planning is the minimum clearance between the path and obstacles. In particular, sampling is restricted in the $\delta$-interior of the obstacle-free configuration space $\mathcal{X}_{free}$, denoted by int$_\delta(\mathcal{X}_{free})$ with int$_\delta(\mathcal{X}_{free}) = \{\mathbf{x} \in \mathcal{X}_{free} \, | \, \mathcal{B}_{\mathbf{x}, \delta} \subseteq \mathcal{X}_{free} \}$ and $\mathcal{B}_{\mathbf{x}, \delta}$ being the closed ball of radius $\delta$ centered at $\mathbf{x}$. A small $\delta$ may cause DO's percolation of obstacles when following the planned path. A large $\delta$ is advantageous since more free space is provided for DO motion. But prior to the task, determining a feasible $\delta$ as large as possible in a confined workspace will take multiple trials. The distance between $\mathbf{s}_{i}$ and DO boundary $\mathcal{C}$ is
\begin{equation}
\label{distance sp and C}
    d({\mathbf{s}_{i},  \mathcal{C}}) = \operatorname*{min}_{\mathbf{s}_{j} \in \mathcal{C}} \| \mathbf{s}_{i} - \mathbf{s}_{j} \|_2.
\end{equation}
$\delta > d({\mathbf{s}_{i},  \mathcal{C}})$ is merely necessary for obstacle avoidance for the undeformed DO, which is not necessarily satisfiable in planning. $\delta$ is now assigned as $k_\delta d({\mathbf{s}_{i},  \mathcal{C}})$ for $k_\delta > 1$ and will be further refined. 
Algorithm \ref{RRTStar} shows the RRT$^*$-based optimal path planning for an individual feedback point. The sampling space is restricted to int$_\delta(\mathcal{X}_{free})$ in function \texttt{SampleFree}($\delta$) and the rest follows the same procedure in \cite{Karaman S. 2011 ICRA, Karaman S. 2011}, where the cost associated with a node adopts the shortest path length to the root $\mathbf{s}_{i,init}$.

\subsection{Passage-Aware Path Planning}
Since determining the large feasible $\delta$-interior for planning is hard in constrained environments, we propose to exploit passage passing states in path planning, which allows a suitable path to be found even with a small $\delta$. Denote workspace obstacles as $\mathcal{E}_i (i = 1, ..., M)$. For any pair $\mathcal{E}_i, \mathcal{E}_j$ ($1 \leq i < j \leq M$), it forms a generic passage $(\mathcal{E}_i, \mathcal{E}_j)$. For simplicity, the passage is represented by the obstacle centroid segment with obstacle dimensions extruded. A total of $(^M_{\,2})$ such passages exist, but not all of them correspond to physically valid ones. Using the visibility condition \cite{de B. Mark 2008}, $(\mathcal{E}_i, \mathcal{E}_j)$ is classified as a valid passage if the passage segment is collision-free with other obstacles. Formally, a path in $\mathcal{X}_{free}$ is a continuous function $\sigma: [0, 1] \mapsto \mathbb{R}^{dim(\mathcal{X})}$ with finite length in $\mathbb{R}^{dim(\mathcal{X})}$. To handle constrained environments, path's passage passing information is encoded. For a path node $\sigma(\tau)$ ($\tau \in [0,1]$ is given by path length parameterization), the passages passed by $\sigma$ from the start $\sigma(0)$ to $\sigma(\tau)$ are stored in order as a list $P_{\sigma}(\tau) = \{(\mathcal{E}_i, \mathcal{E}_j), ...,  (\mathcal{E}_p, \mathcal{E}_q)\}$. $P_{\sigma}(\tau, i)$ indexes the $i$-th passage in $P_{\sigma}(\tau)$. $\| (\mathcal{E}_i, \mathcal{E}_j) \|_2$ returns the passage width and $\min \| P_{\sigma}(\tau) \|_2$ returns the minimum passage width if $P_{\sigma}(\tau)$ is nonempty.
In planning, passage passing is updated every time a new edge is checked as Algorithm \ref{Update passage encoding cost}. 
\begin{figure}[t]
    \minipage{1 \columnwidth}
    \centering
    \includegraphics[width= 0.8 \columnwidth]{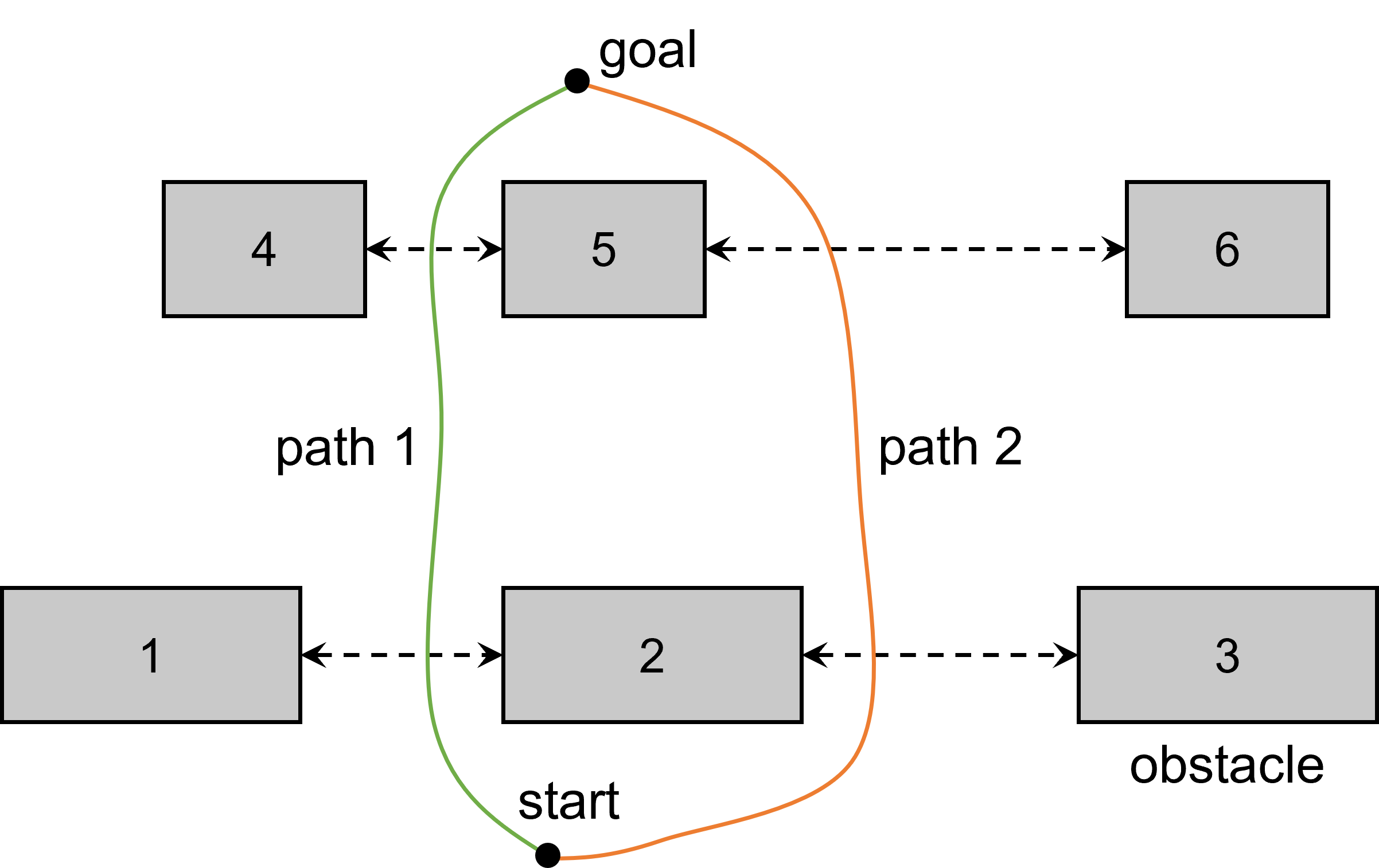}    
    \endminipage \hfill
    \caption{
    Path 1 is optimal under classical path optimization criterion, e.g., path length and execution time, but it admits a quite narrow passage $(\mathcal{E}_4, \mathcal{E}_5)$. Path 2 better trades off the path length and workspace along the path.}
    \label{Passage Encoding}
    \end{figure}

The aim of encoding passages is to select the path with sufficient workspace while being optimal under the chosen criterion. Nevertheless, these two objectives often conflict. In Fig. \ref{Passage Encoding}, two paths connect the same start and goal positions but pass different passages. Path $1$ will be considered better under classical path optimization criteria such as the path length. However, it is also associated with narrower workspace, which is not desirable for DOM in constrained environments. As aforementioned, determining a large $\delta$-interior in the path planning phase of a single point on DO is difficult. Instead, workspace traversed by a single path is more accessible. The goal is to pick the path best trading off the path length cost and narrow passages it goes through. To this aim, a composite cost function is defined as  
\begin{equation}
\label{Cost Function for Passage Passing}
    f_{\sigma} = Len(\sigma) / f_P(\sigma)
\end{equation}
where $Len(\sigma)$ returns the path length. $f_P(\sigma)$ can be taken as $\min \| P_{\sigma}(1) \|_2$ to reflect the preference for wider passages. In case of no traversed passages ($P_{\sigma}(1) = \emptyset$), a large value $\overline{\varepsilon}_P$ is assigned to $f_P$ for each path node.

To find the optimal path depicted in (\ref{Cost Function for Passage Passing}), the node cost is computed as  (\ref{Cost Function for Passage Passing}) in Algorithm \ref{RRTStar} subroutines $\texttt{GetParent()}$ and $\texttt{Rewire()}$. Since (\ref{Cost Function for Passage Passing}) is monotonic in terms of path concatenation and bounded, sampling-based optimal planners are asymptotically optimal to attain the optimal path \cite{Karaman S. 2011}. In practice, small $Len(\sigma)$ and $f_P(\sigma)$ can occur concurrently. To further ensure exclusion of overly narrow passages, $f_P(\sigma)$ is truncated downwards. In sum, $f_P(\sigma)$ is
\begin{equation}
    f_P(\sigma) = 
    \begin{cases}
        \overline{\varepsilon}_P    & P_{\sigma}(1) = \emptyset  \\
        \underline{\varepsilon}_P    & \min \|  P_{\sigma}(1) \|_2 \leq \underline{f}_P \\
        \min \|  P_{\sigma}(1) \|_2    &\text{otherwise}.
    \end{cases}
\end{equation}
$\underline{\varepsilon}_P$ is a small positive value satisfying $0 < \underline{\varepsilon}_P \ll \underline{f}_P$. $\underline{f}_P$ can be interpreted as the minimum passage width requirement for the ongoing DOM task. In this way, paths passing passages with widths smaller than $\underline{f}_P$ will be rejected. 
\begin{algorithm}[t]
\nl Input an $\mathbf{s}_{near} \in S_{near}$ and $\mathbf{s}_{new}$\;
\nl $L(\mathbf{s}_{new}) \leftarrow \mathbf{s}_{near}.cost * \mathbf{s}_{near}.min\_passage\_width + \|l(\mathbf{s}_{near}, \mathbf{s}_{new}) \|_2$\;
\nl $\mathbf{s}_{new}.min\_passage\_width \leftarrow \mathbf{s}_{near}.min\_passage\_width $\;
\nl \ForEach{\textnormal{valid passage} $(\mathcal{E}_i, \mathcal{E}_j)$} {
    \nl \If {$l(\mathbf{s}_{near}, \mathbf{s}_{new})$ \textnormal{passes} $(\mathcal{E}_i, \mathcal{E}_j)$ \textnormal{AND} $\|(\mathcal{E}_i, \mathcal{E}_j)\|_2 < \mathbf{s}_{new}.min\_passage\_width$} {
            \nl $\mathbf{s}_{new}.min\_passage\_width \leftarrow \|(\mathcal{E}_i, \mathcal{E}_j)\|_2$\;
        }
}
\nl $\mathbf{s}_{new}.cost \leftarrow L(\mathbf{s}_{new}) / {\mathbf{s}_{new}.min\_passage\_width}$\;   
\caption{Update A New Node Cost in (\ref{Cost Function for Passage Passing}).}
\label{Update passage encoding cost}
\end{algorithm}

\subsection{Prerequisites for Visual Feedback Vector Path Set Planning}
Optimal path planning and its passage-aware variant for a single DO point deal with only one $\mathbf{s}_i \in S$. This subsection concentrates on some crucial preparations for the development of the final path set planning method: the determination of $S$ target $S_d$ and the feasibility requirement for a path set. Path set planning for visual feedback vector aims to generate coordinated feasible paths for all separated points in $S$ simultaneously. Despite the apparent similarity, it is different from typical multi-robot path planning \cite{Yi Guo 2002}-\cite{R. Luna 2011} because points in $S$ are physically related rather than independent agents.

\subsubsection{Target Determination for Visual Feedback Vector}
A prerequisite of path planning is a determinate target position. Though the completeness requirement of $\mathbf{y}$ helps to restrict possible candidates, $\mathbf{y}_d$ does not suffice to define $S_d$ uniquely if infinitely many $S_d$ satisfy $\mathbf{y}_d = \mathcal{F}(S_d)$ (see Fig. \ref{Angle Feature Target Determination} for an example). Extra criteria thus need to be introduced to figure out an appropriate $S_d$. In such cases, we formulate $S_d$ determination as the following optimization problem
\begin{equation}
\label{sd optimization}
    \begin{split}
    \arg \, &\operatorname*{min}_{S} \; \mathcal{J}(S_0, S) \\
    \st & \; \mathcal{F}(S) = \mathbf{y}_d  \\
      &\mathbf{c}(T)\T \mathbf{1}_{3} = 0
    \end{split}
\end{equation}
where $\mathcal{J}(\cdot) \geq 0$ is the evaluation function for $S$ candidates which takes both obstacle avoidance and manipulation cost into account. $S_0$ is the initial value of $S$. The all-ones vector $\mathbf{1}_{3}$ indicates that all constraints are imposed in the target state. The specific form of $\mathcal{J}(\cdot)$ and solution details are provided in Appendix A.
In the following development, we will assume that $S_d$ is available and a \textit{pivot} $\mathbf{s}_p \in S$ is identified to be used in single point path planning. The pivot can be seen as the point in $S$ most influential by criteria such as feature Jacobian Frobenius norm or displacement in the task.

\subsubsection{Feasibility Requirement for Path Set}
This part discusses the feasibility requirement for path sets in manipulation from $S_0$ to $S_d$. One fundamental feasibility requirement for paths in the path set is related to path homotopy. Two collision-free/feasible paths $\sigma_1, \sigma_2$ with the same initial and final positions are path homotopic if there exists a continuous function called homotopy $\psi: [0, 1] \mapsto \Sigma_{free}$, where $\Sigma_{free}$ is the set of paths in $\mathcal{X}_{free}$, such that $\psi(0) = \sigma_1, \psi(1) = \sigma_2$ and $\psi(x) \in \Sigma_{free}, \, \forall \, x \in [0,1]$. Intuitively, homotopic paths can be continuously transferred to one another in $\mathcal{X}_{free}$ \cite{J.R. Munkres 2000}. Since feedback points are picked on the DO, their paths are path homotopic-like, which is simply verified by noticing that $\mathbf{s}_i$ can always exchange their positions within the space occupied by the DO (thus in $\mathcal{X}_{free}$) at any moment.

Two paths share identical initial and final positions in the homotopic relationship. For this reason, a straight-line path is first employed to construct such point paths. Suppose $\sigma_i$ is the path of $\mathbf{s}_i$ from $\mathbf{s}_{0,i}$ to $\mathbf{s}_{d,i}$. $\sigma_{i,j|d}$ is the straight-line path from $\mathbf{s}_{d,i}$ to $\mathbf{s}_{d,j}$, which connects the termini of $\sigma_i$ and $\sigma_j$ with the segment in between and satisfies $\sigma_{i,j|d}(0) = \sigma_{i}(1)$. $\sigma_{i,j|d} \in \Sigma_{free}$ commonly holds since such path connection is a local operation on the path. Similarly, the straight-line path $\sigma_{i,j|0}$ connects $\mathbf{s}_{0,i}$ and $\mathbf{s}_{0,j}$. For path $\sigma_1$ and $\sigma_2$ satisfying $\sigma_1(1) = \sigma_2(0)$, their concatenated path can be defined with path length parametrization as
\begin{equation}
    \sigma_1 * \sigma_2 (\tau) = 
    \begin{cases}
        \sigma_{1}(\tau / r)                & \tau \in [0, r],  \\
        \sigma_{2}((\tau - r) / (1 - r))    &\tau \in (r, 1].
    \end{cases}
\end{equation}
$r$ is the ratio between $\sigma_1$ length and the concatenated $\sigma_1 * \sigma_2$ length. By the associativity of path concatenation, the concatenated path of $\sigma_{j,i|0}, \, \sigma_i$, and $\sigma_{i,j|d}$ denoted by $\sigma_{i,j}'$ is
\begin{equation}
    \sigma_{i,j}' = \sigma_{j,i|0} \,*\, \sigma_i \,*\, \sigma_{i,j|d}.
\end{equation}
In this work, $\sigma_{i}$ and $\sigma_{j}$ are said to be \textit{path homotopic-like} if $\sigma_{i,j}'$ and $\sigma_j$ are path homotopic. It is easy to see that $\sigma_{j,i}'$ and $\sigma_i$ are path homotopic will also lead to $\sigma_i$ and $\sigma_j$ are path homotopic-like, simply implying that the path homotopic-like relation is symmetric (see Appendix B for proof). 
Symmetry makes the following homotopy properties and path transfer operations between any two point paths to be unordered. 
Let $\Sigma_{S}$ be the set consisting of paths of points $\mathbf{s}_i \in S$. For the path set $\Sigma_{S}$, we say it is \textit{set homotopic-like} if any two paths in $\Sigma_{S}$ are path homotopic-like. 
\begin{figure}[t]
    \minipage{1 \columnwidth}
    \centering
    \includegraphics[width= 0.95 \columnwidth]{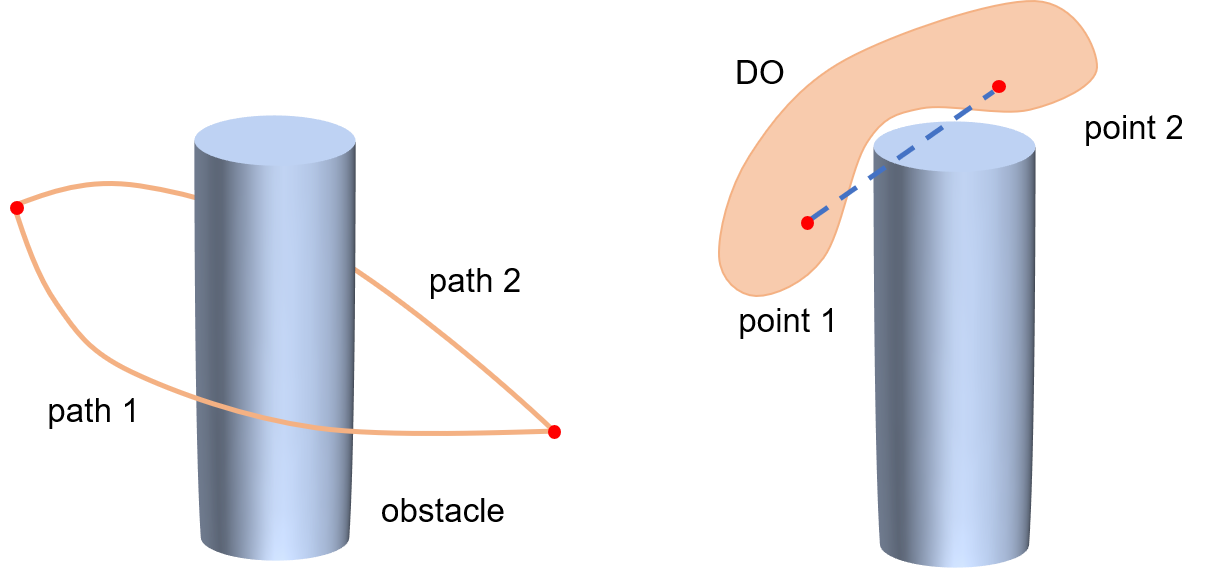}    
    \endminipage \hfill
    \caption{Path 1 and 2 are path homotopic. The composed path set (suppose paths are processed to have the same path ends) is infeasible. On the right, the segment between point 1 and 2 collides with the obstacle. Their path set is not strong homotopic-like, but the shown pose is feasible in DOM.}
    \label{Path Homotopy}
    \end{figure} 

$\Sigma_S$ is required to be set homotopic-like. While in Cartesian space, this is insufficient to guarantee the manipulation feasibility of the DO. See Fig. \ref{Path Homotopy} for an example where a homotopic-like path set is infeasible. To fix this, we further need $\Sigma_S$ to be \textit{strong homotopic-like}. For homotopic paths $\sigma_1, \sigma_2$, their straight-line homotopy has the following form
\begin{equation}
    \psi(x, \tau) = (1 - x) \sigma_1(\tau) + x \sigma_2(\tau), \;\; x, \tau \in [0, 1].
\end{equation}
If the straight-line homotopy $\psi(x, \tau)$ always lies in $\Sigma_{free}$, $\sigma_1, \, \sigma_2$ are said to be \textit{strong path homotopic}, which infers that the hypersurface swept by $\psi(x, \tau)$ is collision-free \cite{L. Jaillet 2008}. Similarly to above, $\sigma_i$ and $\sigma_j$ are strong path homotopic-like if $\sigma_{i,j}' \, (\sigma_{j,i}')$ and $\sigma_j \, (\sigma_i)$ are strong path homotopic. For a set of feasible paths, the corresponding definition is given as 
 \begin{definition}
 \label{Strong homotopic-like definition}
(Strong homotopic-like path set) A set of feasible paths $\Sigma_{S}$ is said to be strong homotopic-like if any two paths in $\Sigma_S$ are strong path homotopic-like.
\end{definition}
\begin{figure*}[t]
    \minipage{2 \columnwidth}
    \centering
    \includegraphics[width= 0.86 \columnwidth]{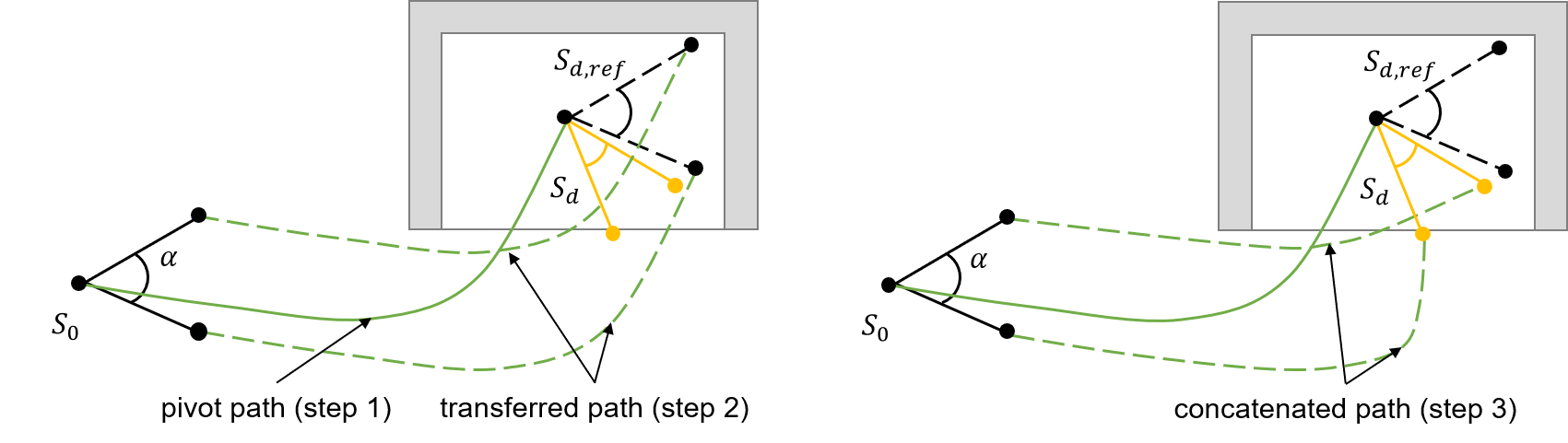}    
    \endminipage \hfill
    \caption{The pivot is the angle vertex and its path is designated by the solid green line. The transferred paths are designated by dashed green lines, which connect $S_0$ and $S_{d,ref}$. The final paths after truncation and concatenation are shown on the left.}
    \label{Path Set Planning Procedure}
    \end{figure*}

Such a homotopy relationship is also utilized similarly in path analysis of unmanned aerial vehicles (UAVs) as uniform visibility deformation \cite{B. Zhou 2020, B. Zhou 2021}, originated from the concept of visibility domain \cite{E. Schmitzberger 2002}, but has not been applied to DOM. 
By convention, both the empty set and singleton are strong homotopic-like. In practice, being strong homotopic-like represents a strong condition for path set feasibility. More precisely, it is sufficient but not necessary for $\Sigma_S$ to be feasible in DOM. For instance, Fig. \ref{Path Homotopy} demonstrates a case where $\Sigma_S$ is not strong homotopic-like but can be feasible (also see \cite{S. Bhattacharya 2012}). Physically, the essential condition for a path set to be feasible can be stated as, for any two paths (with end concatenation) in the set, there exists a homotopy that locates in the space traversed by the DO, but this is difficult to verify \textit{a priori} as the DO-traversed space is unavailable and hard to predict in the planning phase. In this article, our discussion is restricted to the strong homotopic-like path set with simplicity, verifiability, and fairly good generality.

\section{Path Set Generation Based on Path Transfer}
\label{Path Set Transfer Section}
For the goal of generating a feasible $\Sigma_S$ depicted above, a novel method built upon translation among feedback point paths and local path deformation is proposed in this section. The basic procedure for occasions where prespecified transfer assumptions hold is first elaborated. After that, path transfer in more general constrained conditions is presented. The chief advantages of the methods consist in 1) efficiency based on optimal path planning for the pivot, 2) guarantee of the strong homotopic-like property by construction.

\subsection{Basic Procedure}
To illustrate the basic procedure of path set generation based on path transfer, we first address the simple situation where sufficient workspace is available. There are three steps: 1) Reference pivot path generation, 2) Path transfer within $S$, and 3) Postprocessing. The first step plans a path $\sigma_p$ for the pivot $\mathbf{s}_p$ using optimal planners, which later will serve as the base path. A large $\delta$ set in the sampling space int$_\delta(\mathcal{X}_{free})$ benefits the validity of transferred paths. But a too large $\delta$ will result in a smaller sampling space and may prohibit finding a feasible path. To get an appropriate $\delta$, we consider the largest distance to $\mathbf{s}_p$ among points in $S$, i.e., $\Theta(S, \mathbf{s}_p) = \max_{\mathbf{s}_i \in S}(\| \mathbf{s}_i - \mathbf{s}_p \|_2)$. In the algorithm, we can further take the maximum between $\Theta(S_0, \mathbf{s}_{0,p})$ and $\Theta(S_d, \mathbf{s}_{d,p})$, i.e.,
\begin{equation}
\label{delta value}
    \delta_p = \max(\Theta(S_0, \mathbf{s}_{0,p}), \, \Theta(S_d, \mathbf{s}_{d,p})).
\end{equation}

It is possible that sampling in int$_{\delta_p}(\mathcal{X}_{free})$ makes the planner unable to find a feasible path for $\mathbf{s}_p$, because $\delta_p$ may be an overlarge clearance and the path search fails in a shrunk interior of $\mathcal{X}_{free}$. For simplicity and clarity in illustrating the basic procedure, the following \textit{transfer assumptions} are made in this part: 
\begin{enumerate}
\item A feasible path of $\mathbf{s}_p$ can be found in int$_{\delta_p}(\mathcal{X}_{free})$.
\item $S_{d,ref}$ is a feasible configuration. 
\end{enumerate}
$S_{d,ref}$ is the configuration of $S$ directly transferred by the pivot motion $\mathbf{v} = \mathbf{s}_{d,p} - \mathbf{s}_{0,p}$, i.e., $S_{d,ref} = \{ \mathbf{s}_{d, ref} \,|\, \mathbf{s}_{d,ref} = \mathbf{s}_{0,i} + \mathbf{v}, \mathbf{s}_{0,i} \in S_0 \}$.
Note $\sigma_p$ is planned without passage encoding when the transfer assumptions are met since int$_{\delta_p}(\mathcal{X}_{free})$ guarantees sufficient workspace for path sets.

In the second step, $\sigma_p$ is transferred to other points in $S$ to form a strong homotopic-like path set. If a path $\sigma$ is transferred by a vector $\mathbf{v}_t \in \mathbb{R}^{dim(\mathcal{X})}$, the new path is determined as $\sigma_{t}(\tau) = \sigma(\tau) + \mathbf{v}_t$. For each point in $S$, there is a path $\sigma_{t,i}$ transferred from $\sigma_p$ given by
\begin{equation}
\label{Basic path transfer}
    \sigma_{t,i} = \sigma_p + \mathbf{s}_i - \mathbf{s}_p.
\end{equation}
The resultant path set is denoted as $\Sigma_{t}(S, \mathbf{s}_p, \sigma_p)$ with triple arguments. Clearly, $\Sigma_{t}(S, \mathbf{s}_p, \sigma_p)$ is strong homotopic-like since $\sigma_p \subseteq $ int$_{\delta_p}(\mathcal{X}_{free})$ and $\|\mathbf{s}_i - \mathbf{s}_p\|_2 \leq \delta_p, \, \forall \, \mathbf{s}_i \in S$, which implies that all the transferred paths are feasible and moreover, the straight-line homotopy between paths lies in int$_{\delta_p}(\mathcal{X}_{free})$ since paths are located within a tunnel centered at $\sigma_p$. When performing path transfer, there exist two different options since both $\Sigma_t(S_0, \mathbf{s}_{0,p}, \sigma_p)$ and $\Sigma_t(S_d, \mathbf{s}_{d,p}, \sigma_p)$ can be adopted. Considering the chronological order of $S_0$ and $S_d$, if $\Sigma_t(S_0, \mathbf{s}_{0,p}, \sigma_p)$ is utilized, we call the transfer \textit{forward path transfer}. Correspondingly, \textit{backward path transfer} refers to the transfer of $\Sigma_t(S_d, \mathbf{s}_{d,p}, \sigma_p)$. 
\begin{algorithm}[t]
    \nl $\Sigma_S \leftarrow \emptyset$\;
    \nl $ \mathbf{s}_p, \, S_{d,ref}, \, S_d \leftarrow$ {Algorithm \ref{Target Feedback Vector Determination Algorithm}}\;
    \nl $\delta_p \leftarrow \max(\Theta(S_0, \mathbf{s}_{0,p}), \, \Theta(S_d, \mathbf{s}_{d,p}))$\;
    \nl $ \sigma_p \leftarrow$ {RRT$^*$ in Algorithm \ref{RRTStar}}\;
    \nl \ForEach{$\mathbf{s}_{i} \in S_0$} {
        \nl $\sigma_{t,i} \leftarrow \sigma_p + \mathbf{s}_i - \mathbf{s}_{0,p}$\;
        // \textit{search path $\sigma_{t,i}$ and return $\mathbf{s}_{min, i}$} \\
        \nl $\mathbf{s}_{min,i} \leftarrow \texttt{SearchByDistance}(\sigma_i, \mathbf{s}_{d,i}, L_i) $\; 
        \nl $\sigma_i \leftarrow \sigma_{t,i} * l(\mathbf{s}_{min,i}, \mathbf{s}_{d,i})$\;
        \nl $\Sigma_S \leftarrow \Sigma_S \cup \{\sigma_i\}$\;
    }
    \nl \Return $\Sigma_S$\;
\caption{Path Set Generation Based on Forward Path Transfer.}
\label{Path Set Generation}
\end{algorithm}

The transferred path set has not yet connected $S_0$ and $S_d$ because of their different point distributions. In postprocessing (the third step), local path concatenation is conducted to complete the transferred paths. The strong homotopic-like property can be preserved in general due to the spatial locality of postprocessing. In forward path transfer, the destination of $\Sigma_t(S_0, \mathbf{s}_{0,p}, \sigma_p)$ is $S_{d,ref}$ with identical point distribution of $S_0$. From $S_{d,ref}$ to $S_d$, point distribution deforms and various path completion methods exist, e.g., straight-line concatenation and minimum distance projection to the transferred path. In this work, the points on $\sigma_{t,i}$ with distance equal to the segment length $L_i = \|\mathbf{s}_{d,i} - \mathbf{s}_{ref,i}\|_2$ to $\mathbf{s}_{d,i}$ are found as the connection position candidates
\begin{equation}
\label{Path Concatenation Point}
    S_{L,i} = \{\mathbf{s}_{l} \in \sigma_{t,i} \,\, | \,\, \|\mathbf{s}_{d,i} - \mathbf{s}_l\|_2 = L_i \}
\end{equation}
where $\mathbf{s}_{ref,i} = \sigma_{t,i}(1) \in S_{L,i}$ as a trivial member. The point in $S_{L,i}$ with the shortest accumulated path length is selected as $\mathbf{s}_{min,i}$. To complete the path to $\mathbf{s}_{d,i}$, the portion of $\sigma_{t,i}$ from $\mathbf{s}_{min,i}$ to its end $\sigma_{t,i}(1)$ is truncated and the left path, denoted as $\sigma_{t,i}'$, is reconnected to $\mathbf{s}_{d,i}$ with a straight-line path, i.e., $\sigma_i = \sigma_{t,i}' * l(\mathbf{s}_{min,i}, \mathbf{s}_{d,i})$ where $l(\cdot)$ is the straight-line path between two points, which is generally feasible. A similar procedure also applies to backward path transfer.

See Fig. \ref{Path Set Planning Procedure} for an example and conceptual illustration of the basic forward path transfer procedure and Algorithm \ref{Path Set Generation} for the algorithmic implementation. From an ideal path tracking perspective, the backward transferred path set $\Sigma_S(S_d, \mathbf{s}_{d,p}, \sigma_p)$ usually has inferior practicality since deformation (local path concatenation part) needs to be accomplished first when there is much manipulation (transferred path part) to be finished. Therefore, the forward path transfer is selected.  
When the transfer assumptions hold, the probabilistic completeness of the path set planning method using the basic procedure is guaranteed by the algorithm for pivot path planning. The path set feasibility is ensured with the path set optimality reflected by the pivot path optimality. 
\begin{figure}[t]
    \minipage{1 \columnwidth}
    \centering
    \includegraphics[width= 0.81 \columnwidth]{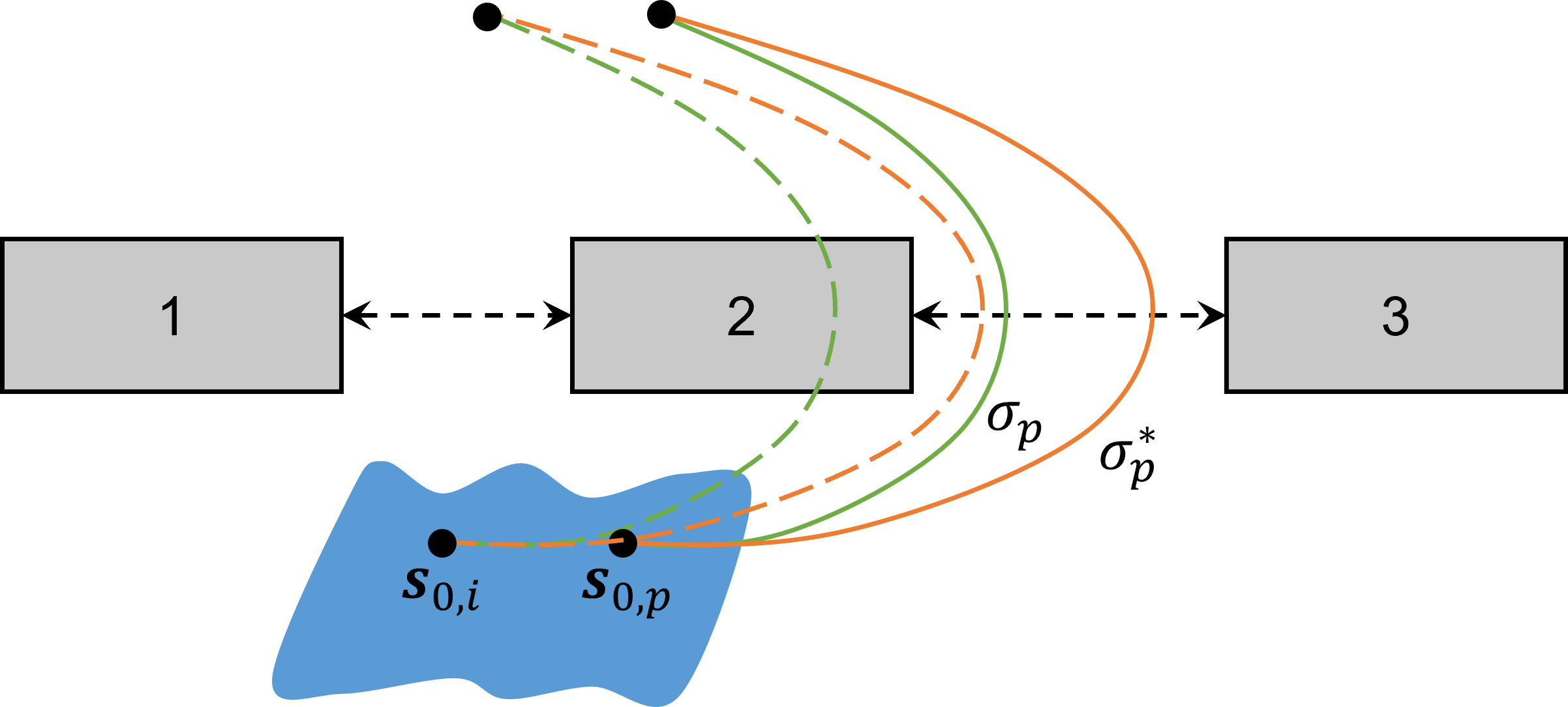}    
    \endminipage \hfill
    \caption{Directly planned pivot path $\sigma_p$ may be not beneficial for generating feasible transferred paths. It is repositioned to benefit the path transfer process. Dashed lines represent directly transferred paths.}
    \label{Pivot Path Repositioning}
    \end{figure}

\subsection{Path Set Generation in General Constrained Conditions} 
In fact, transfer assumptions impose rather strict restrictions. In general constrained conditions, they may easily fail to hold. Then, the objective of pivot path planning lies in finding a path $\sigma_p$ with a large distance to obstacles while being optimal under the chosen criterion. To achieve this, an exhaustive trial of the interior clearance could be performed by decreasing it from $\delta_p$ to zero in a binary-search way, which requires invoking the path planning routine multiple times. Furthermore, as $\sigma_p$ is no longer necessarily in int$_{\delta_p}(\mathcal{X}_{free})$, direct path transfer in (\ref{Basic path transfer}) will lead to infeasible path segments if the transferred point $\sigma_p(\tau) + \mathbf{s}_i - \mathbf{s}_p$ is in collision. Also, postprocessing via direct path concatenation will be infeasible if $S_{d, ref}$ is infeasible. These make up the main challenges of path set generation in general constrained conditions. To address these, passage-aware encoding path planning is utilized for $\sigma_p$. Then, for coordinated path set generation, a deformable path transfer procedure is proposed with two parts: 1) Pivot path repositioning, and 2) Deformable path transfer. 
\begin{algorithm}[t]
    \nl $\Sigma_t, \Sigma^*_S \leftarrow \emptyset$\;
    \nl $ \mathbf{s}_p, \, S_d \leftarrow$ Algorithm \ref{Target Feedback Vector Determination Algorithm}\;
    \nl $ \sigma_p \leftarrow$ {RRT$^*$ in Algorithm \ref{RRTStar} with  composite cost (\ref{Cost Function for Passage Passing})}\;
    \nl $\Sigma_t \leftarrow$ {(\ref{Basic path transfer}) using $\sigma_p$}\;
    \nl \ForEach{$P_{\sigma_p}(1,i) \in P_{\sigma_p}(1)$} {
        \nl $\{\sigma_{t,1}(\eta_{1,i}), ..., \sigma_p(\eta_{p,i}), ..., \sigma_{t,K}(\eta_{K, i})\} \leftarrow \Sigma_t \cap P_{\sigma_p}(1,i)$\;
        \nl $ \| \Sigma_t \cap P_{\sigma_p}(1,i) \|_2 \leftarrow$ (\ref{Chord Length})\;
        \nl \If{$\| P_{\sigma_p}(1,i) \|_2 < \| \Sigma_t \cap P_{\sigma_p}(1,i) \|_2$} {
            \nl $\sigma_p^*(\eta_{p,i}) \leftarrow$ (\ref{Pivot path replacement optimization})\;
        }    
        \nl \ElseIf{$(\Sigma_t \cap P_{\sigma_p}(1,i)) \nsubseteq  P_{\sigma_p}(1,i)$} {
            \nl $\sigma_p^*(\eta_{p,i}) \leftarrow$ Move $\sigma_p(\eta_{p,i})$ in passage such that $\Sigma_t \cap P_{\sigma_p}(1,i) \in$ init$_{\underline{\delta}}(\mathcal{X}_{free})$\;
        }
        \nl \Else {
            \nl Discard $P_{\sigma_p}(1,i)$\ and $\sigma_p(\eta_{p,i})$;
        }
    }    
    \nl \ForEach{$[\eta_{p,i}, \eta_{p,i+1}]$} {
    // \textit{including the first interval} $[0, \eta_{p,1}] \,\, \textit{and the last interval} \,\, [\eta_{p, -1}, 1]$ \\
        \nl $\sigma_p^*(\tau) \leftarrow$ (\ref{Deform Pivot Path})\;
    }    
    \nl $\Sigma_t \leftarrow$ {(\ref{Basic path transfer}) using $\sigma^*_p$}\;
    \nl \ForEach{$\mathbf{s}_i \in S_0$} {
        \nl  $\{\sigma_{t,i}^*(0), ..., \sigma_{t,i}^*(\eta_{i,j}), ..., \sigma_{t,i}^*(1)\} \leftarrow$ (\ref{Reposition Intersection Points})\;
        \nl $\sigma_{t,i}^* \leftarrow$ Reposition $\sigma_{t,i}$ as (\ref{Deform Pivot Path})\; 
        \nl $\Sigma^*_S \leftarrow \Sigma^*_S \cup \sigma_{t,i}^*$\;
     }
    \nl \Return $\Sigma^*_S$\;
\caption{Path Set Generation Using Deformable Path Transfer in General Constrained Conditions.}
\label{General Path Set Generation}
\end{algorithm}

\subsubsection{Pivot Path Repositioning}
Pivot path repositioning modifies the pivot path to achieve more coordinated transferred paths. As illustrated in Fig. \ref{Pivot Path Repositioning}, a raw $\sigma_p$ found by (\ref{Cost Function for Passage Passing}) does not take account of its specific passage passing positions. As such, unfavorable passage passing locations often exist. This will make transferred paths overly infeasible because infeasible parts of transferred paths are far away from the passage. To tackle this, the pivot path is repositioned to a more reasonable configuration. Denote the intersection point between $\sigma_p$ and the passage line of $P_{\sigma_p}(1,i)$ as $\sigma_p(\eta_{p,i})$. For a directly transferred path $\sigma_{t,j}$ obtained by (\ref{Basic path transfer}), its intersection point with the same passage line is $\sigma_{t,j}(\eta_{j,i})$. For the directly transferred path set $\Sigma_{t}(S, \mathbf{s}_p, \sigma_p)$, $\{\sigma_{t,1}(\eta_{1,i}), ..., \sigma_p(\eta_{p,i}), ..., \sigma_{t,K}(\eta_{K, i})\}$ collects all the intersection points with $P_{\sigma_p}(1,i)$ lines. The chord $\Sigma_t \cap P_{\sigma_p}(1,i)$ is defined as the intersection segment of passage $P_{\sigma_p}$ and path set $\Sigma_t$, whose length is given by
\begin{equation}
\label{Chord Length}
    \| \Sigma_t \cap P_{\sigma_p}(1,i) \|_2 = \max_{1 \leq k,j \leq K} \| \sigma_k(\eta_{k,i}) - \sigma_j(\eta_{j,i}) \|_2.
\end{equation}
Namely, a chord is determined by the intersection ends between the path set and passage line. Also note that $ \| \Sigma_t \cap P_{\sigma_p}(1,i) \|_2$ can be greater than $\| P_{\sigma_p}(1,i) \|_2$ because the chord investigates the passage line rather than the segment.

Pivot path repositioning adjusts $\sigma_p$ and consequently the distribution of $\{\sigma_1(\eta_{1,i}), ..., \sigma_p(\eta_{p,i}), ..., \sigma_K(\eta_{K, i})\}$ to facilitate following path transfer. For $\| P_{\sigma_p}(1, i) \|_2 < \| \Sigma_t \cap P_{\sigma_p}(1,i) \|_2$,  the following centering criterion is considered
\begin{equation}
\label{Pivot path replacement optimization}
    \begin{split}
    \arg \, &\operatorname*{min}_{\sigma_p(\eta_{p,i})} \; \| \bar{\sigma}_{1-K,i} - \bar{P}_{\sigma_p}(1,i) \|_2  \\
    &\st \; \sigma_p(\eta_{p,i}) \in P_{\sigma_p}(1, i) \cap \text{init}_{\underline{\delta}}(\mathcal{X}_{free}) \\ 
    \end{split}
\end{equation}
where $\bar{\sigma}_{1-K,i}$ is the center of the chord. $\bar{P}_{\sigma_p}(1,i)$ is the center of passage $P_{\sigma_p}(1,i)$. $P_{\sigma_p}(1, i) \cap \text{init}_{\underline{\delta}}(\mathcal{X}_{free})$ sets a small clearance $\underline{\delta}$ between $\sigma_p$ and obstacles. If $\| P_{\sigma_p}(1, i) \|_2 \geq \| \Sigma_t \cap P_{\sigma_p}(1,i) \|_2$, sufficient passage space is available. $\sigma_p(\eta_{p,i})$ can be simply translated along the passage to make directly transferred paths lie in init$_{\underline{\delta}}(\mathcal{X}_{free})$.  
If the chord totally lies within the passage,  $\sigma_p(\eta_{p,i})$ does not need to be repositioned. $\sigma_p(\eta_{p,i})$ and the passage $P_{\sigma_p}(1, i)$ are discarded to avoid deteriorating the final path's smoothness.

In each remaining passage, a new intersection point of $\sigma_p$, designated as $\sigma_p^*(\eta_{p,i})$, is given by (\ref{Pivot path replacement optimization}) which is readily solvable. Using new passage intersection points and path end points as references, $\sigma_p$ is repositioned in an iterative manner. We utilize a linear mapping between two successive points for repositioning. $\sigma_p^*(\eta_{p,i}) - \sigma_p(\eta_{p,i})$ is the shift at $P_{\sigma_p}(1, i)$. Analogously, the next shift is $\sigma_p^*(\eta_{p,i+1}) - \sigma_p(\eta_{p,i+1})$. The path segment $\sigma_p(\tau)$ for $\eta_{p, i} \leq \tau \leq \eta_{p, i+1}$ is given as 
\begin{multline}
\label{Deform Pivot Path}
    \sigma_p^*(\tau) = \sigma_p(\tau) + \frac{\tau - \eta_{p, i}}{\eta_{p, i+1} - \eta_{p, i}}(\sigma_p^*(\eta_{p,i+1}) - \sigma_p(\eta_{p,i+1})) \\ + \frac{\eta_{p, i+1}  - \tau}{\eta_{p, i+1} - \eta_{p, i}}(\sigma_p^*(\eta_{p,i}) - \sigma_p(\eta_{p,i})).
\end{multline}
As passage regions are assumed small compared to the total path length, the feasibility of the shifted path $\sigma^*_p(\tau)$ is assured by the fact that two path segment ends are placed within two successive passages respectively.
\begin{figure}[t]
    \minipage{1 \columnwidth}
    \centering
    \includegraphics[width= 0.92 \columnwidth]{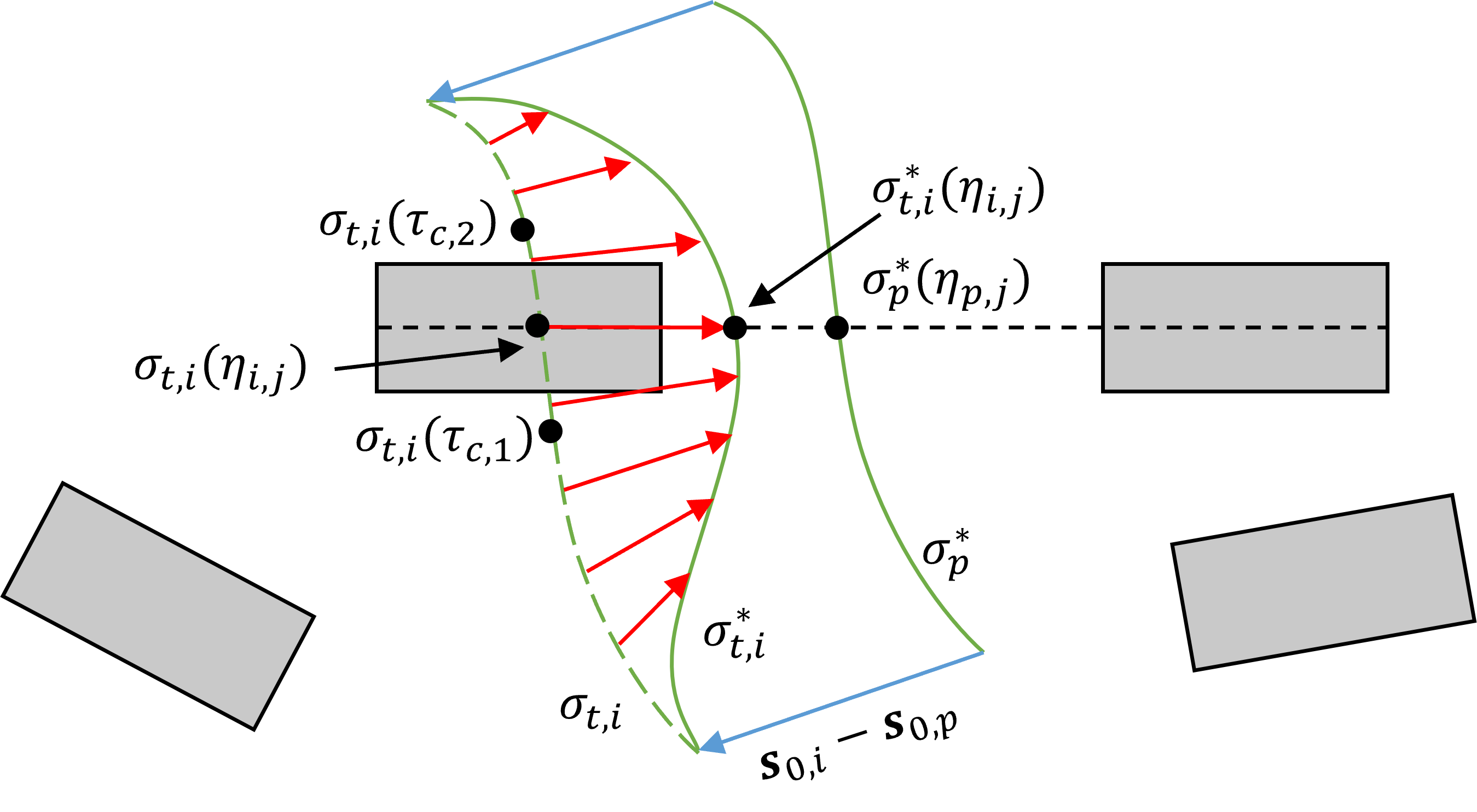}
    \endminipage \hfill
    \caption{Though the pivot path is repositioned, the directly transferred paths are still possible to be infeasible. Deformable path transfer deforms the infeasible path segments to obtain feasible transferred paths.}
    \label{Transferred Path Repositioning}
    \end{figure}

\subsubsection{Deformable Path Transfer}
Pivot path repositioning adjusts $\sigma_p$ locally without changing its topological properties. Deformable path transfer allows $\sigma_p$ to be transferred even if the transfer assumptions fail. The core to achieving this is a topology-preserving path deforming process. Firstly, transferred paths $\sigma_{t,i}$ are regenerated utilizing the basic procedure with the repositioned pivot path $\sigma_p^*$. To tailor potential infeasible parts of transferred paths, deformable path transfer is proposed. If $\sigma_{t,i}$ collides with an obstacle $\mathcal{E}_{near}$ nearby, it needs to reshape the collision part. Since $\sigma_p^*$ is repositioned, the collision of $\sigma_{t,i}$ indicates that $\mathcal{E}_{near}$ is associated with some narrow passages. Suppose this passage is $P_{\sigma_{t,i}}(1, j)$ and the intersection point is $\sigma_{t,i}(\eta_{i,j})$. $\sigma_{t,i}(\eta_{i,j})$ collides with $\mathcal{E}_{near}$ and the path collision segment is characterized by two path nodes before and after collision $\sigma_{t,i}(\tau_{c,1}), \sigma_{t,i}(\tau_{c,2}) \, (\tau_{c,1} < \eta_{i,j} < \tau_{c,2})$ respectively (see Fig. \ref{Transferred Path Repositioning}).

Similarly to the path-guided optimization (PGO) approach in \cite{B. Zhou 2021} which turns a locally infeasible path to a feasible one effectively, the infeasible segment of $\sigma_{t,i}$ is modified with $\sigma_{p}^*$ acting as the guiding path. Particularly, in the direction of $\sigma_p^*(\eta_{p,j}) - \sigma_{t,i}(\eta_{i,j})$, $\sigma_{t,i}(\eta_{i,j})$ is moved to $\mathcal{X}_{free}$. The exact shifted position of $\sigma_{t,i}(\eta_{i,j})$ is determined by proportionally compressing the chord of the directly transferred path set. Using the pivot intersection point as a fixed reference, each intersection point is shifted within the passage, i.e.,
\begin{equation}
\label{Reposition Intersection Points}
   \frac{\| \sigma_{t,i}^*(\eta_{i,j}) - \sigma_p^*(\eta_{p,j}) \|_2}{\|  \sigma_{t,i}(\eta_{i,j}) - \sigma_p^*(\eta_{p,j}) \|_2} = \min(\frac{\gamma_1}{\beta_1}, \frac{\gamma_2}{\beta_2})
\end{equation}
where $\sigma_{t,i}^*(\eta_{i,j})$ represents the new intersection point. Suppose $P_{\sigma_p}(1,j) = (\mathcal{E}_{j,1}, \mathcal{E}_{j,2})$. Then $\gamma_1 = d(\sigma^*_p(\eta_{p,j}), \mathcal{E}_{j,1})$ is the distance between $\sigma^*_p(\eta_{p,j})$ and $\mathcal{E}_{j,1}$. $\beta_1$ is the maximum distance between $\sigma^*_p(\eta_{p,j})$ and other intersection points which is closer to $\mathcal{E}_{j,1}$ than $\mathcal{E}_{j,2}$, i.e.,
\begin{multline}
\label{l Defination}
       \beta_1 = \max_{i} \| \sigma_{t,i}(\eta_{i,j}) - \sigma^*_p(\eta_{p,j} \|_2 \;\; \text{for} \\ d(\sigma_{t,i}(\eta_{i,j}), \mathcal{E}_{j,1}) < d(\sigma_{t,i}(\eta_{i,j}), \mathcal{E}_{j,2}).
\end{multline}
$\gamma_2$ and $\beta_2$ are analogously defined for $\mathcal{E}_{j,2}$.
In the sense of (\ref{Reposition Intersection Points}), it is guaranteed that all paths have feasible passage intersection points to lead the following deformable path transfer.

Lastly, deformable path transfer leverages a similar procedure of linear mapping in (\ref{Deform Pivot Path}) to obtain feasible transferred paths. Transferred paths are feasible in general if obstacle dimensions are negligible. Collision in the vicinity of the passage can be resolved by adding more repositioning points in the collision region to push the path away from obstacles.
To concatenate the transfer path end and the desired $\mathbf{s}_i$, the end position of $\sigma_{t,i}$ is set as $\mathbf{s}_{d,i}$ in deformable path transfer so that no extra postprocessing is required in a unified pipeline. See Algorithm \ref{General Path Set Generation} for the overall procedure of path set generation in general constrained conditions.
\begin{figure}[t]
    \minipage{1 \columnwidth}
    \centering
    \includegraphics[width= 0.76 \columnwidth]{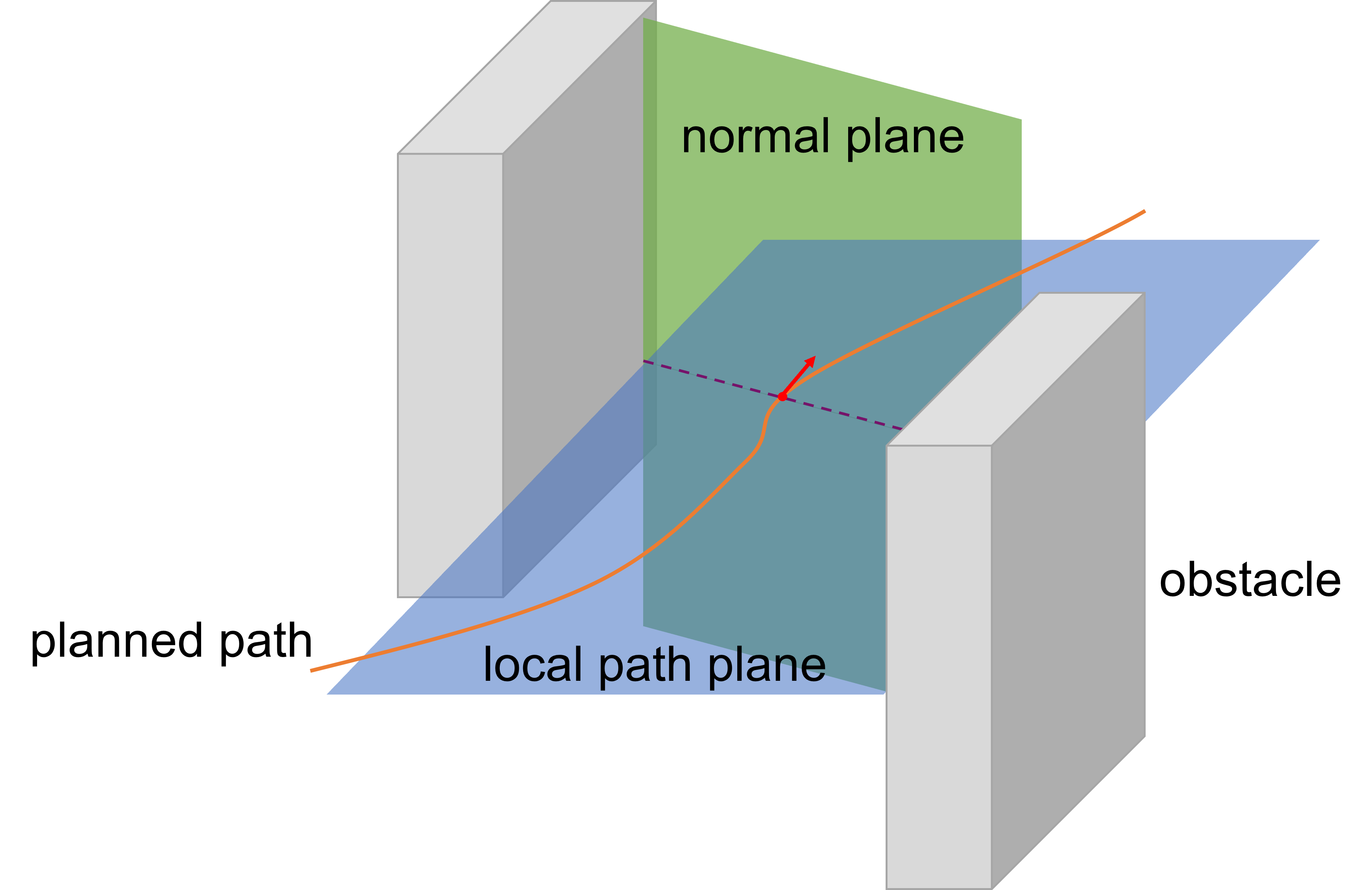}
    \endminipage \hfill
    \caption{Conceptual illustration of the local path width. The dashed purple line is the local path width of a path point (the starting point of the red normal vector arrow), which is truncated by obstacles.}
    \label{Local Path Width}
    \end{figure}

\section{Path Set Tracking Control for Task Execution}
\label{Control Section}
This section elaborates on the manipulation control of DOM tasks formulated in Section \ref{Formulation Section} and endowed with the planned path set of the feedback vector in Section \ref{Planning Section}, \ref{Path Set Transfer Section}. The proposed control architecture builds up a path set tracking controller composed of several core components. First, constraint regulation is discussed. Then, a path set tracking control structure interleaving path set tracking and end-effector path tracking is presented to ensure the task progress.

\subsection{Constraint Regulation} 
Constraint regulation determines the activation states of constraints dynamically imposed in (\ref{Compact Optimizaion}). Relaxation of a constraint implies that under the given condition, it is difficult to let manipulation proceed with the constraint imposed. In principle, constraints should only be relaxed when necessary and reimposed timely. In this work, environment information is exploited in conjunction with the planned path set to determine the activation states of releasable constraints $c_1$ and $c_3$. Specifically, local path width serves as the metric of constraint violation risk (see Fig. \ref{Local Path Width}). Intuitively, it gauges the available workspace volume in the vicinity of the path. The associated local path width of $\sigma$ is $\mu: [0, 1] \mapsto \mathbb{R}_+$ defined as
\begin{equation}
    \mu (\tau)  = \|\, \mathcal{N}(\sigma, \tau) \cap \mathcal{P}(\sigma, \tau) \cap \mathcal{X}_{free} \,\|_2
\end{equation}
where $\mathcal{N}(\sigma, \tau)$ represents the path's normal plane at $\sigma(\tau)$. 
$\mathcal{P}(\sigma, \tau)$ is the plane on which the path lies locally near $\sigma(\tau)$. Both are available analytically utilizing path's local geometry properties. Further intersection with $\mathcal{X}_{free}$ gives rise to the local path width segment at $\sigma(\tau)$. 
\begin{figure}[t]
    \minipage{1 \columnwidth}
    \centering
    \includegraphics[width= 1 \columnwidth]{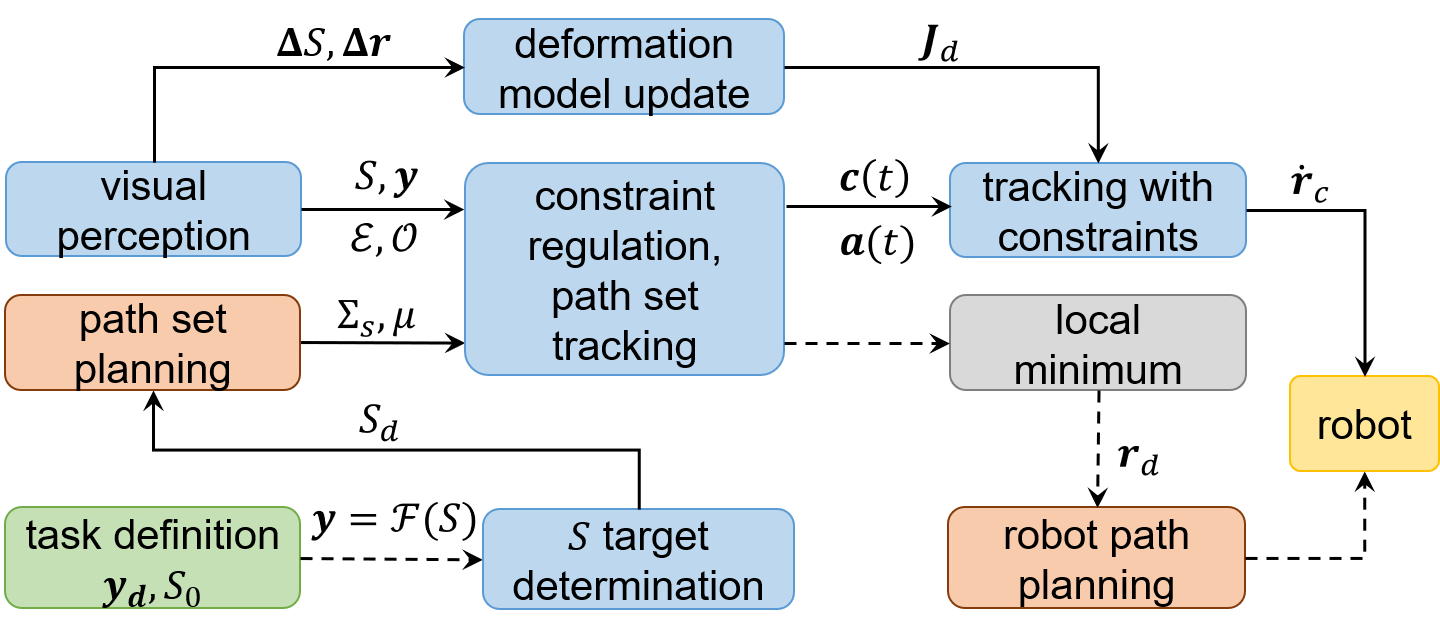}    
    \endminipage \hfill
    \caption{Block diagram of the overall control system. The dashed line implies that this flow is active in certain conditions.}
    \label{System Flow Chart}
\end{figure}

Because of the passage encoding of paths, local path width computation can be saved when locating narrow workspace. Narrow passages will correspond to local minima of $\mu(\tau)$. It thus suffices to recognize narrow passages' positions. There are in general multiple paths in $\Sigma_S$, but only one needs to be considered due to the shared passage encoding information and surroundings of paths in $\Sigma_S$. In particular, the pivot position $\mathbf{s}_p(t)$ and its planned path $\sigma_p$ are employed as references in constraint regulation. The risky segments in terms of constraint violation are obtained by
\begin{equation}
    W = \{\sigma_p(\tau) \,|\, \tau \in P_{\sigma_p}^{N}(1) \}.
\end{equation}
$P_{\sigma_p}^N(1)$ stands for the narrow passage neighborhoods, which can be selected from $P_{\sigma}(1)$ via $\| P_{\sigma_p}(1, i) \|_2 \leq \| \Sigma_t \cap P_{\sigma_p}(1,i) \|_2$ or specified conditions related to the ongoing task. $W$ serves as the main criterion for the  activation of $c_1$, i.e., suppose the path is perfectly tracked, $a_1$ is simply given by the following rule
\begin{equation}
    a_1 = 
    \begin{cases}
        0       &\mathbf{s}_p(t) \in W \\
        1       &\text{otherwise}.
    \end{cases}
\end{equation}
A looser activation criterion of the DO shape constraint $c_3$ is employed considering its temporal characteristic. Transient violation of $c_3$ is acceptable and adverse effects on the DO are assumed to be caused only after a long violation. Therefore, $c_3$ is allowed to be temporarily relaxed in the task and will be checked in the target configuration.

\subsection{Path Set Tracking And Constraint Adjustment}
Fig. \ref{System Flow Chart} details the overall control system.
Note that the planned path set serves the gross motion reference purpose for achieving the target $S_d$ by tracking $\Sigma_S$. However, precise path set tracking for multiple paths throughout the task is difficult under constraints and also not a must. In other words, the ultimate goal is making the DO achieve $S_d$ while being feasible during the task. For clarity, the tracking control is first discussed without constraint violation, followed by scenarios with active constraints. For real-time visual feedback vector $S(t)$, its distance to the path set $\Sigma_S$ is defined as
\begin{equation}
    \mathbf{e}(t) = S(t) - \Sigma_S
\end{equation}
with $\mathbf{e}_i = \mathbf{s}_i - \sigma_i := \mathbf{s}_i - \arg \, \operatorname*{min} \|\mathbf{s}_i - \mathbf{s}_{\sigma_i}\|_2, \, \mathbf{s}_{\sigma_i} \in \sigma_i$. The corresponding path argument of $\mathbf{s}_{\sigma_i}$ is $\tau_{e,i}$, i.e., $\mathbf{s}_{\sigma_i} = \sigma_i(\tau_{e, i})$. The target position of $S(t)$ at the current step is obtained with a small increment $\xi$ along the path, i.e., $\sigma_i(\tau_{e,i} + \xi)$, and $\tau_{e,i} + \xi$ saturates at one. The tracking error term is 
\begin{equation}
\label{path error}
    \mathbf{e}_S(t) = S(t) - \Sigma_S(\tau_e + \xi)
\end{equation}
where $\Sigma_S(\tau_e + \xi)$ returns the forward shifted projection vector $S
_\Sigma = [\sigma_1(\tau_{e,1} + \xi)\T, ..., \sigma_k(\tau_{e,k} + \xi)\T]\T$. A simple feedback controller for $S$ can be constructed as $\dot{S} = -\mathbf{K}_S \mathbf{e}_S$ where $\mathbf{K}_S $ is a positive definite gain matrix.

The developed methods are compatible with different deformation modeling techniques. Here we simply assume that the (approximate) deformation Jacobian $\mathbf{J}_d$, which relates end-effector velocity $\dot{\mathbf{r}}$ and $\dot{S}$ by $\dot{S} = \mathbf{J}_d \dot{\mathbf{r}}$, is accessible via means like Broyden's method \cite{J. Nocedal 2006}. Ideally, the tracking control law
\begin{equation}
\label{velocity no constraint}
    \dot{\mathbf{r}}_S = -\mathbf{J}_d^\dagger \mathbf{K}_S \mathbf{e}_S
\end{equation}
where $\square^\dagger$ denotes the Moore-Penrose pseudoinverse, can be utilized. In typical constraint-free deformation control studies, e.g., \cite{D. Navarro-Alarcon 2014 ijrr, D. Navarro-Alarcon 2016 TRO}, $S$ will eventually converge to $S_d$ to accomplish the task with a bounded error.
\begin{table}[t]
    \caption{Cost Functions of Constraints and Transposed Gradients}
    \label{tab: Cost functions and gradients}
    \centering
    \begin{tabular}{|c|c|c|c|}
    \hline
        {$c_i$} &${c_1}$  &$c_2$  &$c_3$ \\
        \hline
        \textbf{$f_{c_i}$}  
        &$\frac{k_{\mathcal{E}, 1}}{2 d^2(\mathcal{O, E})}$ 
        &$\frac{k_{\mathcal{E}, 2}}{2 d^2(\mathbf{r}, \mathcal{E})}$ 
        &$\frac{1}{2} \| \mathcal{H}(S) - \frac{\underline{\mathbf{h}} + \overline{\mathbf{h}}}{2} \|^2$ \\
        \hline
        \textbf{$\pdv{f_{c_i}}{\mathbf{r}}$} 
        &$\frac{-k_{\mathcal{E},1} (\mathbf{p}_o - \mathcal{E})\T \mathbf{J}_{d,i}}{d^4(\mathcal{O, E})}$  
        & $\frac{-k_{\mathcal{E},2}\mathbf{d}(\mathbf{r}, \mathcal{E})\T}{d^4(\mathbf{r}, \mathcal{E})}$  
        &$(\mathcal{H}(S) - \frac{\underline{\mathbf{h}} + \overline{\mathbf{h}}}{2})\T \mathbf{J}_{\mathcal{H }}$ \\
        \hline
    \end{tabular}
\end{table}

In the presence of constraints, the velocity command $\dot{\mathbf{r}}_{c_i}$ adjusting a single constraint $c_i$ is readily attainable by gradient descent of the cost function $f_{c_i}$.
As shown in Table \ref{tab: Cost functions and gradients}, $f_{c_i}$ is formally simple, e.g., repulsive potential for collision constraints and squared distance to the admissible range center. $f_{c_2}$ and $f_{c_3}$ are directly related to $\mathbf{r}$. But the gradient of $f_{c_1}$ w.r.t. $\mathbf{r}$ is not easily obtainable.
To execute the move induced by $\dot{\mathbf{r}}_{c_1}$, the motion transformation from $\mathbf{r}$ to $\mathbf{p}_o = \text{arg min} d(\mathbf{p}_i, \mathcal{E}), \mathbf{p}_i \in \mathcal{O}$ is needed. Nonetheless, it is unachievable in model-free approaches due to the uncertainty of $\mathbf{p}_o$. To address this, an approximation is used assuming $\mathbf{p}_o$ shares a similar motion transformation with its nearest feedback point. If $\mathbf{p}_{s_i}$ has the shortest inner-distance to $\mathbf{p}_o$ (shortest path length inside the DO shape silhouette \cite{H. Ling 2007}), its deformation Jacobian $\mathbf{J}_{d,i}$ is leveraged for motion command generation
\begin{equation}
    \dot{\mathbf{r}}_{c_1} =  k_{c_1} \mathbf{J}_{d,i}\T(\mathbf{p}_o - \mathcal{E})
\end{equation}
where $\mathbf{J}_{d,i}$ is the submatrix associated with $\mathbf{s}_i$ in $\mathbf{J}_d$. This approximation is assumed valid if the DO size is not too large. $f_{c_3}$ is in a quadratic form. In $\pdv{f_{c_3}}{\mathbf{r}}$, $\mathbf{J}_{\mathcal{H}} = \pdv{\mathcal{H}}{\mathbf{r}}$ is the deformation Jacobian of $\mathcal{H}(\cdot)$. 
\begin{figure}[t]
    \minipage{1 \columnwidth}
    \centering
    \includegraphics[width= 0.88 \columnwidth]{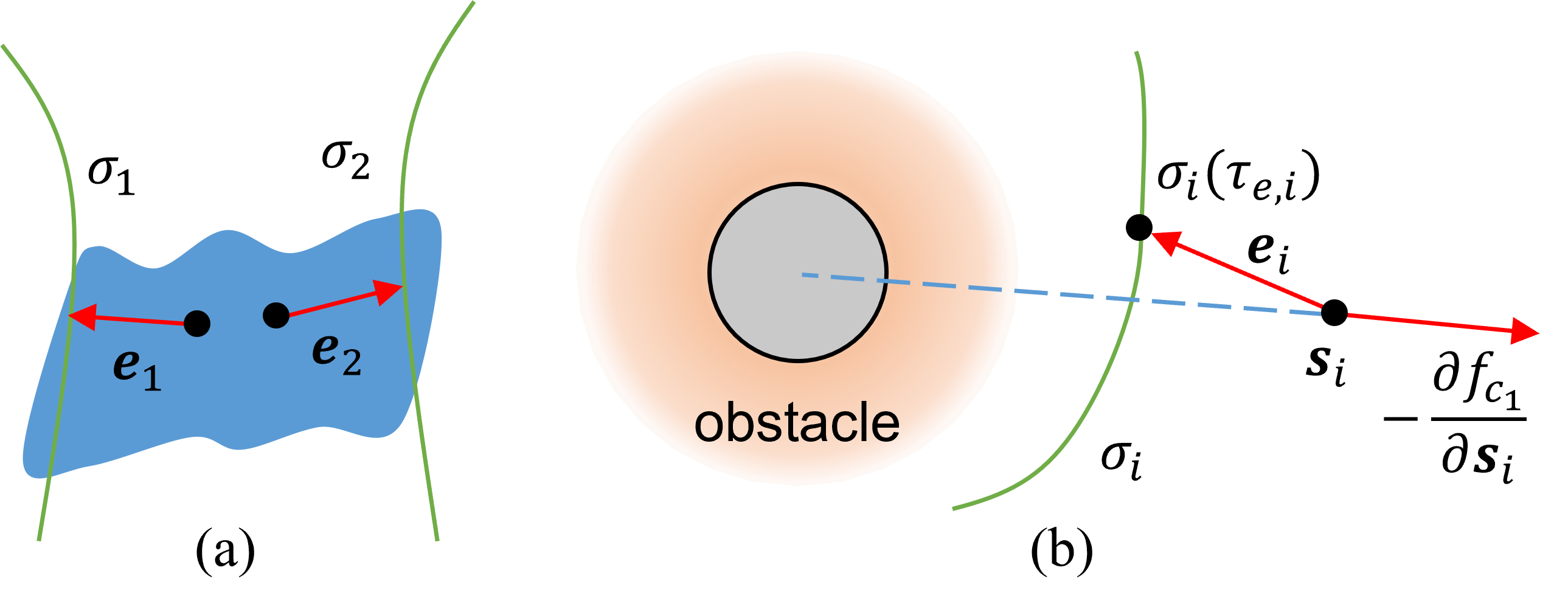}   
    \endminipage \hfill
    \caption{ (a) When path set tracking errors are conflicting, the generated robot motion can be very small to stick the task. (b) 
    Take the single point case as an example, the path tracking motion and the constraint motion can conflict each other easily.}
    \label{Conflict Motion Example}
\end{figure}

With multiple constraints, priority scheduling first satisfies hard constraints and then imposes secondary ones. Constraints are sequenced in a prioritized order: $c_2 \rightarrow c_1 \rightarrow c_3$, depending on the practical importance and potential severity of violation. The robot collision constraint $d(\mathbf{r}, \mathcal{E}) > 0$ is always enforced for safety. Constraints related to DOs rank lower. By iterative nullspace projection of velocity, the constrained motion velocity is obtained as
\begin{equation}
\label{velocity_eq_0}
    \dot{\mathbf{r}}_{c} = \dot{\mathbf{r}}_{c_2} + \mathbf{N}_{c_2} \dot{\mathbf{r}}_{c_1} + \mathbf{N}_{c_2} \mathbf{N}_{c_1} \dot{\mathbf{r}}_{c_3}
\end{equation}
where $\mathbf{N}_{c_i}$ is the nullspace projector to $\dot{\mathbf{r}}_{c_i}$. If the constraint is not active, $a_i = 0, \dot{\mathbf{r}}_{c_i} = \mathbf{0}, \mathbf{N}_{c_i} = \mathbf{I}$.

\subsection{Interleaving Path Set Tracking and Direct End-effector Path Tracking}
In practice, single-arm or dual-arm robotic systems are often used. It is hard to coordinate $\dot{\mathbf{r}}_S$ and $\dot{\mathbf{r}}_c$ due to the limited DOFs provided by the end-effectors. Notably, the manipulator is unable to precisely track the path set when the dimension of $S$ exceeds end-effector DOFs. Consider the Lyapunov function $\mathcal{Q} = \mathbf{e}_S^\mathsf{T} \mathbf{e}_S / 2$. Using the tracking controller in (\ref{velocity no constraint}), its time derivative yields
\begin{equation}
    \dot{\mathcal{Q}} = \mathbf{e}_S^\mathsf{T}(\dot{S} - \dot{S}_\Sigma) = -\mathbf{e}_S^\mathsf{T} \mathbf{K}_S \mathbf{J}_d \mathbf{J}_d^\dagger \mathbf{e}_S - \mathbf{e}_S^\mathsf{T} \dot{S}_\Sigma
\end{equation}
where $\dot{\mathbf{e}}_S = \dot{S} - \dot{S}_\Sigma$. $\mathbf{J}_d$ has more rows ($2K$) than columns ($2$ for a single arm, $4$ for dual arms) in most scenarios, resulting in an underactuated system. As such, $\mathbf{J}_d \mathbf{J}_d^\dagger$ is only positive semi-definite. In path set tracking, we can reasonably assume that $\mathbf{e}_S^\mathsf{T} \dot{S}_\Sigma = 0$ with a small increment $\xi$ in $\Sigma_S(\tau_e + \xi)$. When $S$ is near $S_d$, the forward tracking term on path saturates at $S_d$, i.e., $\dot{S}_\Sigma = 0$. These make $\dot{\mathcal{Q}} \leq 0$ and local minima possible to happen.

More importantly, the overall task execution can easily get stuck if $\dot{\mathbf{r}}_S$ and $\dot{\mathbf{r}}_c$ are simply switched with the constraint activation state. This is because tracking and constraint adjustment motions are often inconsistent or even conflicting as illustrated in Fig. \ref{Conflict Motion Example}(b). A high-level task conduction pipeline ensuring the task progress is thus required. For this, a scheme which interleaves path set tracking with constraints and direct end-effector control is introduced. Firstly, path set tracking coordination is used for a unified tracking target. Note that $\mathbf{e}_i$ terms may be conflicting and counteract each other, rendering $\dot{\mathbf{r}}_S$ quite small and $S_\Sigma$ almost intact to stick the tracking as Fig. \ref{Conflict Motion Example}(a). To alleviate this, a more coordinated path set tracking error is needed. Using the pivot path $\sigma_p$ as a unified tracking reference, the tracking error term is given as
\begin{equation}
    \mathbf{e}_S(t) = S(t) - \Sigma_S(\tau_{e,p} + \xi).
\end{equation}
In this way, tracked positions on all paths have the same path length parameter to decrease the probability of conflicting $\mathbf{e}_i$. Further, if a severe conflict is detected by calculating the inner product with $\mathbf{e}_p = \mathbf{s}_p - \sigma_p(\tau_{e, p} + \xi)$, the conflicting component in $\mathbf{e}_i$ is removed by orthogonal projection.

Next, to efficiently circumvent collision regions and proceed with the task, direct end-effector control is interleaved with path set tracking. The manipulation system tends to enter the same collision region repeatedly because of the task setup similarity after conducting $\dot{\mathbf{r}}_{c}$, making itself sluggish. An effective option is to explicitly control the robot to get out of the collision region by tracking a locally planned end-effector path $\sigma_{EE}$. $\sigma_{EE}$ is planned locally with an appropriate end-effector target position $\mathbf{r}_{d}$ that helps the DO pass the collision region. 
Feasibility of $\sigma_{EE}$ is evaluated by the homotopy property of the path set $\{\sigma_p, \sigma_{EE}\}$. When tracking $\sigma_{EE}$, DO-related constraints, i.e., $c_1, c_3$, are not enforced as $\dot{\mathbf{r}}$ is directly controlled. Once the plant leaves the collision region, tracking of the path set $\Sigma_S$ will be recovered.

\section{Experimental Results}
\label{Experiment Section}
We have implemented and tested the proposed path set planning and tracking methods for DOM tasks in different constrained setups. Both the implementation of path set planning algorithms and physical robot experiments with constraints are displayed in this section.

\subsection{Implementation Details}
The camera image space is the default configuration space in path set planning. The upstream recognition of obstacles and DOs is assumed available and obstacles are processed as polygons. Feedback points and derived deformation features can be specified manually. For single path planning, RRT$^*$ is amended with the composite cost in (\ref{Cost Function for Passage Passing}) and built upon the kd-tree. The cost and the minimum passage width passed through by the optimal path from the start node to each path node are stored as node attributes. The passage list $P_{\sigma_p}(1)$ is retrieved via a one-way traversal of the planned $\sigma_p$. Then the forward transferred path set and chords are obtained for repositioning $\sigma_p$ to $\sigma_p^*$. If infeasible transferred path segments exist, deformable path transfer is subsequently performed. 
In robot experiments, the 7-DOF Flexiv Rizon manipulators with a clamp-type end-effector are used.
Grasps are preset properly but can also be founded actively with actuated grippers using our method in \cite{J. Huang 2023}. All computations are performed on a laptop with Ubuntu 18.04, Intel Core i7-7700 CPU$@\SI{2.8}{\GHz} \times 8$, and 16 GB of RAM. The visual module (image perception, segmentation, and point tracking using Lucas-Kanade method) and online numerical $\mathbf{J}_d$ update run at the camera frame rate of $\SI{30}{\hertz}$. The end-effector velocity command is generated at a lower frequency of $\SI{25}{\hertz}$. Raw sampled paths are smoothed and densified in quadratic B-spline format for tracking. The whole architecture is implemented in the robot operating system (ROS) to simplify the communication and synchronization between the sensing and execution units.
\iftrue
\begin{figure}[t]
    \minipage{1 \columnwidth}
    \centering
    \subfigure[] {
    \includegraphics[width= 0.48 \columnwidth]{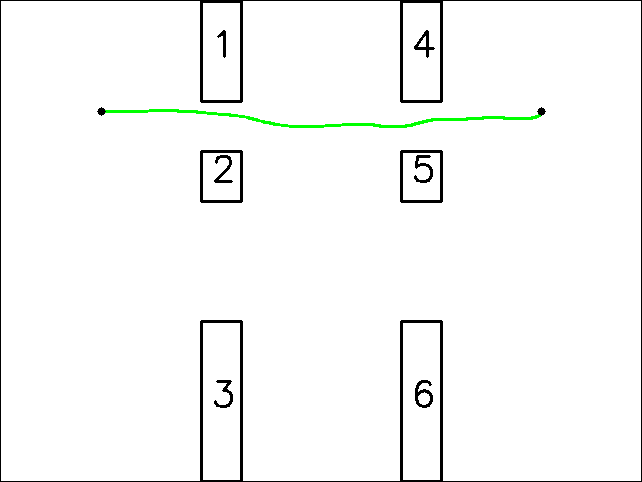}}
    \subfigure[] {
    \includegraphics[width= 0.48 \columnwidth]{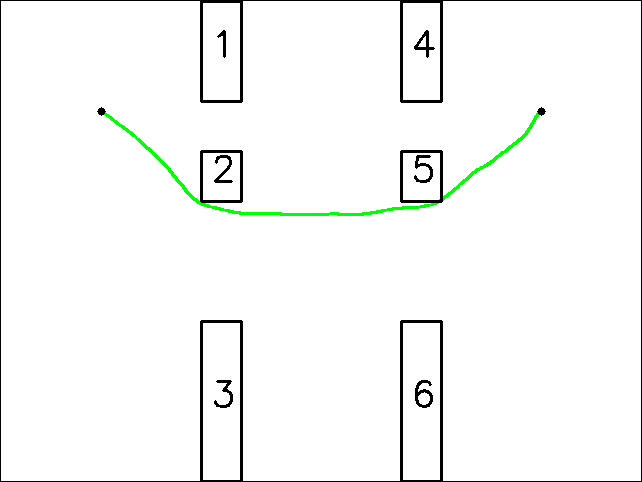}}
    \endminipage \hfill
    \minipage{1 \columnwidth}
    \centering
    \subfigure[] {
    \includegraphics[width= 0.48 \columnwidth]{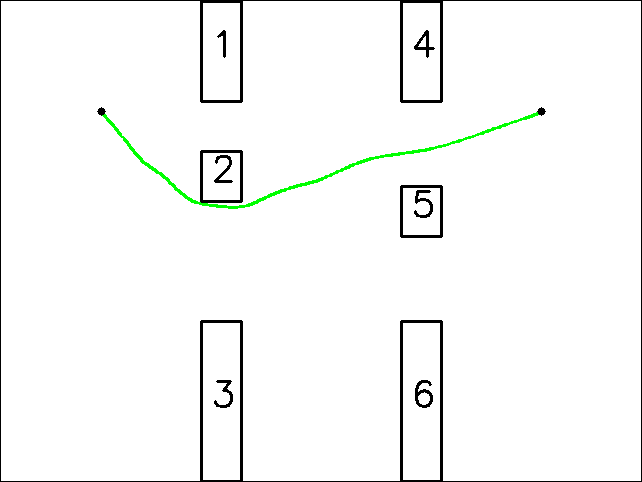}}
    \subfigure[] {
    \includegraphics[width= 0.48 \columnwidth]{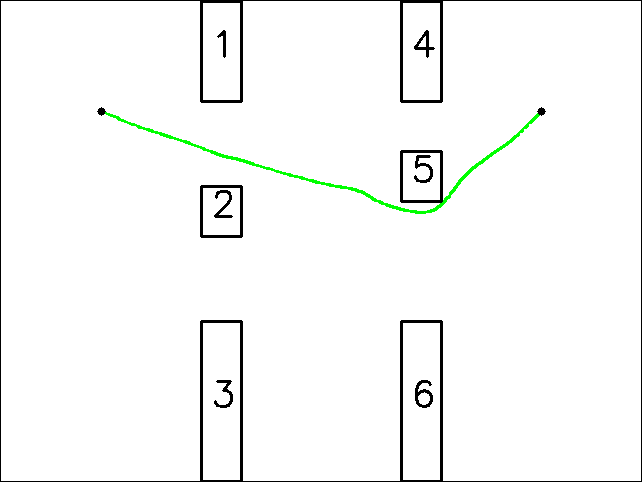}}
    \endminipage    
    \caption{Path planning comparison between the path length cost and the composite cost in RRT$^*$. (a) Path length cost. (b)-(d) Composite cost (\ref{Cost Function for Passage Passing}).}
    \label{RRTStar Planning Comparison of Different Costs}
    \end{figure}
\fi

\subsection{Path Set Generation Results}
\begin{figure}[t]
    \minipage{1 \columnwidth}
    \centering
    \subfigure[] {
    \includegraphics[width= 0.48 \columnwidth]{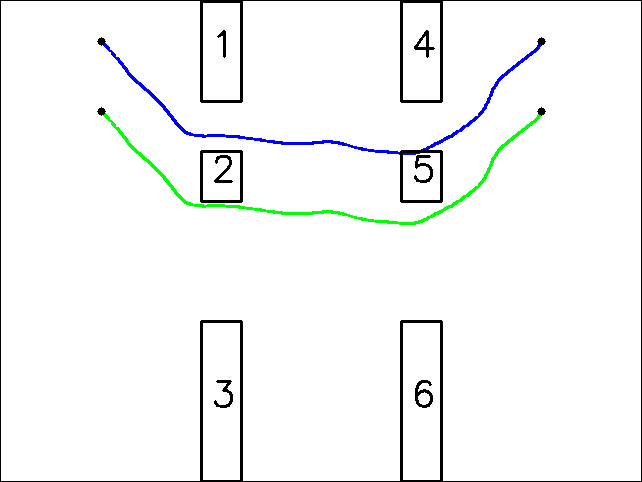}}
    \subfigure[] {
    \includegraphics[width= 0.48 \columnwidth]{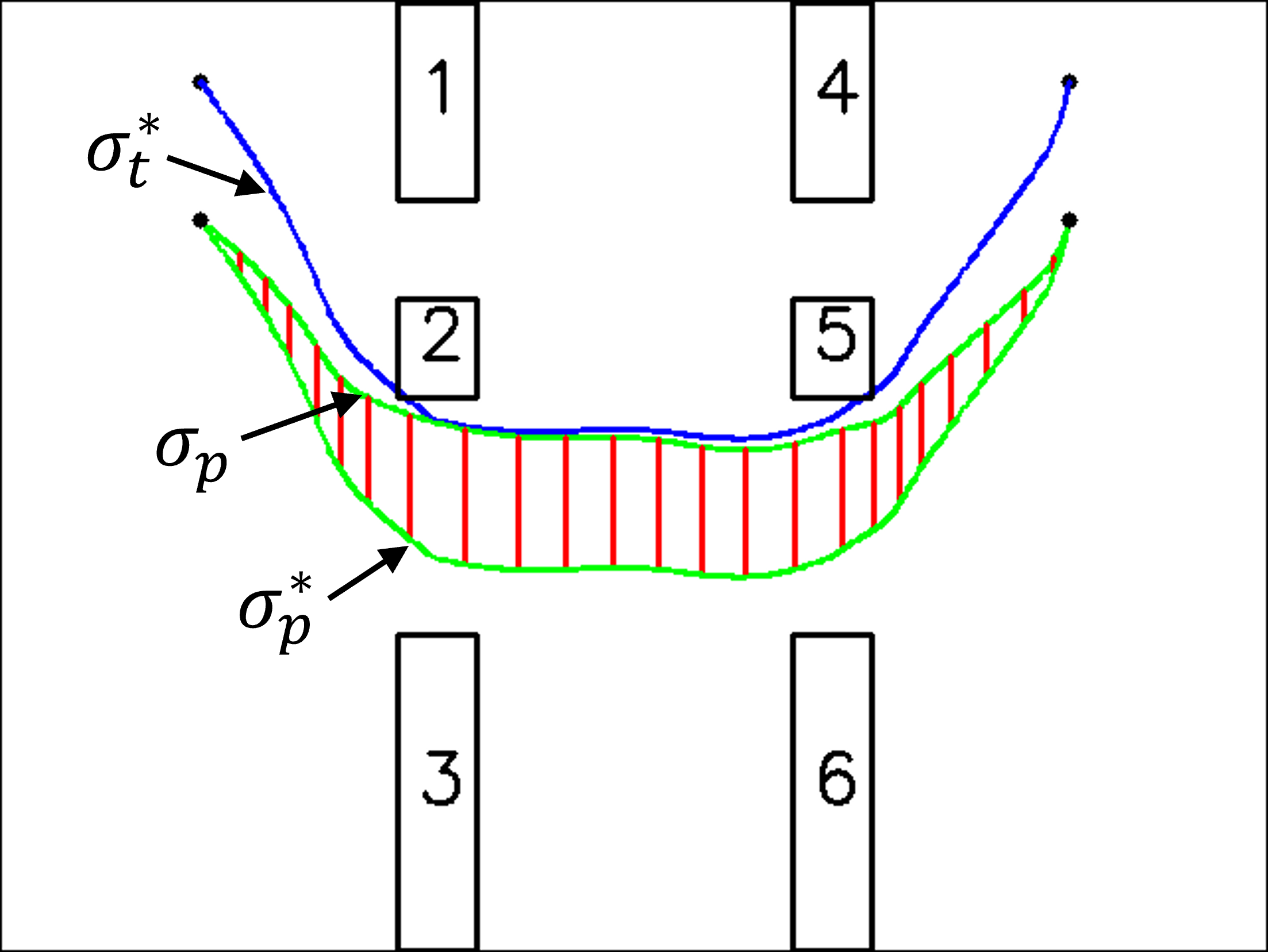}}
    \minipage{1 \columnwidth}
    \centering
    \subfigure[] {
    \includegraphics[width= 0.48 \columnwidth]{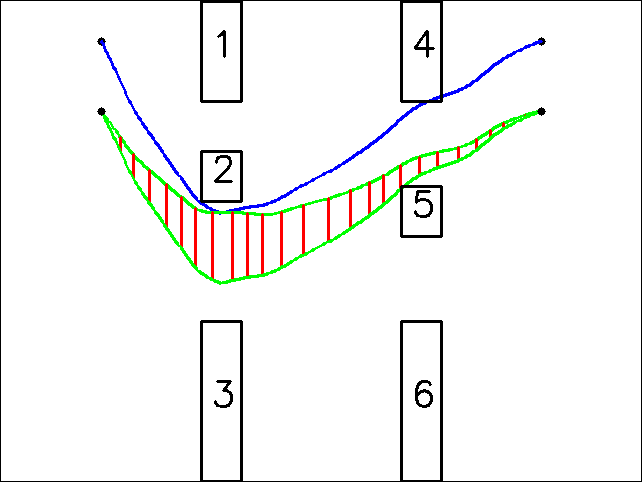}}
    \subfigure[] {
    \includegraphics[width= 0.48 \columnwidth]{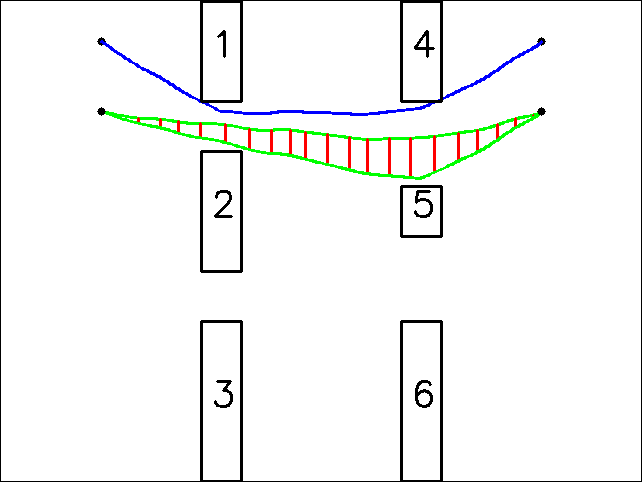}}
    \endminipage        
    \endminipage \hfill
    \caption{Path set generation. Pivot paths are in green. Transferred paths are in blue. Red segments show the repositioning mapping between $\sigma_p$ and $\sigma_p^*$. (a) Direct forward path transfer. (b)-(d) Path transfer in constrained conditions.}
    \label{Path Repositioning}
    \end{figure}

\begin{figure}[t]
    \minipage{1 \columnwidth}
    \centering
    \centering
    \subfigure[] {
    \includegraphics[width= 0.48 \columnwidth]{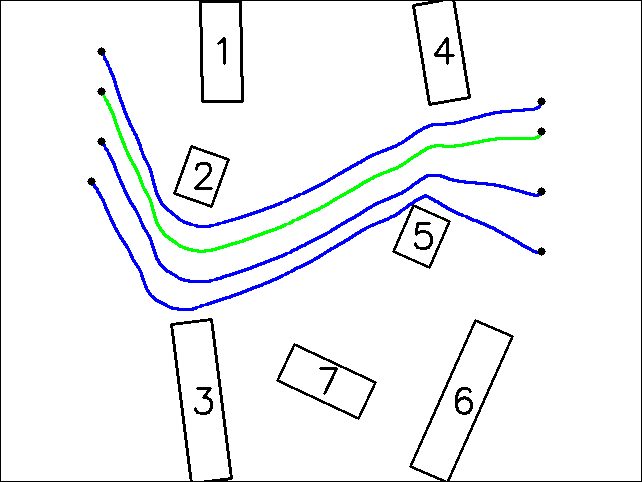}}
    \subfigure[] {
    \includegraphics[width= 0.48 \columnwidth]{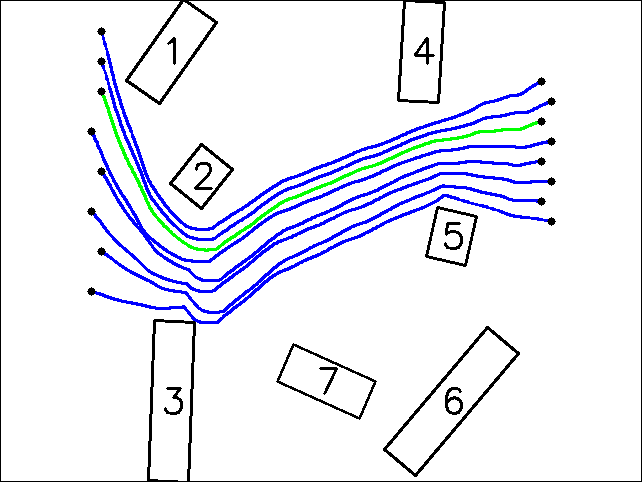}}
    \endminipage \hfill
    \caption{Path set generation with multiple points. (a) Four feedback points ($K = 4$). (b) Eight feedback points ($K = 8$).}
    \label{Path Repositioning More Points}
    \end{figure}

We first show the path set planning implementation results. Examples are showcased from Fig. \ref{RRTStar Planning Comparison of Different Costs} to Fig. \ref{Path Repositioning More Points}. In single pivot path planning, the shortest path length cost is compared with. As expected, for the same planning problem, the planner using path length cost may result in limited workspace along the path, e.g., Fig. \ref{RRTStar Planning Comparison of Different Costs}(a). Using the composite cost in (\ref{Cost Function for Passage Passing}) which also takes into account traversed passages, workspace, and path length are traded off to attain the path. In \ref{RRTStar Planning Comparison of Different Costs}(b)-(d), there are four vertical valid passages in the environment. When the passage configuration changes, the planner returns different paths. Specifically, the displacement from the start to the target is horizontal. The planned paths choose the wider passage between vertically aligned passages ($(\mathcal{E}_1, \mathcal{E}_2)$ vs. $(\mathcal{E}_2, \mathcal{E}_3)$,  $(\mathcal{E}_4, \mathcal{E}_5)$ vs. $(\mathcal{E}_5, \mathcal{E}_6)$). If passages are of similar widths, such as $(\mathcal{E}_4, \mathcal{E}_5)$ and $(\mathcal{E}_5, \mathcal{E}_6)$ in Fig. \ref{RRTStar Planning Comparison of Different Costs}(c), the one leading to a shorter path length is selected.
Such passage encoding and selection capacities in pivot path planning are important in restrictive environments where the transfer assumptions fail. More workspace for manipulation will be provided by the path set transferred from the pivot path.
\begin{figure*}[t]
    \minipage{2 \columnwidth}
    \centering
    \includegraphics[width= 1 \columnwidth]{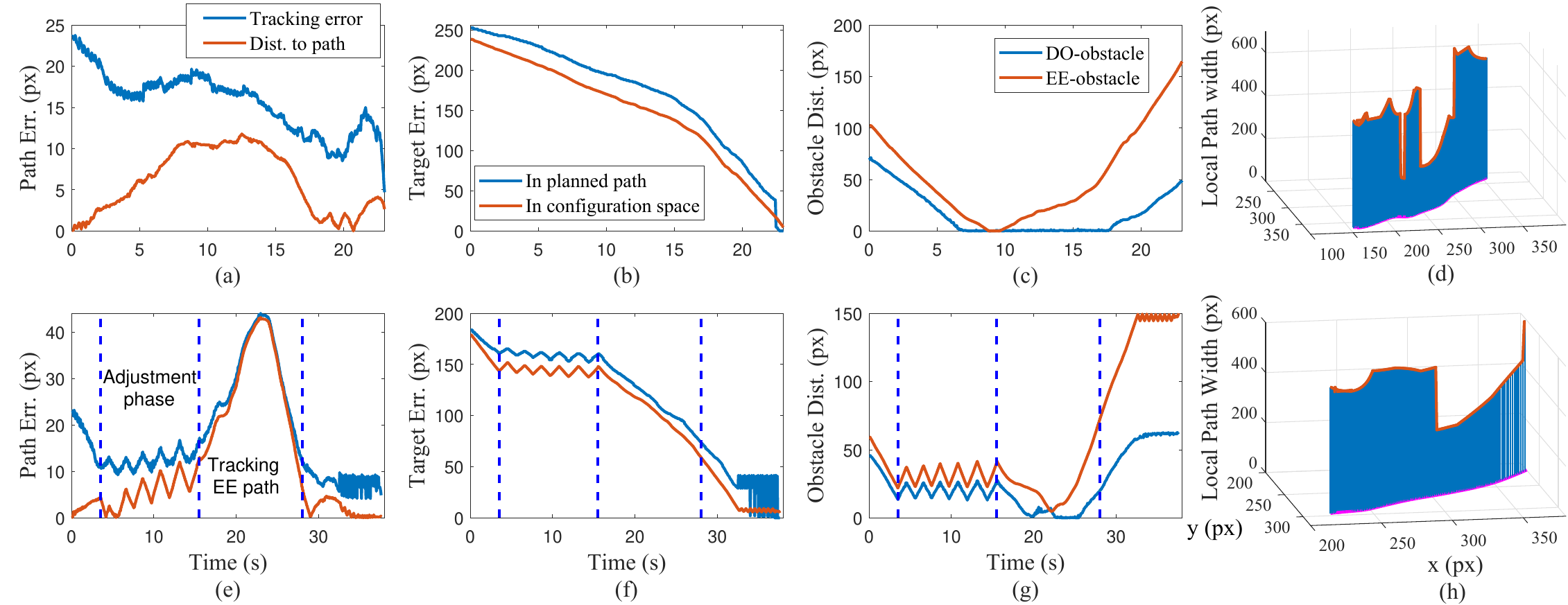}
    \endminipage \hfill
    \caption{(a) Tracking performance: tracking error ($\|\mathbf{s}_1 - \sigma_1(\tau_{e, 1} + \xi)\|_2$), distance to path ($\|\mathbf{s}_1 - \sigma_1\|_2$). (b) Target error: in planned path (path length from $\sigma_1(\tau_{e, 1} + \xi)$ to $\sigma_1(1)$), in configuration space ($\|\mathbf{s}_1 - \mathbf{s}_{d,1}\|_2$). (c) End-effector (EE) and DO's distances to the obstacle. (d) Local path width along $\sigma_1$. Metrics in the constrained experiment are shown in (e)-(h). The processes of constraint adjustment and tracking $\sigma_{EE}$ are depicted by blue dashed lines.}
    \label{Single point data united}
\end{figure*}

Based on passage-aware pivot path planning in constrained environments, path set generation is conducted. Fig. \ref{Path Repositioning} displays the benchmark case of path set generation with only two feedback points. The pivot is arbitrarily appointed in this part. Despite wide passages exist in Fig. \ref{Path Repositioning}(a), the transfer assumptions are not met since no feasible path can be planned in init$_{\delta_p}(\mathcal{X}_{free})$ ($\delta_p = 70, \| (\mathcal{E}_2, \mathcal{E}_3) \|_2 = \| (\mathcal{E}_5, \mathcal{E}_6) \|_2 = 120 < 2\delta_p$). This also shows that the transfer assumptions are very restrictive in practice. The pivot path $\sigma_p$ is then planned with a small $\delta < \delta_p$ in init$_{\delta}(\mathcal{X}_{free})$. $\delta$ can be very small in order to ensure $\sigma_p$ is found since the original transfer assumption $\sigma_p \in$ init$_{\delta_p}(\mathcal{X}_{free})$ will not be used in following path set generation. The directly forward transferred path $\sigma_t$ is infeasible due to collision. Meanwhile, the strong homotopic-like property of the resulting path set does not hold. This entails repositioning of $\sigma_p$ and deformable path transfer proposed for general constrained conditions. As illustrated in Fig. \ref{Path Repositioning}(b)-(d), the repositioned pivot path $\sigma_p^*$ is shifted to create more free space for following path transfer. In Fig. \ref{Path Repositioning}(b) and (c), $\sigma^*_p$ directly leads to feasible transferred paths as the passed passages have widths larger than $\delta_p$. But transferred paths from $\sigma_p^*$ are not guaranteed to be feasible. In Fig. \ref{Path Repositioning}(d), the transferred path translated from the repositioned $\sigma^*_p$ collides with obstacles due to the narrow $(\mathcal{E}_1, \mathcal{E}_2)$. It is thus further deformed to be feasible. The path set generation pipeline is not limited by the feedback number and pivot selection. For instance, Fig. \ref{Path Repositioning More Points} shows examples of four and eight feedback points in different constrained setups with arbitrarily picked pivots.
\begin{figure}[t]
    \minipage{1 \columnwidth}
    \centering
    \subfigure[] {
    \includegraphics[width= 0.45 \columnwidth]{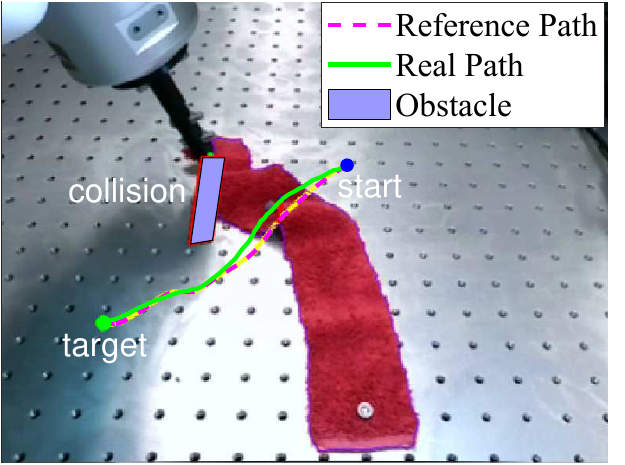}}
    \subfigure[] {
    \includegraphics[width= 0.45 \columnwidth]{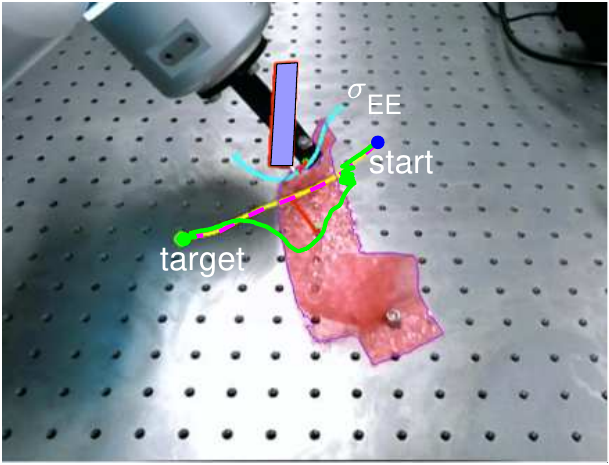}}
    \endminipage  \hfill
    \caption{(a) Planned and real paths of feedback point $\mathbf{s}_1$ on a fabric sheet in the unconstrained experiment. A collision between the sheet and the obstacle occurs. (b) Constrained experiment. 
    }
    \label{Single point path}
    \end{figure}

\subsection{Single Point Manipulation}
Single point manipulation tasks are performed as the benchmark tests. Both the feature vector $\mathbf{y}$ and the feedback vector $S$ only contain a feedback point $\mathbf{s}_1$. The robot needs to manipulate the DO so that $\mathbf{s}_1$ can reach $\mathbf{s}_{d,1}$. For comparison, two sets of tests are performed with similar setups. The unconstrained tests only enable path set tracking. The constrained tests impose collision constraints $c_1, c_2$. 
\begin{figure}[t]
    \minipage{1 \columnwidth}
    \centering
    \subfigure[] {
    \includegraphics[width= 0.45 \columnwidth]{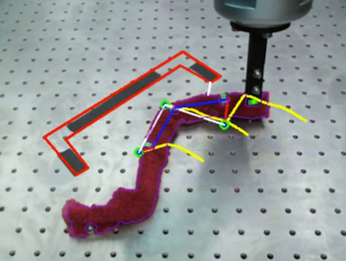}}
    \subfigure[] {
    \includegraphics[width= 0.49 \columnwidth]{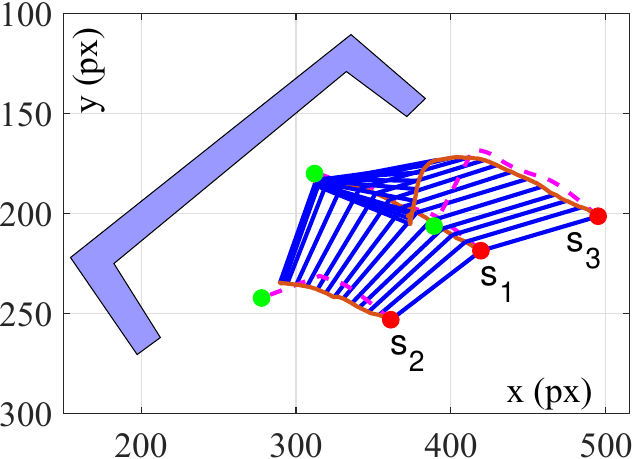}}
    \endminipage  \hfill
    \caption{(a) Bending-move task snapshot.
    (b) Motion process of the angle feature. Red and green points are initial and target feedback points, respectively. Blue lines depict angle sides.}
    \label{Angle simulation and path}
\end{figure}

\subsubsection{Unconstrained Single Point Manipulation} 
In this experiment, a fabric sheet anchored on the operation table is manipulated. In the tracking controller, $\mathbf{K}_S = \mathbf{I}$, the forward shift in the tracking error term $\xi = 0.05$. Fig. \ref{Single point path}(a) shows the planned reference path and the real path of $\mathbf{s}_1$. The associated data are demonstrated in Fig. \ref{Single point data united}(a)-(c). Fairly good tracking performance is achieved with the maximum deviation from the reference path below $10$ px in Fig. \ref{Single point data united}(a). In Fig. \ref{Single point data united}(b), the target error measured in the planned path refers to the path length between $\sigma_1(\tau_{e,1} + \xi)$ and $\sigma_1(1) = \mathbf{s}_{d,1}$ to reflect the tracking progress. The target error in configuration space is $\|\mathbf{s}_1 - \mathbf{s}_{d,1}\|_2$. Both error measurements exhibit a similar temporal evolution tendency to decline to zero monotonically. However, as illustrated in Fig. \ref{Single point path}(a) and Fig. \ref{Single point data united}(c), both the DO and end-effector collide with the obstacle while their distances to the obstacle decrease to zero. Moreover, a long duration ($\sim$ $6.5-\SI{17.5}{\second}$) of DO-obstacle collision is observed. The end-effector collides with the obstacle around $8.5-\SI{10}{\second}$. For this reason, the task would fail if these constraints are strictly imposed as in practice, which necessitates proper constraint handling. Fig. \ref{Single point data united}(d) illustrates the local path width of $\sigma_1$ calculated using centered numerical differentiation. The field of view ($640 \times 480$ px) is set as the outer boundary of $\mathcal{X}_{free}$. The short local minimum indicates the position where a narrow passage region is encountered.

\subsubsection{Constrained Single Point Manipulation} 
A plastic sheet is used in this experiment to avoid significant sheet folding and consequent occlusion of $\mathbf{s}_1$. All experimental settings remain unchanged except that collision constraints $c_1$ and $c_2$ are now imposed, for which distances $d(\mathcal{O}, \mathcal{E})$ and $d(\mathbf{r}, \mathcal{E})$ are extracted in real time to check violation.

At the beginning, path tracking proceeds normally. $d(\mathcal{O}, \mathcal{E})$ approaches the collision threshold ($15$ px) at around $\SI{3.5}{\second}$ for the first time. The following adjustment leads to repetitive motions in $3.5-\SI{15.5}{\second}$ as demonstrated by curve zigzags in Fig. \ref{Single point data united}(e)-(g). On $\mathbf{s}_1$ path in Fig. \ref{Single point path}(b), a chattering section is caused which restricts $\mathbf{s}_1$. As a result, a task-stuck state is recognized. A conservative criterion is adopted here that only after the violation takes place consecutively over five times, the current situation would be regarded as being stuck. To leave this region, i.e., bypass the obstacle, a local end-effector path $\sigma_{EE}$ depicted by the cyan path in Fig. \ref{Single point path}(b) is planned and tracked during $15.5-\SI{28}{\second}$. When $\dot{\mathbf{r}}$ is controlled to track $\sigma_{EE}$, the path tracking error of $\mathbf{s}_1$ will not necessarily decline. In this experiment, a significant deviation from the reference path in Fig. \ref{Single point path}(b) can be observed. Constraints related to the DO are relaxed in tracking $\sigma_{EE}$. Thus temporary DO-obstacle collision is allowed as what happens during $22.5-\SI{25.5}{\second}$ in Fig. \ref{Single point data united}(g) near the local path width minimum in Fig. \ref{Single point data united}(h). After passing the obstacle by following $\sigma_{EE}$, tracking $\sigma_1$ is recovered and interleaved timely. The small chattering at the final stage in Fig. \ref{Single point data united}(e),(f) is caused by frequent switches of the shifted projection term $S_{\Sigma}$, which can be resolved by higher path resolution and more robust algorithmic settings.

\subsection{Multiple-Point Feature Manipulation}
The second series of experiments concentrate on features composed of multiple feedback points that better characterize DO deformation. In particular, we consider the bending-move task described by feature $\mathbf{y} = [\mathbf{s}_1^\mathsf{T} \, \alpha]^\mathsf{T} \in \mathbb{R}^3$, $S = [\mathbf{s}_1^\mathsf{T} \, \mathbf{s}_2^\mathsf{T} \, \mathbf{s}_3^\mathsf{T}]^\mathsf{T}$ and $\alpha = \acos (\mathbf{v}_{12} \cdot \mathbf{v}_{13} \, / \, (\|\mathbf{v}_{12}\|_2\,\|\mathbf{v}_{13}\|_2))$. $\mathbf{v}_{12} = \mathbf{s}_2 - \mathbf{s}_1$, $\mathbf{v}_{13} = \mathbf{s}_3 - \mathbf{s}_1$ are two side vectors. Physically, the task intends to bend the DO to a given angle meanwhile move the angle vertex $\mathbf{s}_1$ to its target region. Both $\mathbf{s}_{d,1}$ and $\alpha_d$ are specified when defining the task. $\mathbf{y}$ is thus complete and $\mathbf{s}_1$ plays the role of the pivot in planning. A symbolic box obstacle of $\sqcup$ shape is placed and the DO is made of a folded fabric sheet. Initially, the DO is outside of the obstacle and unbent. Once $S_0$ and $\mathbf{y}_d$ are provided, the first step is to determine the target feedback vector $S_d$ by solving (\ref{sd optimization}). In this experiment, the DO shape constraint in the target determination is imposed on the angle side length change ratio as $\Delta\|\mathbf{v}_{1i}\|_2 / \|\mathbf{v}_{1i,0}\|_2 \leq \SI{2}{\percent}, i = 2, 3$. Since the transfer assumptions hold in this setup, the basic procedure for path set generation is utilized.
\begin{figure}[t]
    \minipage{1 \columnwidth}
    \centering
    \includegraphics[width= 0.98 \columnwidth]{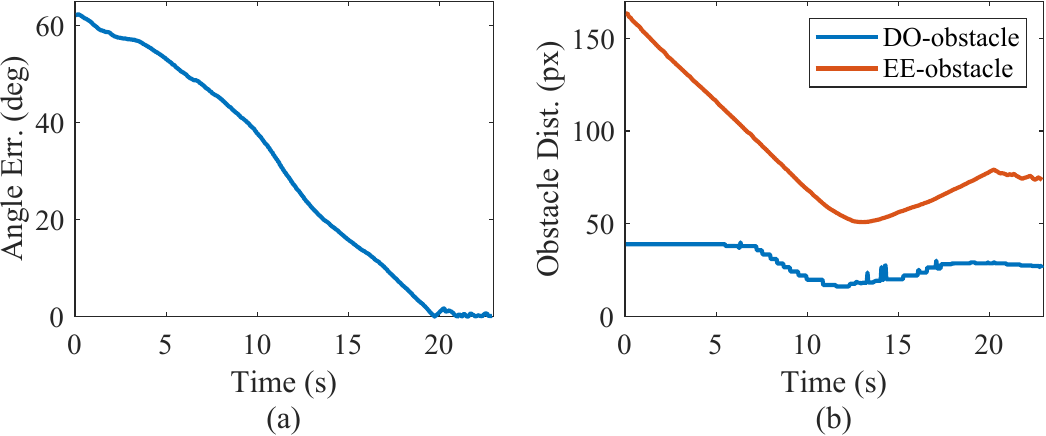}
    \endminipage \hfill
    \caption{(a) Error evolution of the feature angle $|\alpha - \alpha_d|$. (b) DO-obstacle and EE-obstacle distances.}
    \label{Angle manipulation data angle distance}
\end{figure}
\begin{figure}[t]
    \minipage{1 \columnwidth}
    \centering
    \includegraphics[width= 0.98 \columnwidth]{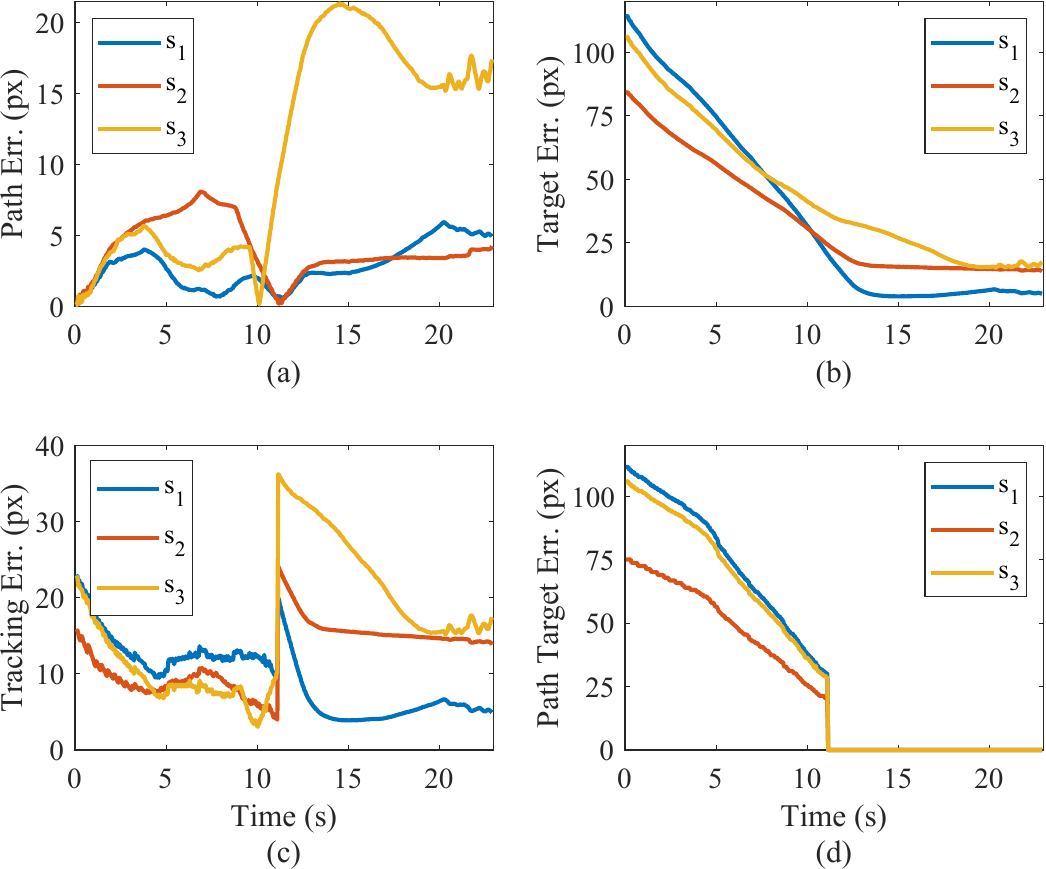}
    \endminipage \hfill
    \caption{(a) $\| \mathbf{s}_i - \sigma_i \|_2$. (b) $\| \mathbf{s}_i - \mathbf{s}_{d,i} \|_2$. (c) $\| \mathbf{s}_i - \sigma_i(\tau_{e,i} + \xi) \|_2$. (d) Path length on $\sigma_i$ from $\sigma_{i}(\tau_{e,i} + \xi)$ to $\sigma_i(1) = \mathbf{s}_{d,i}$.}
    \label{Angle manipulation data}
\end{figure}

Fig. \ref{Angle simulation and path}(b) illustrates the motion processes of each feedback point and the constructed feature. Experimental data are depicted in Fig. \ref{Angle manipulation data angle distance} and Fig. \ref{Angle manipulation data}. Because of the truncation and concatenation of transferred paths, the resulting paths may not be optimal or smooth compared to directly planned paths. Taking $\mathbf{s}_3$ as an example, there exists a sharp turn at the truncation position in $\sigma_3$. But the feasibility of the holistic path set is ensured despite non-optimal transferred paths. For path tracking precision, unsurprisingly, the pivot path $\sigma_1$ is tracked more accurately given its tracking priority, while coarser tracking is exhibited by $\mathbf{s}_2, \mathbf{s}_3$. 
In tracking, the tracked position on each path is coordinated and synchronized in accordance with the pivot tracking progress to avoid conflicting tracking terms. The components in tracking errors $\mathbf{e}_2, \mathbf{s}_3$ contradictory to the pivot tracking error $\mathbf{e}_1$ are removed if a serve conflict is detected, making precise path tracking harder to achieve.  In consequence, when $\mathbf{y}_d$ is reached with the preset error threshold ($5$ px for $\|\mathbf{s}_1 - \mathbf{s}_{d,1}\|_2$, $\SI{5}{\degree}$ for $|\alpha - \alpha_d|$), $\mathbf{s}_2$ and $\mathbf{s}_3$ are not necessarily located in their reference target regions in $S_d$. This is also shown in Fig. \ref{Angle manipulation data}(b), where the static errors of $\|\mathbf{s}_i - \mathbf{s}_{d, i}\|_2, i = 2, 3$ are of notably greater values than the pivot error $\|\mathbf{s}_1 - \mathbf{s}_{d,1}\|_2$.

To robustly reach the target, when feedback points approach their targets, their tracked positions on paths will saturate at the path ends, i.e., $\sigma_i(\tau_{e, i} + \xi) = \sigma_i(1) = \mathbf{s}_{d, i}$. This transfer occurs when the pivot's absolute tracking error becomes sufficiently small. At around $\SI{11}{\second}$, all path tracking errors decrease to zero sharply in Fig. \ref{Angle manipulation data}(d) because the pivot $\mathbf{s}_1$ enters the preset neighborhood of $\mathbf{s}_{d,1}$ ($\| \mathbf{s}_1 - \mathbf{s}_{d, 1}\|_2 < 20$ px) and its tracking target is fixed at $\mathbf{s}_{d, 1}$ ever after. This tracking target transfer causes abrupt increases of all tracking errors in Fig. \ref{Angle manipulation data}(c) since $S_\Sigma$ jumps to $S_d$ immediately. Overall, path set tracking is first conducted to complete gross motions until the pivot is close to its target. Then feedback points' tracking targets are fixed at their desired positions for finer modification until $\mathbf{y}_d$ is achieved. When $\mathbf{s}_1$ gets close to $\mathbf{s}_{d,1}$, the feature angle $\alpha$ still presents a large error. As shown by the target errors in Fig. \ref{Angle manipulation data}(b) and the feature in Fig. \ref{Angle simulation and path}(b), manipulation at the final stage mainly bends the object so that $\mathbf{s}_3$ moves to an appropriate position to reach $\alpha_d  = \SI{100}{\degree}$. In Fig. \ref{Angle manipulation data angle distance}(b), no collision is detected. The shape constraint is also obeyed. The multimedia material of more trials in various setups can be found in the supplementary video file.
\begin{figure}[t]
    \minipage{1 \columnwidth}
    \centering
    \subfigure[] {
    \includegraphics[width= 0.465 \columnwidth]{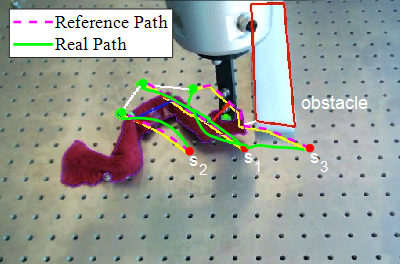}}
    \subfigure[] {
    \includegraphics[width= 0.465 \columnwidth]{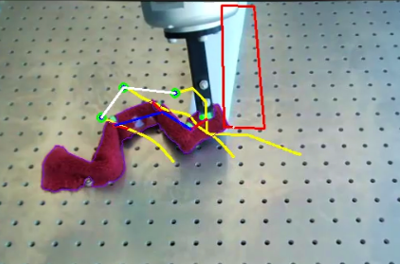}}
    \endminipage  \hfill
    \caption{A foam block is placed close to the folded fabric strip in comparison experiments. (a) Passing the foam without collision in the path set tracking approach. (b) Collision in the pure control method.}
    \label{Angle Comparison Setup}
    \end{figure}
\begin{figure}[t]
    \minipage{1 \columnwidth}
    \centering
    \includegraphics[width= 1 \columnwidth]{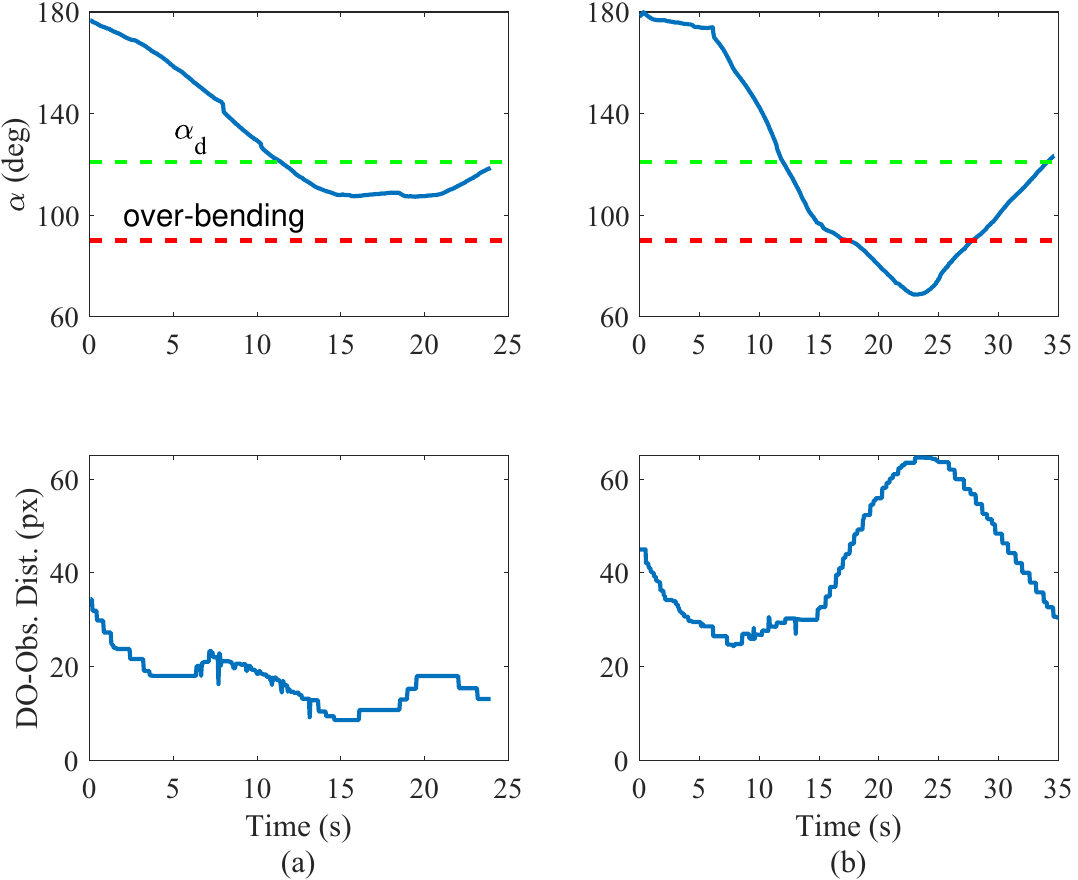}
    \endminipage \hfill
    \caption{Bending angle and DO-obstacle distance in comparison: (a) path set planning and tracking control, (b) pure deformation control.}
    \label{Angle Comparison setup1 unified}
\end{figure}

\subsection{Comparison With Pure Deformation Control}
The path set tracking approach and pure deformation control in DOM are compared in typical constrained conditions. The following classical control method is utilized
\begin{equation}
\label{Pure error-driven control}
    \dot{\mathbf{r}} = -\mathbf{J}_{\mathcal{F}}^\dagger \mathbf{K}_{\mathcal{F}} \mathbf{e}_y
\end{equation}
where $\mathbf{K}_{\mathcal{F}}$ and $\mathbf{e}_y$ are the gain and feature error, respectively. The motion-to-feature Jacobian is $\mathbf{J}_{\mathcal{F}} = \mathbf{J}_y \mathbf{J}_d$ and the feature Jacobian $\mathbf{J}_y = \pdv{\mathbf{y}}{S}$.
The bending-move task remains unchanged but an obstacle (a foam block) is placed near the fabric strip to create a confined environment for the motion of the manipulated sheet end (see Fig. \ref{Angle Comparison Setup}). The initial and target feature configurations are appointed as the same in the two methods. Two metrics are monitored for comparison purpose. One is the obstacle avoidance measured by the distance $d(\mathcal{O}, \mathcal{E})$. The other is the angle value $\alpha$, which is also the shape constraint quantity to reflect whether over-bending occurs. Specifically, the sheet is straight at rest and $\alpha$ is around $\SI{180}{\degree}$. The target bending angle is $\alpha_d = \SI{121}{\degree}$. If the bending angle falls below $\SI{90}{\degree}$, it will be regarded as over-bending. The confined setup of Fig. \ref{Angle Comparison Setup} makes the transfer assumption 1 fail to hold because not all feedback points are in int$_{\delta_p}(\mathcal{X}_{free})$ in (\ref{delta value}). Since there is only one obstacle, a smaller $\delta_p$ is used when planning the pivot path. The collision portion on $\sigma_{t,3}$ is then guided by an inflated obstacle polygon, yielding a path segment enclosing the obstacle in Fig. \ref{Angle Comparison Setup}.

Fig. \ref{Angle Comparison setup1 unified}(a) showcases the metrics in the path set tracking approach. No DO-obstacle collision occurs since the distance remains greater than zero. Meanwhile, no significant over-bending happens and the minimum $\alpha$ is around $\SI{110}{\degree}$, implying that the target is reached safely without extreme sheet bending. The metrics in the experiment utilizing the feedback controller in (\ref{Pure error-driven control}) are depicted in Fig. \ref{Angle Comparison setup1 unified}(b). There is also no collision detected. But the minimum bending angle falls to around $\SI{70}{\degree}$. Furthermore, $\alpha$ is smaller than the over-bending threshold for over $\SI{10}{\second}$. When the foam block is placed closer, the collision between the sheet and obstacle is observed in deformation control as shown in Fig. \ref{Angle Comparison Setup}(b). In pure control methods, constraint violation is often caused because constraints in the task cannot be well formulated and managed. The states of DOs and robots are thus partially indeterminate and undesirable situations may occur. Conversely, such occasions are effectively mitigated with a coordinated path set as the motion and path references in the path set tracking approach.
\begin{figure}[t]
    \minipage{1 \columnwidth}
    \centering
    \subfigure[] {
    \includegraphics[width= 0.465 \columnwidth]{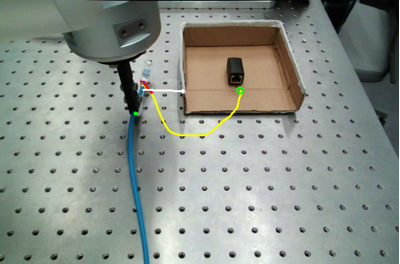}}
    \subfigure[] {
    \includegraphics[width= 0.465 \columnwidth]{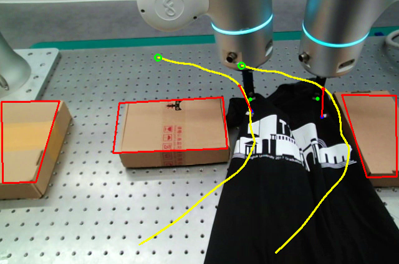}}
    \endminipage  \hfill
    \caption{Insertion and narrow passage passing tasks in application case study. (a) Insert the Ethernet cable to the port region. (b) Lead the clothes to pass the chosen narrow passage.}
    \label{Case Study Robot View}
    \end{figure}

\subsection{Application Case Study}
\begin{figure*}[t]
    \minipage{2 \columnwidth}
    \centering
    \subfigure[] {
    \includegraphics[width= 0.23 \columnwidth]{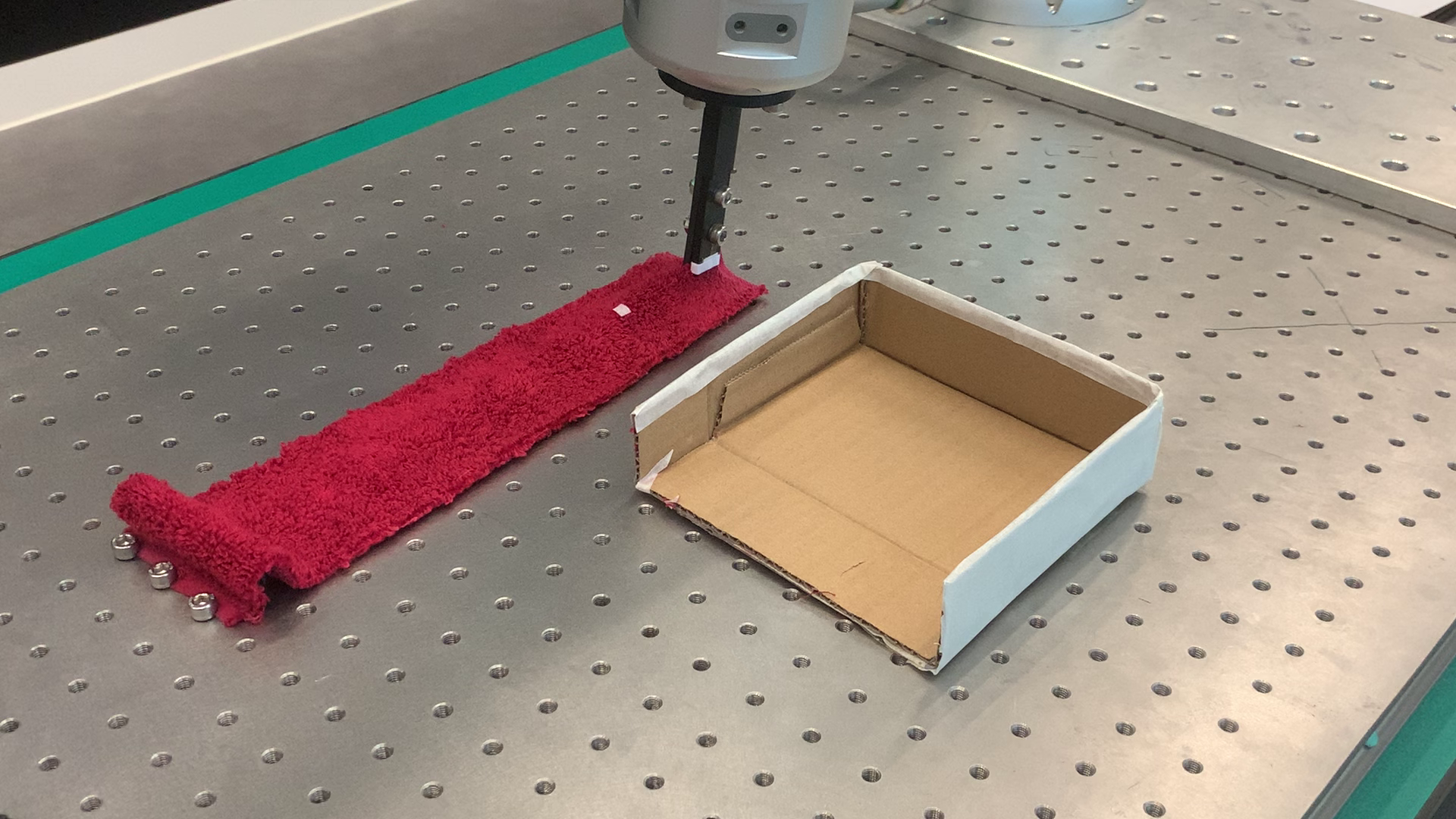}}
    \subfigure[] {
    \includegraphics[width= 0.23 \columnwidth]{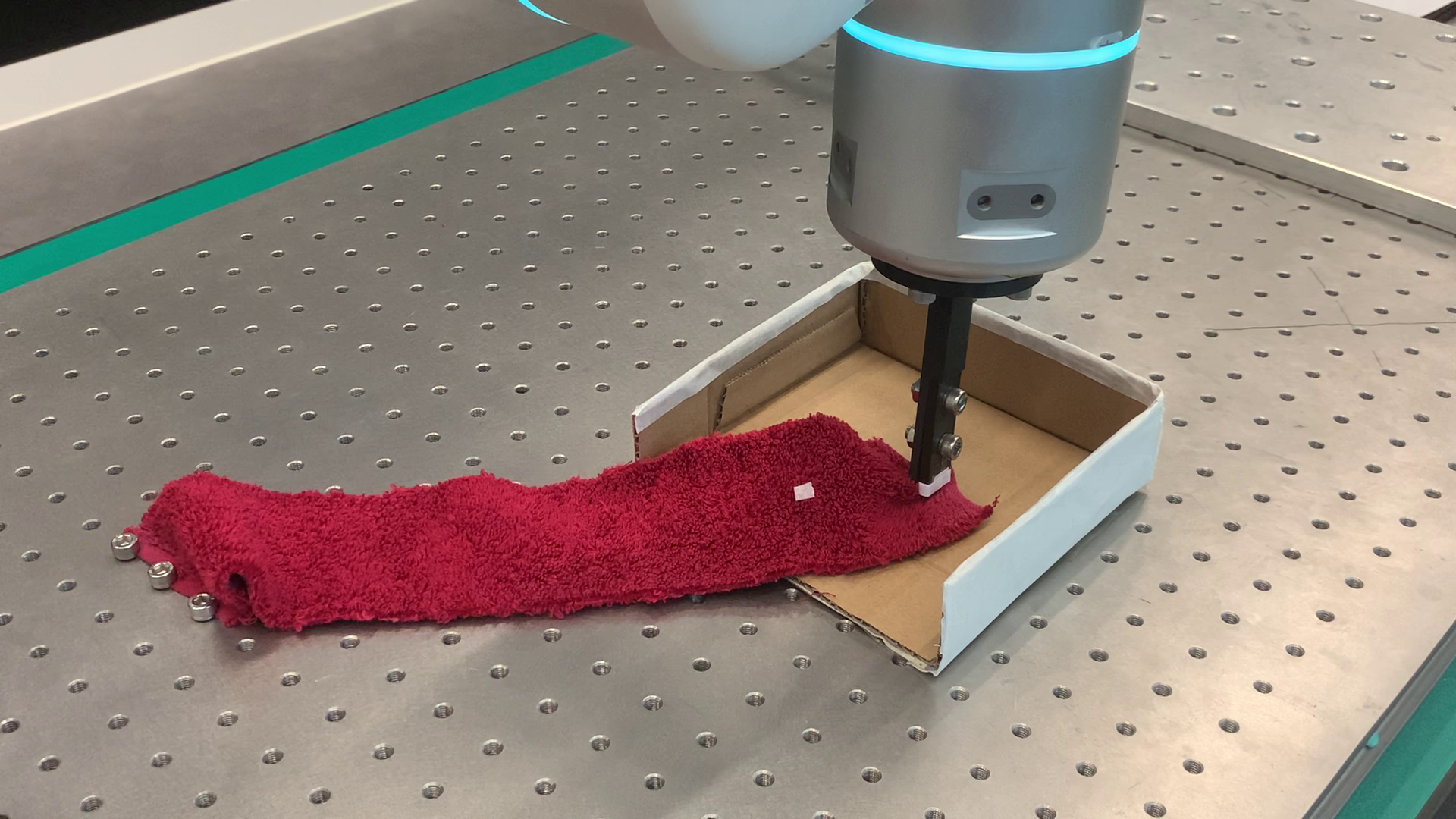}}
    \subfigure[] {
    \includegraphics[width= 0.23 \columnwidth]{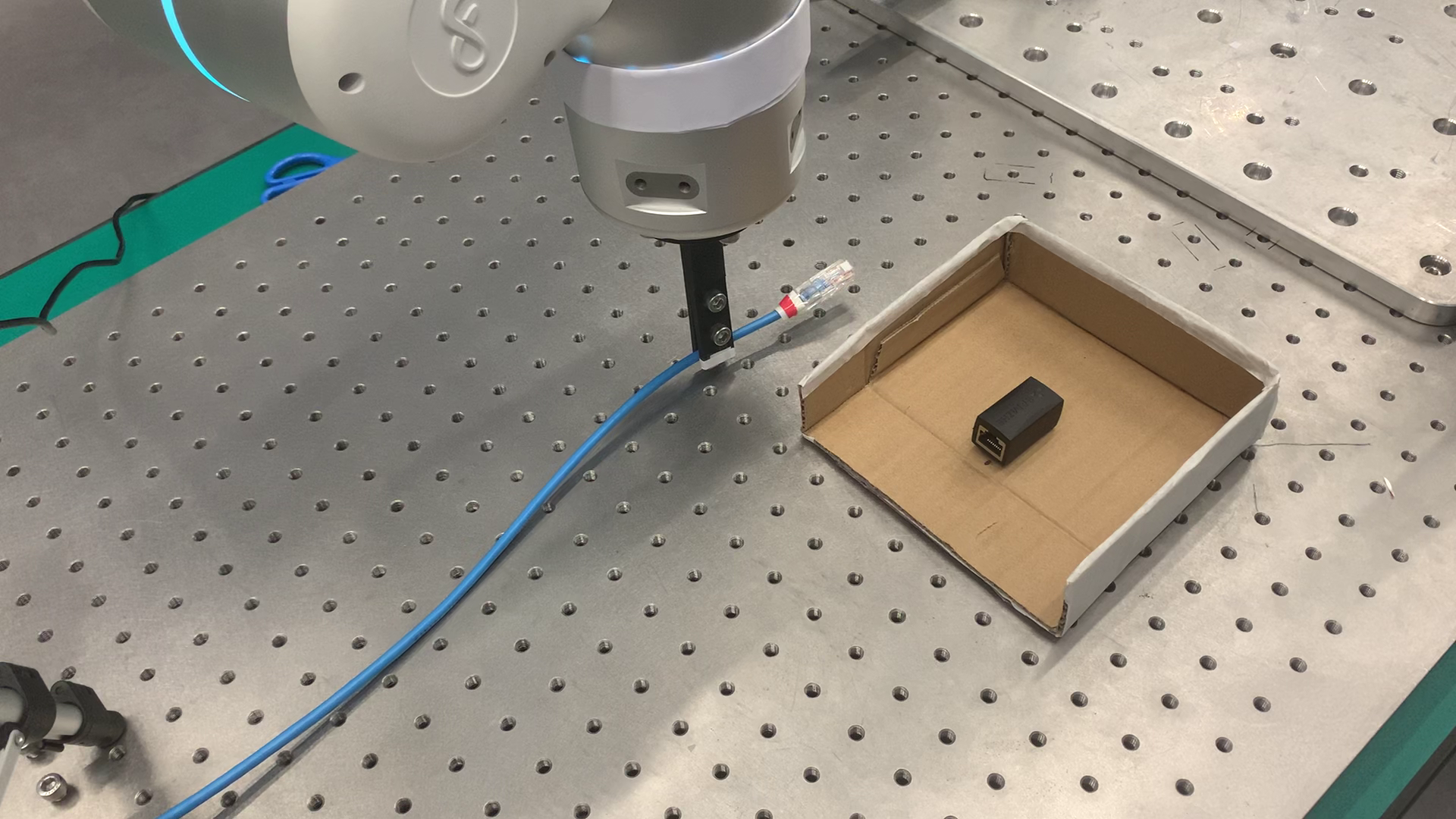}}
    \subfigure[] {
    \includegraphics[width= 0.23\columnwidth]{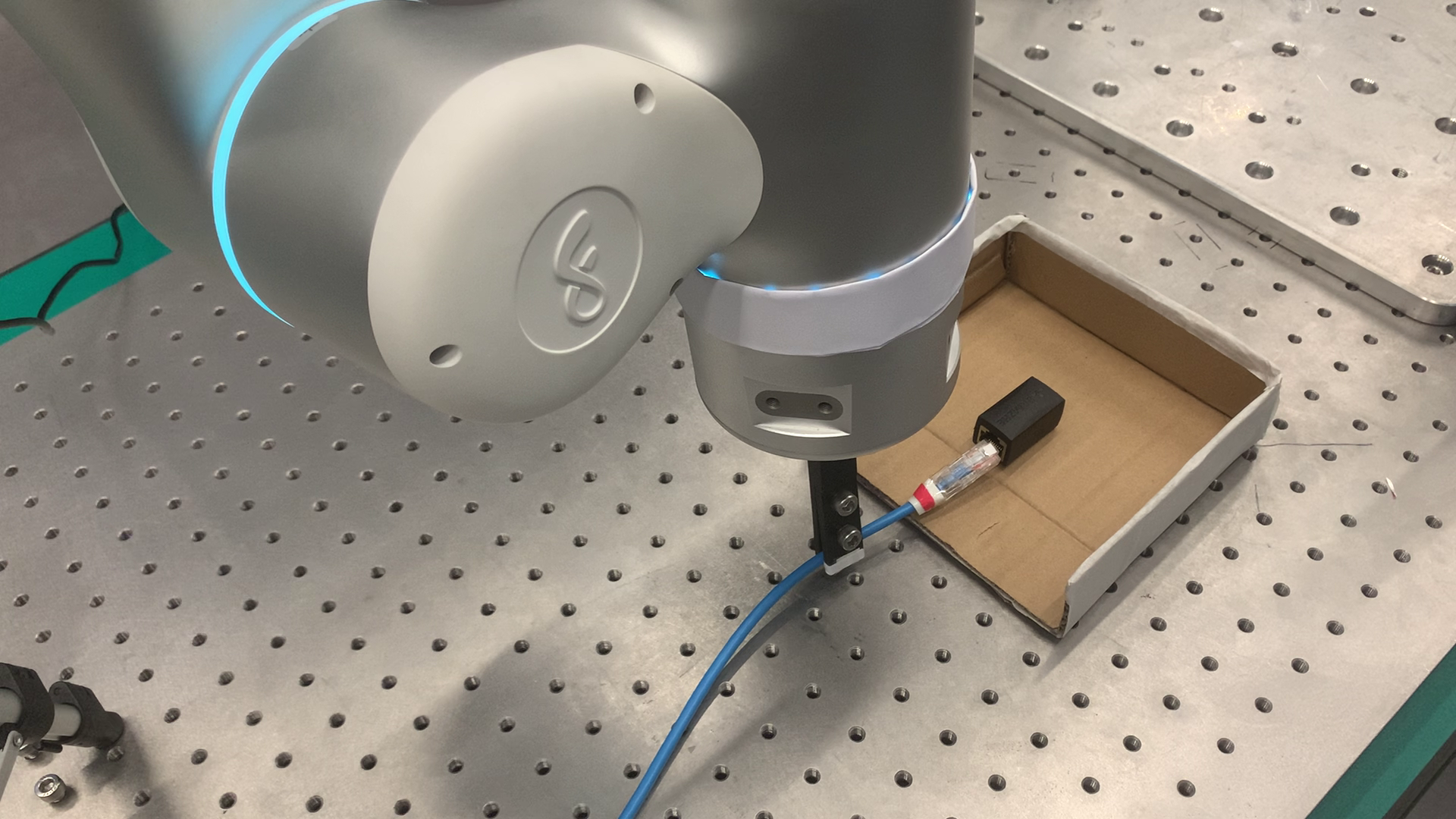}}    
    \endminipage \hfill
    \minipage{2 \columnwidth}
    \centering
    \subfigure[] {
    \includegraphics[width= 0.23 \columnwidth]{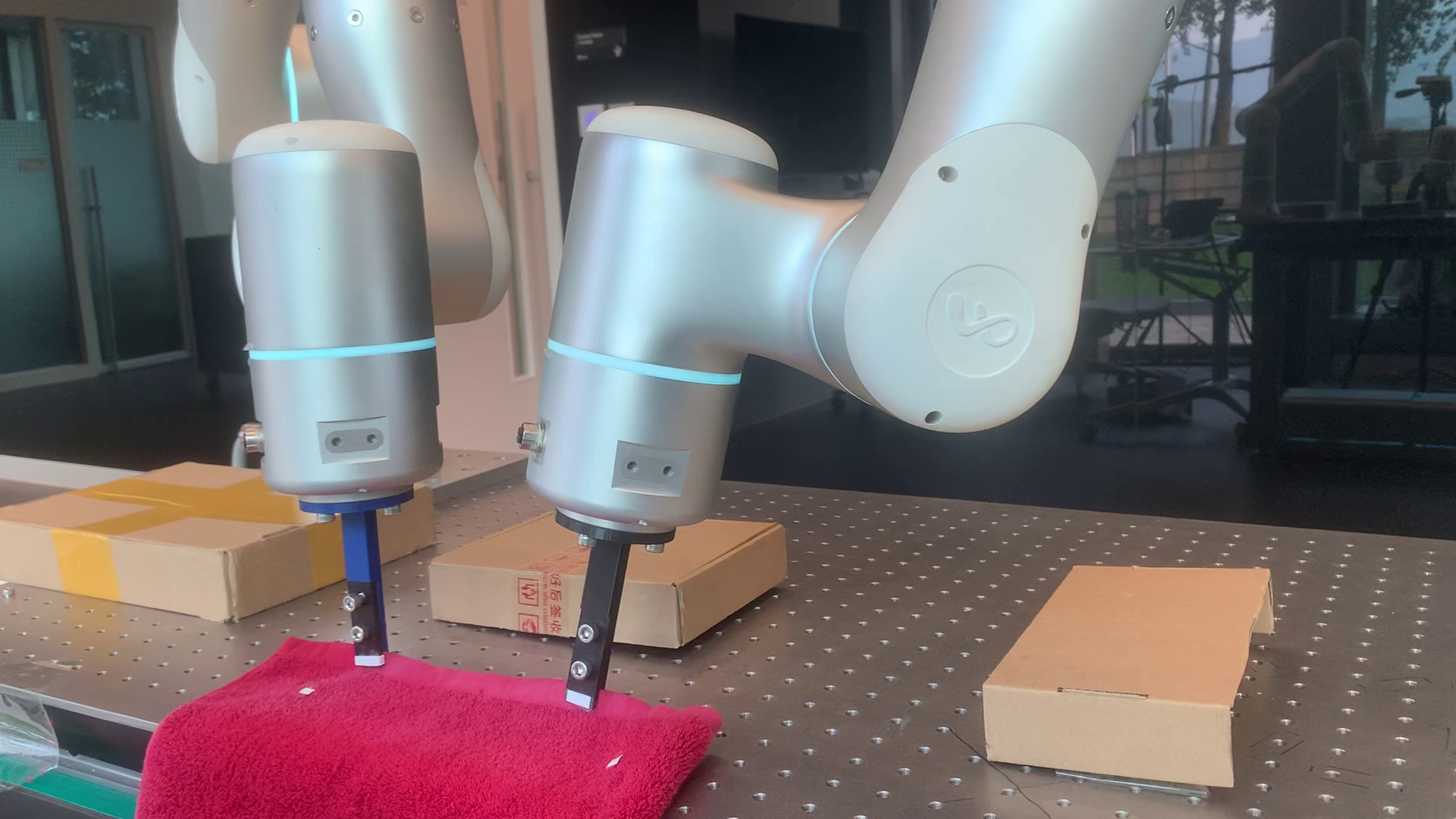}}
    \subfigure[] {
    \includegraphics[width= 0.23 \columnwidth]{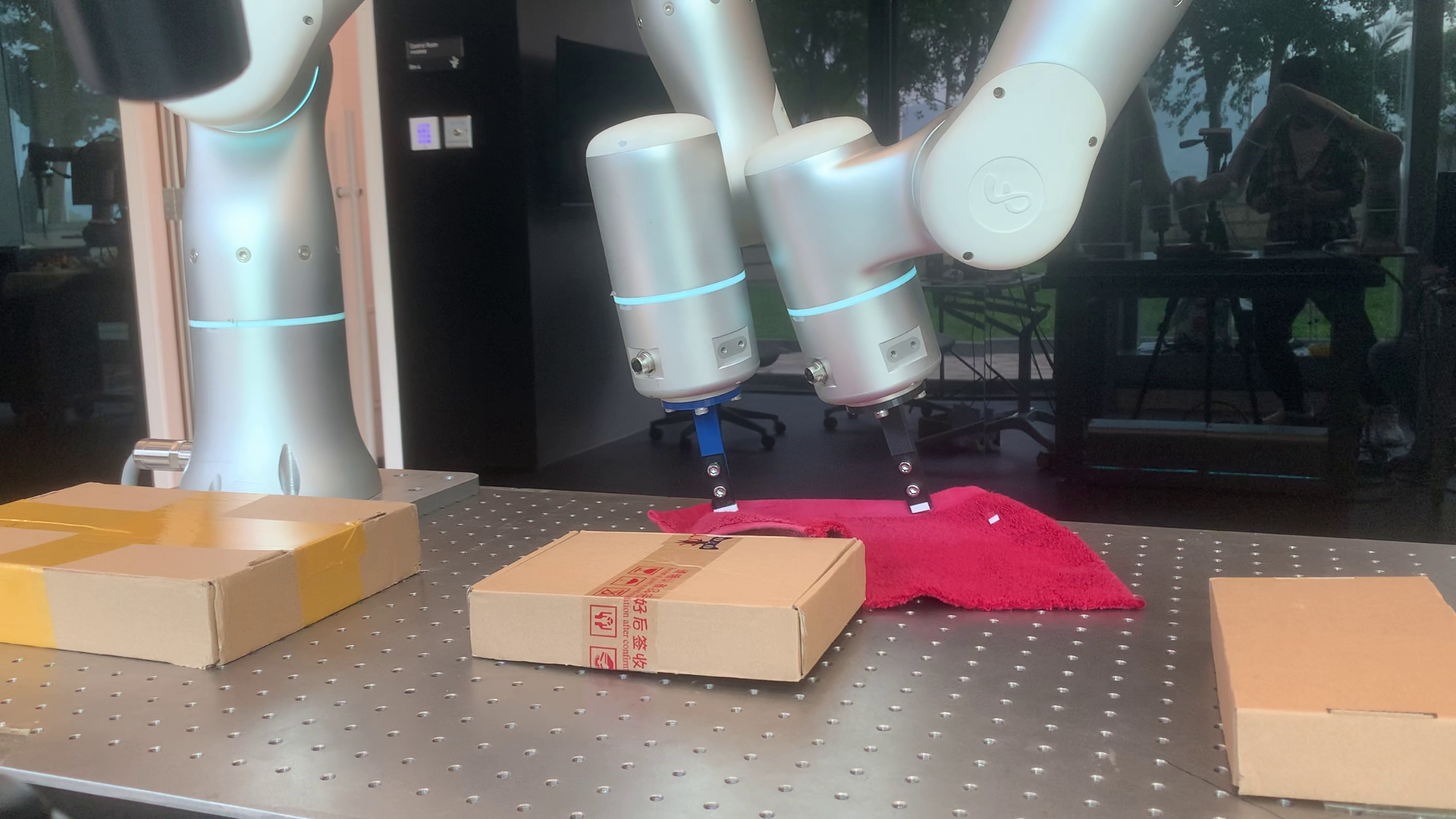}}
    \subfigure[] {
    \includegraphics[width= 0.23 \columnwidth]{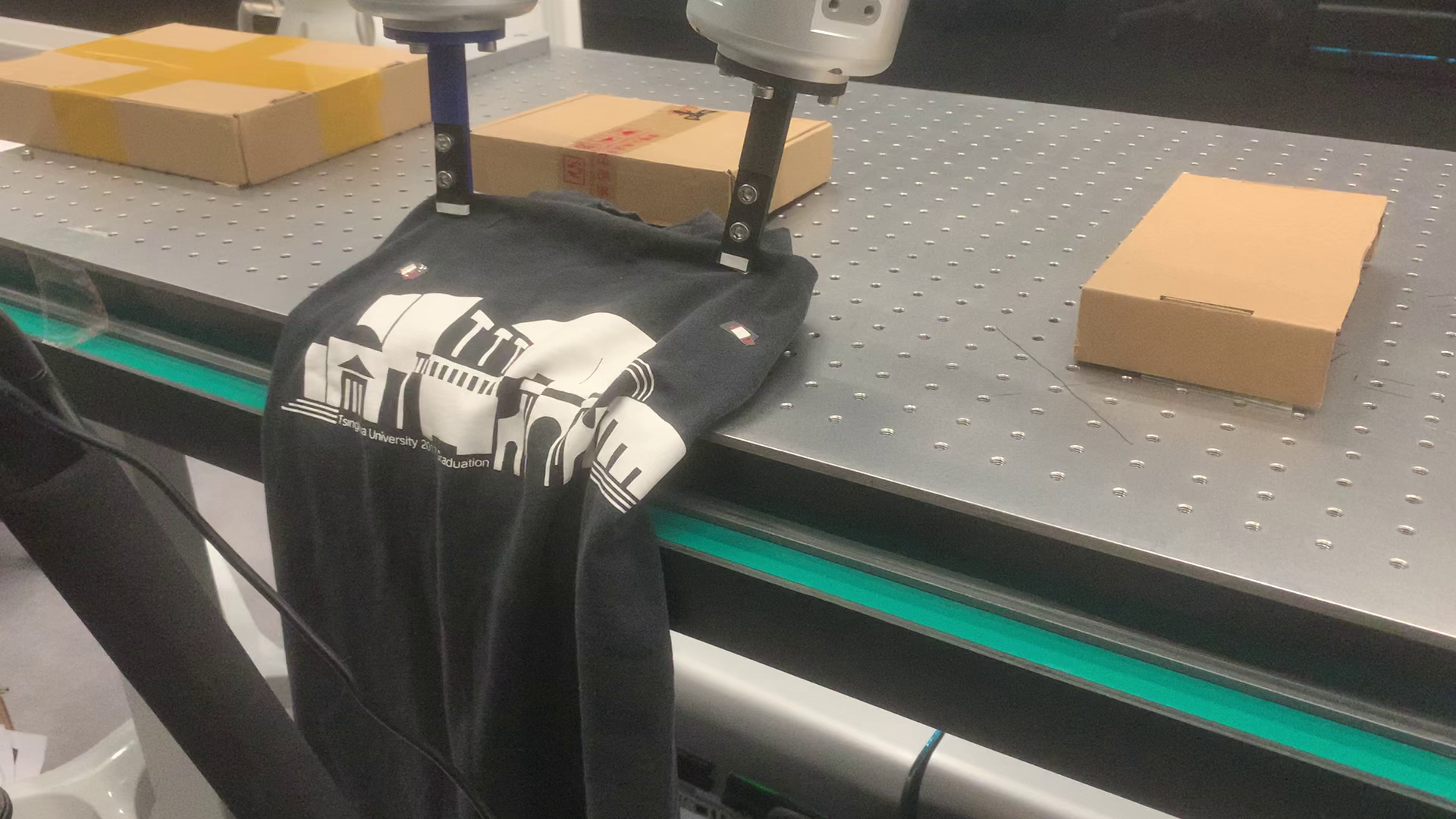}}
    \subfigure[] {
    \includegraphics[width= 0.23 \columnwidth]{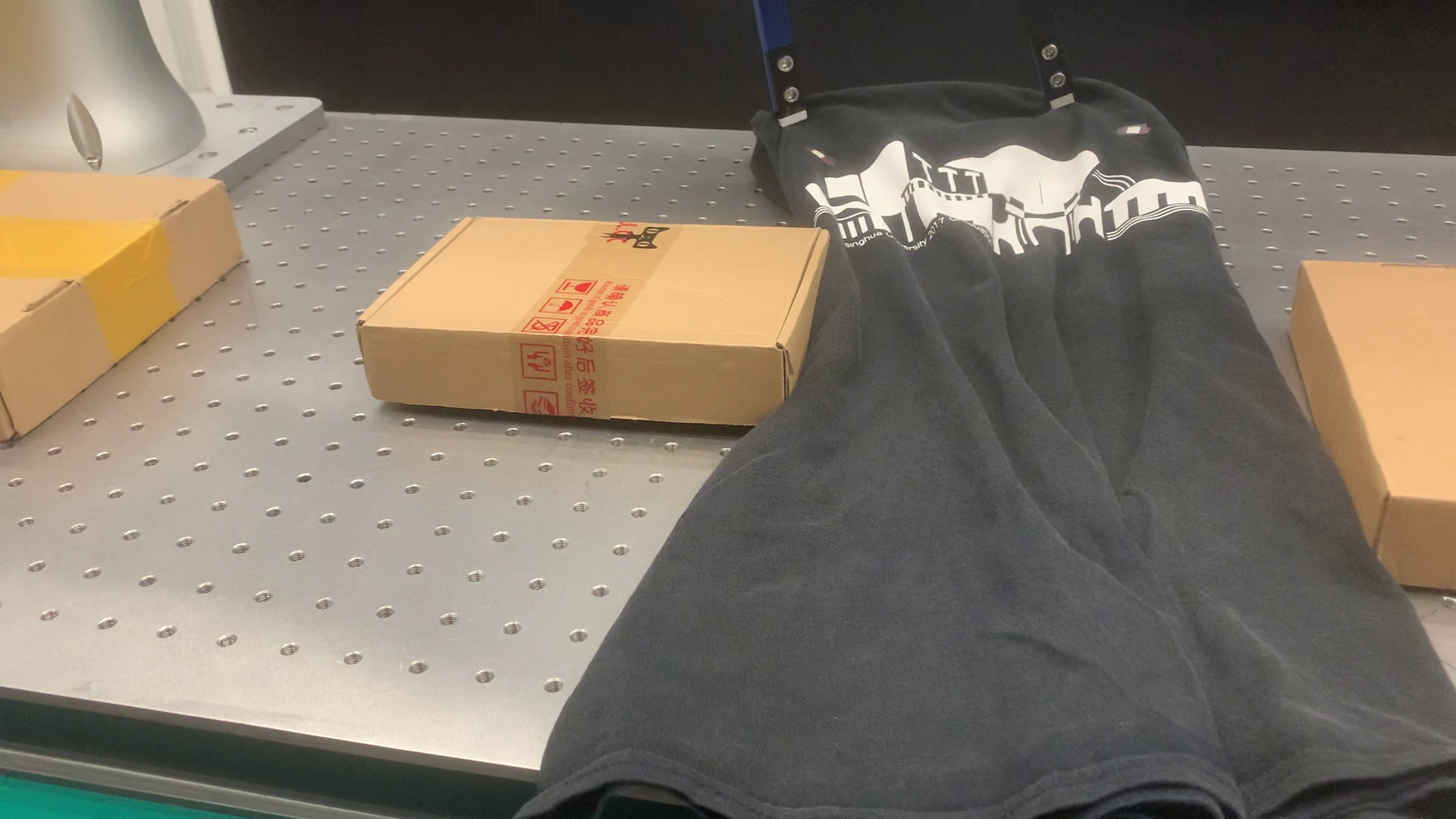}}    
    \endminipage \hfill
    \caption{(a), (b) Initial and final configurations of the fabric sheet in the insertion task. (c), (d) Initial and final configurations of the Ethernet cable in the insertion task. (e), (f) Towel at rest and after passing the chosen passage. (g), (h) T-shirt at rest and after passing the chosen passage.}
    \label{Case Study Snapshots}
    \end{figure*}
In the last part, we present application cases of the overall scheme in some common domestic and industrial scenarios. Two classes of DOM tasks are carried out: box insertion and narrow passage passing. Box-insertion is the combination of single-point manipulation and the box-like obstacle in previous subsections. Specifically, the DO of a prolate profile is initially positioned outside of an open box in the box's close vicinity. Analogously to Fig. \ref{Narrow Passage Task}, the robot needs to move the object into a target region central in the box in a feasible manner. The whole process mimics common intersection motions. The major challenge of accomplishing it lies in figuring out a feasible DO motion and deformation process. Conventional approaches either rely on a deformation process simulated in advance or plan robot motions interactively, both of which are heavily model-based and computationally costly. By simply employing DO feedback points' spatial path set and tracking, the problem is converted into a more tractable one able to involve constraints easily.

Fig. \ref{Case Study Snapshots}(a)-(d) displays insertion tasks manipulating a fabric sheet and an Ethernet cable. Inserting the fabric sheet is similar to experiments previously carried out. The differences are that the obstacle is more complex as a box with one open side (represented by a polygon of three walls) and that the target region is located inside the box. To accomplish the task, the sheet should be first compressed from its initial configuration and then expanded. Such a procedure is naturally encoded in the planned feedback point path. In particular, the task can be clearly described by only a feedback point $\mathbf{s}_1$ picked in the top sheet part, whose spatial path is planned and tracked in execution. Since $\mathbf{s}_1$ lies inside the fabric and the sheet width is large compared to the box and clearance, tracking $\sigma_1$ still may cause a collision. Interleaved local end-effector path planning and tracking will be triggered if the obstacle cannot be bypassed in collision adjustment.

Ethernet cables belong to DLOs that demonstrate distinct properties in contrast to DOs of larger sizes and volumes. DLO manipulation can involve complex procedures like knotting, which explicitly change the homotopy properties of both the DLO itself and the paths of feedback points on it. Moreover, execution of DLO manipulation tasks usually requires iterative re-grasps, which greatly complicates the task. Thus, many specially-designed methods have been developed for DLO manipulation, e.g., \cite{M. Yu 2023}, \cite{T. Tang 2018}-\cite{S. Jin 2022}. Our approach is applicable to DLOs if the prespecified feedback point path homotopy properties hold in the task and manipulation can be conducted with fixed grasps. In the experiment, the feedback point is selected near the cable head and the robot grasps the cable at a distance behind. An Ethernet port is fixed at the box center as the target region. Due to the slim cable size, collision is avoided when following the planned path of the feedback point if a suitable planning interior is used.

The second class of tasks involve constrained environmental setups composed of multiple obstacles to evaluate path set generation and tracking under general practical conditions. As demonstrated in Fig. \ref{Case Study Snapshots}(e)-(h), three obstacles on the table construct two narrow passages of different widths. DOs of large areas are intended to be manipulated to reach the target configuration preset on the other side of passages. A dual-arm system is utilized to execute these tasks considering DOs' larger sizes. Narrow passage passing tasks are common and also studied in \cite{D. Mcconachie 2020} where robot motion is planned based on the simulated DO model. However, the robot motion plan only takes into account feasibility. Selection among passages is not considered. The feature consists of two feedback points $\mathbf{s}_1, \mathbf{s}_2$ placed inside DO but near the boundary to better represent DO's overall size. The pivot is chosen as $\mathbf{s}_1$ on the left. As shown in Fig. \ref{Case Study Robot View}(b), the planned pivot path goes through the wider passage though this results in a longer path length. As the passage width is smaller than $\| \mathbf{s}_{0,1} - \mathbf{s}_{0,2} \|_2$, direct forward path transfer will lead to an infeasible path $\sigma_{t,2}$. Therefore, the final feasible path set is generated following the procedure for general constrained conditions. When passing the passage in tracking, the collision constraint between obstacles and the DO is relaxed to promote the execution progress. DOs of different properties, e.g., towels and T-shirts of significantly larger sizes, are tested in the experiments.

\section{Conclusion}
\label{Conclusion Section}
In this article, we investigated DOM with constraints and proposed the visual feedback vector path set planning and tracking scheme. Particularly, DOM was extended to more general and practical scenarios where constraints were present.
We first formulated constrained tasks in an optimization formalism versatile to involve common constraints and enabled a dynamic constraint imposition mechanism. For task conduction, path set of the visual feedback vector was leveraged and the central issue in its planning was simultaneous feasible path generation for multiple feedback points in constrained environments. To achieve this, passage-aware pivot path planning was proposed and feasibility requirements for path sets were analyzed as planning prerequisites. Then, efficient algorithms for path set generation based on passage-aware pivot path planning, repositioning, and deformable path transfer were designed for general constrained conditions. In manipulation control, we proposed a holistic tracking control architecture for both normal path set tracking and tracking with constraints, in which the constraint regulation and local minimum resolution were embedded. Experiments of DOM in different constrained setups showed that the proposed methods could empower robots to perform DOM under constraints and had superior performances over conventional pure control methods in challenging situations.

The recent development of pixel-level tracking techniques in computer vision will provide the proposed scheme with more robust segmentation and tracking capacities \cite{Q. Wang 2023}. However, the path set tracking control for task execution is still low-level and restricted, unable to plan higher-level of robot motions in complex DOM tasks. 
In the future, we plan to connect the current pipeline with automatic task-specific DO feature and key point extraction modules in 2D and 3D vision. Moreover, the presented methods have good generality in many other robotics domains. For instance, methods in the planning part can be readily extended to higher dimensional spaces with necessary modifications. In fields such as mobile robotics and aerial manipulation by swarms, the concepts and approaches of path sets in this work are promising.

\iftrue
\section*{Appendix A\\ Target Determination for Visual Feedback Vector}
\label{Appendix A}
If $S_d$ is not entirely specified by $\mathbf{y}_d$, it needs to be determined to provide a definite target for path set planning and tracking. To construct the $S_d$ evaluation function $\mathcal{J}(\cdot)$ in (\ref{sd optimization}), it first considers the obstacle avoidance of the DO approximated by $S_d$'s distance to the obstacle. Then, it further includes the manipulation cost quantified by the spatial distribution difference between $S$ and $S_d$. So $\mathcal{J}(\cdot)$ is defined as
\begin{equation*}
    \mathcal{J}(S_0, S) = (1 - \lambda) \mathcal{D}(S_0, S) - \lambda d(S, \mathcal{E}).
\end{equation*}
$d(S, \mathcal{E})$ is a generic distance between $S$ and obstacle $\mathcal{E}$ to approximate $d(\mathcal{O}, \mathcal{E})$. $\mathcal{D}(S_0, S)$ gauges the manipulation cost. $\lambda \in [0, 1]$ adjusts their weights.

For clarity, feedback points in $S$ are also called \textit{complete} if their target positions are given in $\mathbf{y}_d$ and \textit{incomplete} otherwise, designated by $S_c$ and $S_{ic}$, respectively, i.e., $S = [S_c\T, \, S_{ic}\T]\T$. $d(S, \mathcal{E})$ is computed as the average of incomplete points' distances to the obstacle to measure obstacle avoidance
\begin{equation*}
    d(S, \mathcal{E}) = \frac{1}{\text{card}(S_{ic})} \sum_{\mathbf{s}_i \in S_{ic}} d(\mathbf{s}_i, \mathcal{E}).
\end{equation*}

$\mathcal{D}(S_0, S)$ is a deformation-energy-like function reflecting the manipulation cost through the structural difference between $S$ and $S_0$. Analogously to soft robot shape modeling \cite{G. Fang 2020}, the central idea is that if $S$ and $S_0$ share similar relative distributions of points, $S$ can be achieved more easily from $S_0$. To quantify this structural difference, a reference distribution for $S_d$ is first generated. The pivot $\mathbf{s}_p$ refers to the point in $S_c$ most influential in the task under some given criterion. For instance, $\mathbf{s}_i$ most influential on the feature $\mathbf{y}$ can be picked by the Frobenius norm of the feature Jacobian, i.e.,
\begin{equation*}
   \mathbf{s}_{d,p} = \arg \, \operatorname*{max}_{\mathbf{s}_i \in S_{d|c}} \; \| \pdv{\mathbf{y}}{\mathbf{s}_i} \|_F^2
\end{equation*}
where $S_{d|c}$ is the known target of $S_c$ and $\mathbf{s}_{d,p}$ is the desired position of $\mathbf{s}_p$. Other criteria, such as the point with the maximum displacement, may also be employed. The translational vector $\mathbf{v} = \mathbf{s}_{d,p} - \mathbf{s}_{0,p}$ is then utilized to define the position variation for incomplete points. The reference target of an incomplete point is then given by
\begin{equation*}
    \mathbf{s}_{ref,i} = \mathbf{R}_{S_c} \mathbf{s}_{0,i} + \mathbf{v}.
\end{equation*}
$\mathbf{R}_{S_c}$ is the rotation matrix induced by complete components $S_c$, which can be extracted by considering the transformation between $S_{0 | c}$ and $S_{d | c}$ using registration techniques \cite{B. K. Horn 1988}-\cite{H. Yang 2021}. If there is only one complete point, $\mathbf{R}_{S_c} = \mathbf{I}$.
\begin{figure}[t]
    \minipage{1 \columnwidth}
    \centering
    \includegraphics[width= 0.9 \columnwidth]{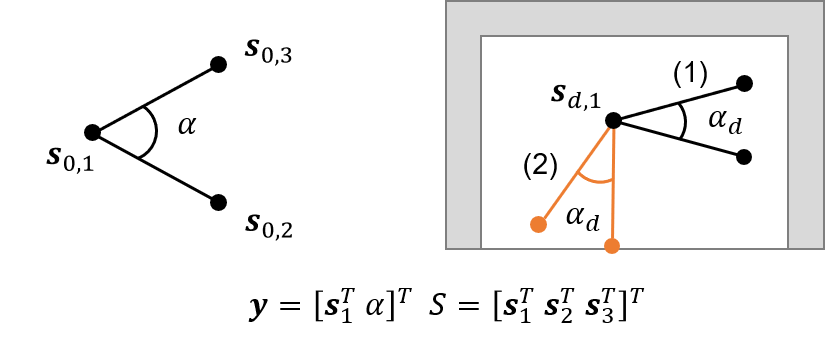}    
    \endminipage \hfill
    \caption{In feature-based deformation description, the desired feedback vector $S_d$ can be undetermined for a specified desired feature $\mathbf{y}_d$. For the point-angle feature above, there are infinitely many $S$ with $\mathbf{y}_d = \mathcal{F}(S)$ such as the shown different configurations of $(1)$ and $(2)$. The aim is to find the $S_d$ optimizing some constructed criteria in $\mathcal{J}(\cdot)$.}
    \label{Angle Feature Target Determination}
    \end{figure}
\begin{algorithm}[t]
    \nl $ \mathbf{s}_{d,p} \leftarrow \arg \, \operatorname*{max}_{\mathbf{s}_i \in S_{d | c}} \; \| \pdv{\mathbf{y}}{\mathbf{s}_i} \|_F^2 $\;
    \nl $ \mathbf{v} \leftarrow \mathbf{s}_{d,p} - \mathbf{s}_{0,p}$\;
    \nl $ \mathbf{R}_{S_c} \leftarrow $ \texttt{ExtractRotation}$(S_{0|c}, S_{d|c})$\;
    \nl \ForEach{$\mathbf{s}_{i} \in S_{0 | ic}$} {
        \nl $\mathbf{s}_{ref, i} \leftarrow \mathbf{R}_{S_c}\mathbf{s}_{i} + \mathbf{v}$\;
        \nl $S_{ref | ic} \leftarrow  S_{ref | ic} \cup \{\mathbf{s}_{ref, i}\}$\;
    }
    \nl $S_{d,ref} \leftarrow [S_{d | c}\T, \, S_{ref | ic}\T]\T$\;
    \nl $S_d \leftarrow$ solve (\ref{sd optimization})\; 
    \nl \Return $S_d$\;
\caption{Target Determination for Visual Feedback Vector.}
\label{Target Feedback Vector Determination Algorithm}
\end{algorithm}

In the final step, the reference distribution of $S_d$ is given by $S_{d,ref} = [S_{d|c}\T, \, S_{ref|ic}\T]\T$. $\mathcal{D}(S_0, S)$ is assigned as the average of incomplete points' distances between $S$ and $S_{d,ref}$
\begin{equation*}
    \mathcal{D}(S_0, S) = \frac{1}{\text{card}(S_{ic})} \sum_{\mathbf{s}_i \in S_{ic}} \| \mathbf{s}_i - \mathbf{s}_{ref,i} \|_2.
\end{equation*}
It is worth mentioning that $S_{d,ref}$ may be an infeasible configuration for $S$, but it is acceptable since $S_{d,ref}$ only acts as a reference. Algorithm \ref{Target Feedback Vector Determination Algorithm} outlines the proposed procedure of $S_d$ determination and an example is shown in Fig. \ref{Angle Feature Target Determination}.

\section*{Appendix B\\ Symmetry of Path Homotopic-like Relation}
Upon symmetry of the path homotopic-like relation, homotopy properties and path transfer operations between any two point paths become unordered.  
Assume path $\sigma'_{j,i}$ and $\sigma_i$ are path homotopic, denoted by $\sigma'_{j, i} \simeq_p \sigma_i$. $\sigma_j$ and $\sigma_i$ are thus path homotopic-like by our definition. To prove symmetry of the path homotopic-like relation (i.e., $\sigma_i$ and $\sigma_j$ are also path homotopic-like), we need to show $\sigma'_{i,j} \simeq_p \sigma_j$. Suppose $\psi'_{j,i}(x, \tau)$ is a homotopy from $\sigma_i$ to $\sigma'_{j,i}$ and
\begin{align*}
    \psi'_{j,i}(0, \tau) &= \sigma_i(\tau) \\
    \psi'_{j,i}(1, \tau) &= \sigma'_{j,i}(\tau).
\end{align*}

Assume that on path $\sigma'_{j,i}$, $\sigma'_{j,i}(\tau_1) = \mathbf{s}_{0,j}$ and $\sigma'_{j,i}(\tau_2) = \mathbf{s}_{d,j}$ for two path parameters $\tau_1 < \tau_2$ given by path length parametrization. Then $\psi'_{j,i}(x, \tau_1)$ and $\psi'_{j,i}(x, \tau_2)$ for $x \in [0, 1]$ can represent the paths from $\sigma_i(\tau_1)$ to $\mathbf{s}_{0,j}$ and $\sigma_i(\tau_2)$ to $\mathbf{s}_{d,j}$ determined by the homotopy $\psi'_{j,i}(x, \tau)$, respectively. For a fixed $x_0 \in [0, 1]$, we define the intermediate positions of $\mathbf{s}_{0,j}$ and $\mathbf{s}_{d,j}$ as
\begin{align*}
    \mathbf{s}'_{0,j} &= \psi'_{j,i}(x_0, \tau_1) \\ \mathbf{s}'_{d,j} &= \psi'_{j,i}(x_0, \tau_2)
\end{align*}
(see Fig. \ref{Homotopic-like Symmetry Proof}). Consider the concatenated path
\begin{equation*}
    \sigma'_{x_0} = \psi'_{j,i}(1 \sim x_0, \tau_1) * \psi'_{j,i}(x_0, \tau_1 \sim \tau_2) * \psi'_{j,i}(x_0 \sim 1, \tau_2)
\end{equation*}
where $\sim$ represents the traversal of the closed interval from left to right. Intuitively, $\sigma'_{x_0}$ starts at $\mathbf{s}_{0,j}$, passes through $\mathbf{s}'_{0,j}$ and $\mathbf{s}'_{d,j}$ in order, and terminates at the final point $\mathbf{s}_{d,j}$ as shown by the path marked by green arrows in Fig. \ref{Homotopic-like Symmetry Proof}. 
\begin{figure}[t]
    \minipage{1 \columnwidth}
    \centering
    \includegraphics[width= 0.81 \columnwidth]{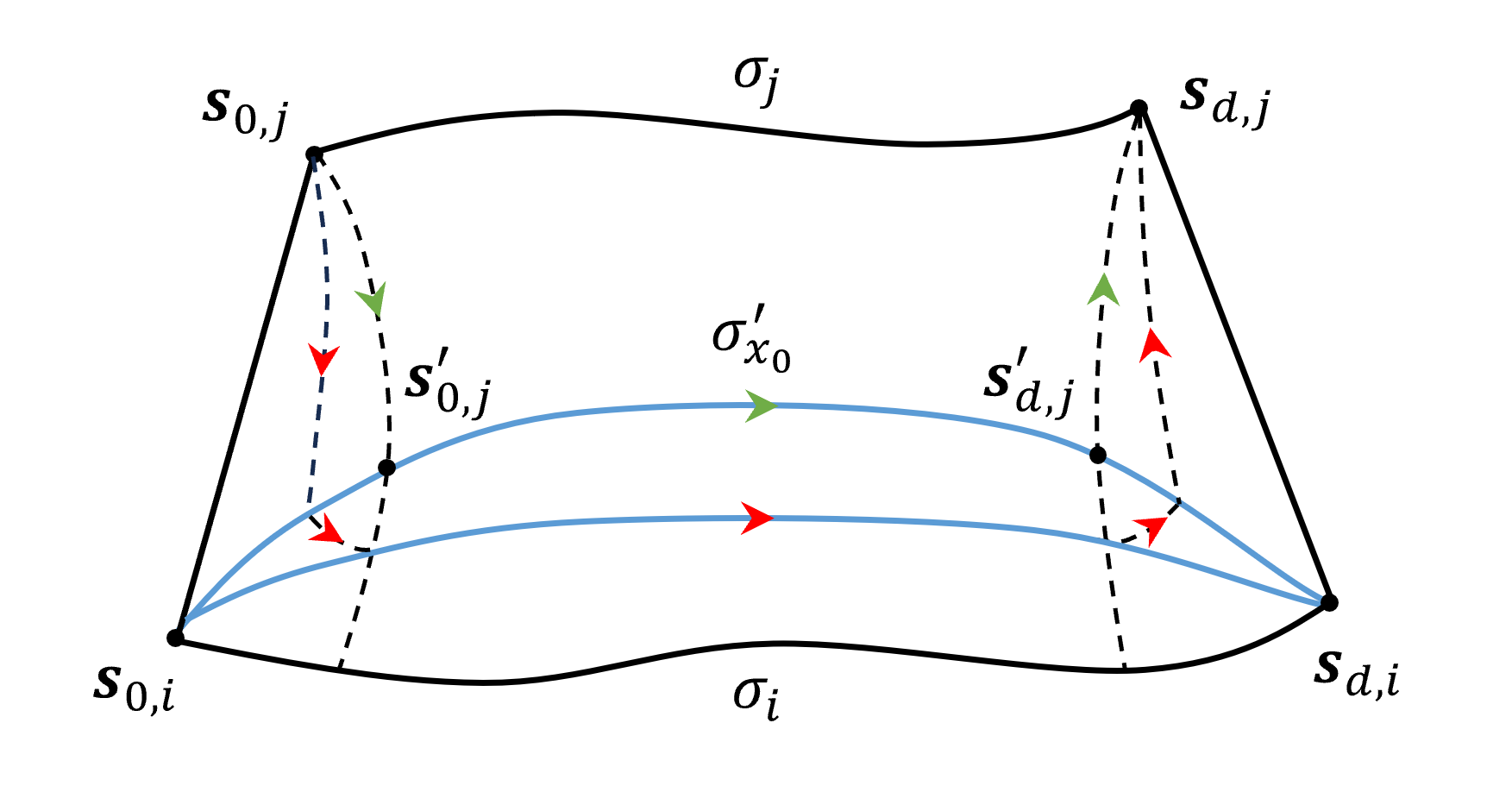}
    \endminipage \hfill
    \caption{$\sigma'_{x_0}$ is the path marked by green arrows. $\psi_2(x)$ is composed of the five segments with red arrows. $\psi'_{j,i}(x, \tau_1)$ and $\psi'_{j,i}(x,\tau_2)$ are depicted by the dashed black lines from $\mathbf{s}_{0,j}$ and $\mathbf{s}_{d,j}$ to path $\sigma_i$.}
    \label{Homotopic-like Symmetry Proof}
\end{figure}

Next, we will show that $\sigma'_{x_0} \simeq_p \sigma_j$ and $\sigma'_{x_0} \simeq_p \sigma'_{i,j}$. The former one is easy to validate since one homotopy $\psi_1$ from $\sigma'_{x_0}$ to $\sigma_j$ can be constructed as
\begin{multline*}
    \psi_{1}(x) = \psi'_{j,i}(1 \sim x_0 + x(1 - x_0), \tau_1) \\ *\psi'_{j,i}(x_0 + x(1 - x_0), \tau_1 \sim \tau_2) * \psi'_{j,i}(x_0 + x(1 - x_0) \sim 1, \tau_2).
\end{multline*}
When only the first argument is provided in the homotopy like $\psi_1(x)$, it refers to the intermediate path $\psi_{1}(x, 0 \sim 1)$ for notational simplicity. 
$\psi_1$ is continuous for $\psi'_{j,i}$ is continuous. Recall that $\sigma_{i,j}' := \sigma_{j,i|0} \,*\, \sigma_i \,*\, \sigma_{i,j|d}$. To demonstrate the path homotopic relation between $\sigma'_{x_0}$ and $\sigma'_{i,j}$, we explicitly construct the homotopy $\psi_2(x)$ composed of the following five sequential segments $\psi^i_2(x), i = 1, 2, 3, 4, 5$:
\begin{align*}
    \psi^1_2(x) &= \psi'_{j,i}(1 \sim x_0, \tau_1 \sim (1 - x)\tau_1)  \\
    \psi^2_2(x) &= \psi'_{j,i}(x_0 \sim (1 - x)x_0, (1 - x)\tau_1 \sim \tau_1)  \\
    \psi^3_2(x) &= \psi'_{j,i}((1 - x)x_0, \tau_1 \sim \tau_2)  \\
    \psi^4_2(x) &= \psi'_{j,i}((1 - x)x_0 \sim x_0, \tau_2 \sim \tau_2 + x(1 - \tau_2))  \\
    \psi^5_2(x) &= \psi'_{j,i}(x_0 \sim 1, \tau_2 + x(1 - \tau_2) \sim \tau_2).
\end{align*}
Namely, $\psi_2(x) = \psi^1_2(x) * \psi^2_2(x) * ..., \psi^5_2(x)$ and $\psi_2(0) = \sigma_{x_0}, \psi_2(1) = \sigma'_{i,j}$. When both arguments in the homotopy $\psi_2(x, \tau)$ are ranges, $x$ and $\tau$ vary linearly to cover their respective ranges. Each segment $\psi^i_2$ is continuous since $\psi'_{j,i}(x, \tau)$ is continuous w.r.t. both $x$ and $\tau$, and so is the resulting concatenated path by the gluing lemma. As for the feasibility, both $\psi_1$ and $\psi_2$ are guaranteed to be feasible because their domains are subsets of the domain of $\psi'_{j,i}$.

Given $\sigma'_{x_0} \simeq_p \sigma_j$ and $\sigma'_{x_0} \simeq_p \sigma'_{i,j}$, we have $\sigma'_{i,j} \simeq_p \sigma_j$ using that $\simeq_p$ is an equivalence relation, and according to the definition, $\sigma_i$ and $\sigma_j$ are path homotopic-like.
\fi

\section*{Acknowledgement}
The authors would like to thank Minjian Feng for his help in experiments in the application case study part.

\end{document}